\documentclass[english]{taudissertation}

\usepackage{amsmath,amssymb}
\usepackage{lipsum}
\usepackage{hyperref}
\usepackage{subcaption}

% ...

\makeglossaries
\usepackage{hyperref}

\newglossaryentry{PMSM}{
name={PMSM},
description={Permanent-magnet synchronous motor}
}

\newglossaryentry{HDMM}{
name={HDMM},
description={Heavy-duty mobile machine}
}

\newglossaryentry{Dof}{
name={Dof},
description={Degrees of freedom}
}

\newglossaryentry{RAC}{
name={RAC},
description={Robust adaptive control}
}

\newglossaryentry{BLF}{
name={BLF},
description={Barrier Lyapunov function}
}

\newglossaryentry{RFB}{
name={RFB},
description={Radial baasis function}
}

\newglossaryentry{DNN}{
name={DNN},
description={Deep neural network}
}

\newglossaryentry{AI}{
name={AI},
description={Artificial intelligence}
}

\newglossaryentry{ISO}{
name={ISO},
description={International Organization for Standardization}
}

\newglossaryentry{EMLA}{
name={EMLA},
description={Electromechanical linear actuator}
}

\newglossaryentry{WMR}{
name={WMR},
description={Wheeled mobile robot}
}

\newglossaryentry{SSWMR}{
name={SSWMR},
description={Skid-steer wheeled mobile robot}
}

\newglossaryentry{HDA}{
name={HDA},
description={Hydraulic-driven actuator}
}

\newglossaryentry{EDA}{
name={EDA},
description={Electrical-driven actuator}
}

\newglossaryentry{LM}{
name={LM},
description={Levenberg–Marquardt}
}

\newglossaryentry{PID}{
name={PID},
description={Proportional–integral–derivative}
}

\newglossaryentry{P}{
name={P},
description={Publication}
}

\newglossaryentry{C}{
name={C},
description={Contribution}
}

\newglossaryentry{RQ}{
name={RQ},
description={Research question}
}

\newglossaryentry{PDA}{
name={HDA},
description={Pneumatic-driven actuator}
}

\newglossaryentry{IWD}{
name={IWD},
description={In-wheel Drive}
}

\newglossaryentry{BEV}{
name={BEV},
description={Battery electric vehicle}
}

\newglossaryentry{CAC}{
name={CAC},
description={Command-filter-approximator-based adaptive control}
}

\newglossaryentry{ANATC}{
name={ANATC},
description={Adaptive neural asymptotic tracking control}
}

\newglossaryentry{CBF}{
name={CBF},
description={Control barrier function}
}

\newglossaryentry{HRI}{
name={HRI},
description={Human-robot interaction}
}

\newglossaryentry{VDC}{
name={VDC},
description={Virtual decomposition control}
}

\newglossaryentry{VPF}{
name={VPF},
description={Virtual power flow}
}

\newglossaryentry{RAMS}{
name={RAMS},
description={Reliability, Availability, Maintainability, and Safety}
}

\newglossaryentry{ODE}{
name={ODE},
description={Ordinary differential equation}
}

\newglossaryentry{BAC}{
name={BAC},
description={Backstepping-based adaptive control}
}

\newglossaryentry{AATC}{
name={AATC},
description={Adaptive active torque control}
}

\newglossaryentry{AFTC}{
name={AFTC},
description={Adaptive fault-tolerant control}
}

\newglossaryentry{DVSC}{
name={DVSC},
description={Decentralized valve-structure control}
}

\newglossaryentry{DRL}{
name={DRL},
description={Deep reinforcement learning}
}

\newglossaryentry{HITL}{
name={HITL},
description={Human-in-the-loop}
}

\newglossaryentry{MPD}{
name={MPD},
description={Multi-purpose deployer}
}

\newglossaryentry{TL}{
name={TL},
description={Teaching–Learning}
}

\newglossaryentry{IHA}{
name={IHA},
description={Innovative hydraulics and automation}
}

\newglossaryentry{TAU}{
name={TAU},
description={Tampere University}
}

\newglossaryentry{HMI}{
name={HMI},
description={Human-machine interface}
}

\newglossaryentry{MFAC}{
name={MFAC},
description={Model-free adaptive control}
}

\newglossaryentry{BSMC}{
name={BSMC},
description={Backstepping sliding mode control}
}

\newglossaryentry{HDRM}{
name={HDRM},
description={Heavy-duty robotic manipulator}
}

\newglossaryentry{SBC}{
name={SBC},
description={Subsystem-based Control}
}

\newglossaryentry{DSC}{
name={DSC},
description={Dynamic surface control}
}

\newglossaryentry{DADS}{
name={DADS},
description={Deadzone-adapted disturbance suppression}
}

\hypersetup{hidelinks}

\begin{document}
\let\cleardoublepage\clearpage

\makeatletter
\renewcommand\maketitle{%
  \thispagestyle{empty}
  \hspace{0pt}\vfill
  {\centering\LARGE\sffamily\MakeUppercase{\@author}\par}
  \vspace{30pt}
  {\centering\huge\rmfamily\@title\par}
  \vspace{1.5em}
  {\centering\large\itshape\rmfamily\@subtitle\par}
  \vfill
  \clearpage 
}
\makeatother


\frontmatter


\title{Robust Deep Learning Control with Guaranteed Performance for Safe and Reliable Robotization in Heavy-Duty Machinery}
\author{Mehdi Heydari Shahna}
\maketitle


\chapter*{Preface}
This thesis reflects the author’s dedicated work from 2023 to 2025, driven by a commitment to contribute meaningfully to the field. The research aims to bridge the gap between control theory and learning methods, with particular attention to the ethical integration of AI into real-world applications where safety and innovation must go hand in hand. This project was supported by the Business Finland Partnership Project, 'Future All-Electric Rough Terrain Autonomous Mobile Manipulators' under Grant No. 2334/31/2022.

I am profoundly grateful to Prof. Jouni Mattila for allowing me to begin my doctoral studies under his supervision. His constant smile and endless trust not only fostered a reassuring atmosphere but also taught me, perhaps unintentionally, that great leadership often begins with a good sense of humor and an encouraging nod.

I also extend my sincere appreciation to the preliminary examiners, Prof. Mario Duran and Prof. Alessandro Macchelli, for their meticulous review of this thesis and their insightful, constructive, and encouraging feedback. I’m also profoundly thankful to the colleagues who walked this path with me; your support and collaboration meant the world.

My deepest gratitude goes to my beloved wife, Fatemeh, whose pride in my achievements, unwavering dedication, and boundless forgiveness created a foundation of unconditional support, giving me the courage and freedom to pursue my ambitions without fear of failure.

At the end of this sweet but challenging road, I offer a reflection born of my experiences: ``Be generous with what you know, for in return, you often receive the distilled wisdom of others, earned over a lifetime, in a moment."

\vspace{15pt}
Mahdi Heydari Shahna

Tampere, Nov 2025

\chapter*{Abstract}

Today’s heavy-duty mobile machines (HDMMs) face two major transitions to advanced technologies: a shift from diesel-hydraulic to clean electric systems driven by climate goals, and a gradual move from human supervision toward greater autonomy. Firstly, diesel-powered hydraulic systems have long been the dominant actuation mechanism in HDMMs. Consequently, transitioning to fully electric systems, whether through direct replacement or complete system redesign, poses significant technological and economic challenges. Secondly, although advanced artificial intelligence (AI) technologies hold great promise for enabling higher levels of autonomy, their adoption in HDMMs remains limited due to stringent safety standards, and these machines still rely heavily on human supervision.

This dissertation aims to develop a novel control framework that: 1) reduces the complexity of control design for electrification of HDMMs by introducing a generic modular approach that is independent of the energy source and facilitates future system modifications; and
2) establishes hierarchical control policies that enable the partial integration of AI technologies into HDMMs while guaranteeing safety-defined performance and system stability.

To achieve this goal, five interrelated research questions (RQs) were formulated, which align with three overarching lines of investigation. The first line focuses on developing a generic robust control strategy for multi-body HDMMs that guarantees strong stability across different actuation types and energy sources. The second line seeks to design control solutions capable of maintaining strict predefined performance levels even under uncertainties and faults, while balancing the inherent trade-off between system robustness and responsiveness. The third line addresses the interpretability and trustworthiness of black-box, learning-based strategies, traditionally difficult to analyze and verify, toward enabling their stable integration into HDMMs in alignment with international safety standards.

The validity and generality of the proposed framework are demonstrated through three distinct case studies, involving different actuation mechanisms and operational conditions, and covering both heavy-duty mobile robotic systems and robotic manipulator systems. Collectively, the findings of this dissertation are documented in five peer-reviewed publications and one unpublished manuscript. This work advances the state of the art in nonlinear control and robotics, laying the foundation for accelerating the two aforementioned transitions in HDMMs.

\tableofcontents

\listoffigures
\listoftables

\glsaddall
\setglossarystyle{taulong}
\setlength{\glsnamewidth}{0.25\textwidth}
\setlength{\glsdescwidth}{0.75\textwidth}
\renewcommand*{\glsgroupskip}{}

\printglossary[title=Abbreviations]

\newpage
% ...

\listofpublications

\chapter*{Author Contribution}

\definecolor{darkgreen}{RGB}{0,150,0}

\section*{Original Publications}
This compendium thesis is built upon the following six publications (\textbf{P-1} to \textbf{P-6}), and one unpublished manuscript (\textbf{P-7}).

\begin{publikeenumerate}
 \item [\textbf{Publication 1}] \underline{\textbf{M. H. Shahna}}, M. Bahari, and J. Mattila, 
    ``Robust decomposed system control for an electromechanical linear actuator mechanism under input constraints,'' 
    \textit{International Journal of Robust and Nonlinear Control}, 2024.

    \item [\textbf{Publication 2}] \underline{\textbf{M. H. Shahna}} and J. Mattila, 
    ``Model-free generic robust control for servo-driven actuation mechanisms with layered insight into energy conversions,'' 
    \textit{American Control Conference (ACC)}, 2025.
    
    \item [\textbf{Publication 3}] \underline{\textbf{M. H. Shahna}} and J. Mattila, 
    ``Exponential auto-tuning fault-tolerant control of n degrees-of-freedom manipulators subject to torque constraints,'' 
    \textit{IEEE Conference on Decision and Control (CDC)}, 2024.
    
    \item [\textbf{Publication 4}] \underline{\textbf{M. H. Shahna}}, M. Bahari, and J. Mattila, 
    ``Robustness-guaranteed observer-based control strategy with modularity for cleantech {EMLA-driven} heavy-duty robotic manipulator,'' 
    \textit{IEEE Transactions on Automation Science and Engineering (T-ASE)}, 2025.
    
    \item [\textbf{Publication 5}] \underline{\textbf{M. H. Shahna}}, P. Mustalahti, and J. Mattila, 
    ``Robust torque-observed control with safe input–output constraints for hydraulic in-wheel drive systems in mobile robots,'' 
    \textit{Control Engineering Practice (CEP)}, 2025.
    
    \item [\textbf{Publication 6}] \underline{\textbf{M. H. Shahna}}, J. P. Humaloja, and J. Mattila, 
    ``Model reference-based control with guaranteed predefined performance for uncertain strict-feedback systems,'' 
    \textit{Control Engineering Practice (CEP)}, 2025.

\end{publikeenumerate}

\begin{figure}[h] 
  \centering
\scalebox{1}
    {\includegraphics[trim={0cm 0.0cm 0.0cm
    0cm},clip,width=\columnwidth]{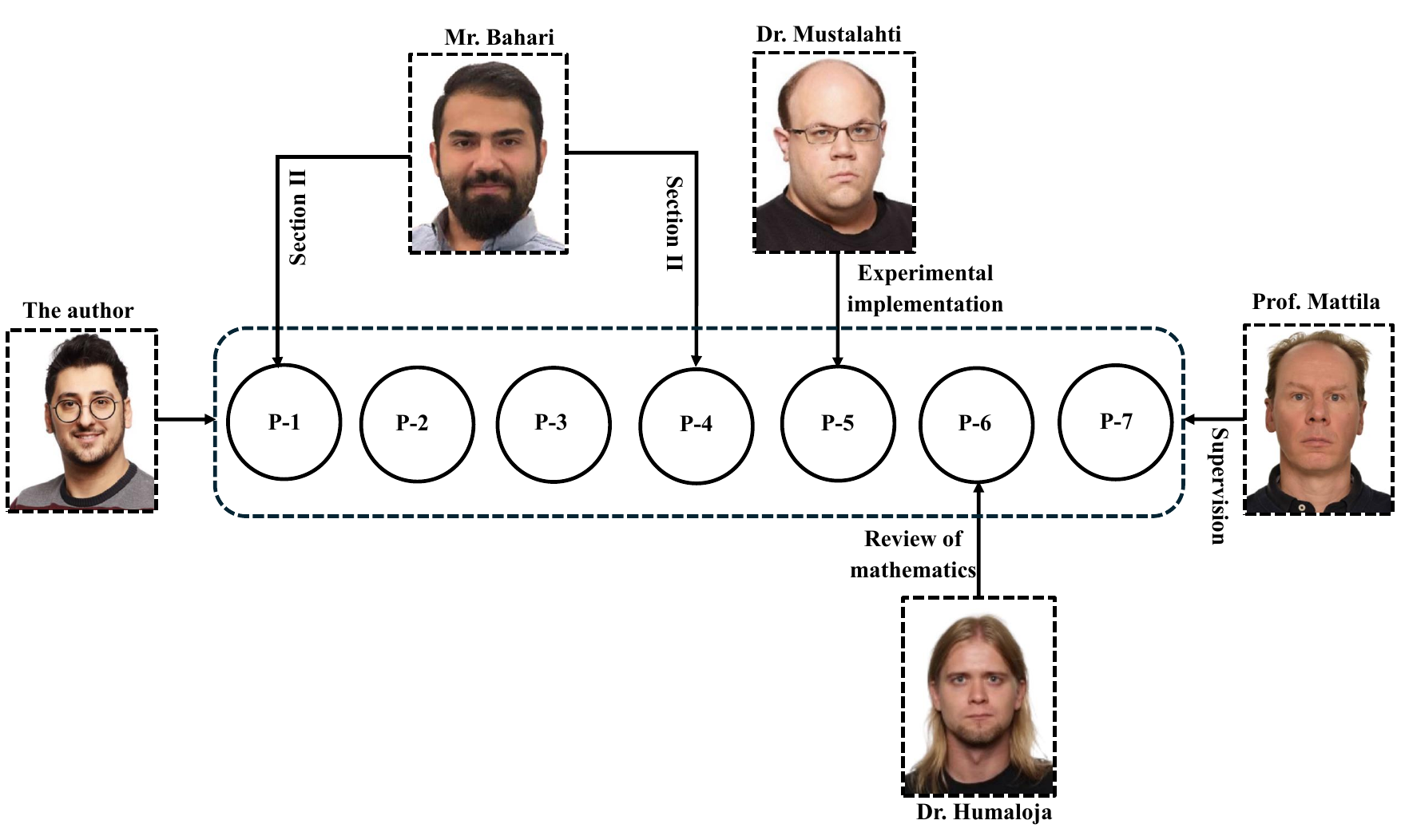}}
  \caption{Illustration of the contribution shares across the main publications of the thesis.}
  \label{P-1_schematicassa}
\end{figure}

In \textbf{P-1} to \textbf{P-6}, the candidate was responsible for the full manuscript writing, theoretical development, controller design, and experimental validation. Prof. Mattila provided supervision through conceptual guidance, critical review of the manuscripts, and constructive feedback. Mr. Bahari contributed to Section II of \textbf{P-1} and \textbf{P-4}, Dr. Mustalahti supported the experimental work in \textbf{P-5}, and Dr. Humaloja reviewed the mathematical formulations in \textbf{P-6}; see Fig. \ref{P-1_schematicassa}.

\section*{Unpublished Manuscript}

Alongside publications \textbf{P-1} to \textbf{P-6}, this thesis further includes the following unpublished manuscript, \textbf{P-7}. 

\begin{publikeenumerate}
    \item[\textbf{P-7}] \underline{\textbf{M. H. Shahna}} and J. Mattila, 
``Synthesis of deep neural networks with safe robust adaptive control for reliable operation of wheeled mobile robots.''
\end{publikeenumerate}

In \textbf{P-7}, the candidate was responsible for the full manuscript writing, theoretical development, controller design, and experimental validation. Prof. Mattila provided supervision through conceptual guidance, critical review of the manuscripts, and constructive feedback.

\section*{Supporting and Related Publications}
During the doctoral studies, the candidate also contributed to eight other publications (\textbf{P-8} through \textbf{P-14}). While these works support and complement the main research narrative, they are not discussed in detail in this thesis. The related publications are listed below:

\begin{publikeenumerate}
    \item [\textbf{Publication 8}] \underline{\textbf{M. H. Shahna}}, S. A. A. Kolagar, and J. Mattila, 
    ``Integrating DeepRL with robust low-level control in robotic manipulators for non-repetitive reaching tasks,'' 
    \textit{IEEE International Conference on Mechatronics and Automation (ICMA)}, 2024.

    \item [\textbf{Publication 9}] \underline{\textbf{M. H. Shahna}}, P. Mustalahti, and J. Mattila, 
    ``Robust model-free control framework with safety constraints for a fully electric linear actuator system,'' 
    \textit{IEEE International Power Electronics and Motion Control Conference (PEMC)}, 2024.

    \item [\textbf{Publication 10}] \underline{\textbf{M. H. Shahna}}, E. Haaparanta, P. Mustalahti, and J. Mattila, 
    ``LiDAR-inertial SLAM-based navigation and safety-oriented AI-driven control system for skid-steer robots,'' 
    \textit{IEEE Conference on Decision and Control (CDC)}, 2025.

    \item [\textbf{Publication 11}] \underline{\textbf{M. H. Shahna}}, P. Mustalahti, and J. Mattila, 
    ``Anti-slip AI-driven model-free control with global exponential stability in skid-steering robots,'' 
    \textit{IEEE/RSJ International Conference on Intelligent Robots and Systems (IROS)}, 2025.

    \item [\textbf{Publication 12}] \underline{\textbf{M. H. Shahna}}, P. Mustalahti, and J. Mattila, 
    ``Fault-tolerant control for system availability and continuous operation in heavy-duty wheeled mobile robots,'' 
    \textit{IEEE/ASME International Conference on Advanced Intelligent Mechatronics (AIM)}, 2025.

    \item [\textbf{Publication 13}] S. A. A. Kolagar, \underline{\textbf{M. H. Shahna}}, and J. Mattila, 
    ``Combining deep reinforcement learning with a jerk-bounded trajectory generator for kinematically constrained motion planning,'' 
    \textit{European Control Conference (ECC)}, 2025.

    \item [\textbf{Publication 14}] M. Bahari, A. Paz, \underline{\textbf{M. H. Shahna}}, P. Mustalahti, and J. Mattila, 
    ``System-level efficient performance of EMLA-driven heavy-duty manipulators via bilevel optimization framework with a leader-follower scenario,'' 
    \textit{IEEE Transactions on Automation Science and Engineering (T-ASE)}, 2025.
\end{publikeenumerate}


\mainmatter


\chapter{Introduction}
\label{ch:Introduction}

\begin{figure}[h] 
  \centering
\scalebox{1}
    {\includegraphics[trim={0cm 0.0cm 0.0cm
    0cm},clip,width=\columnwidth]{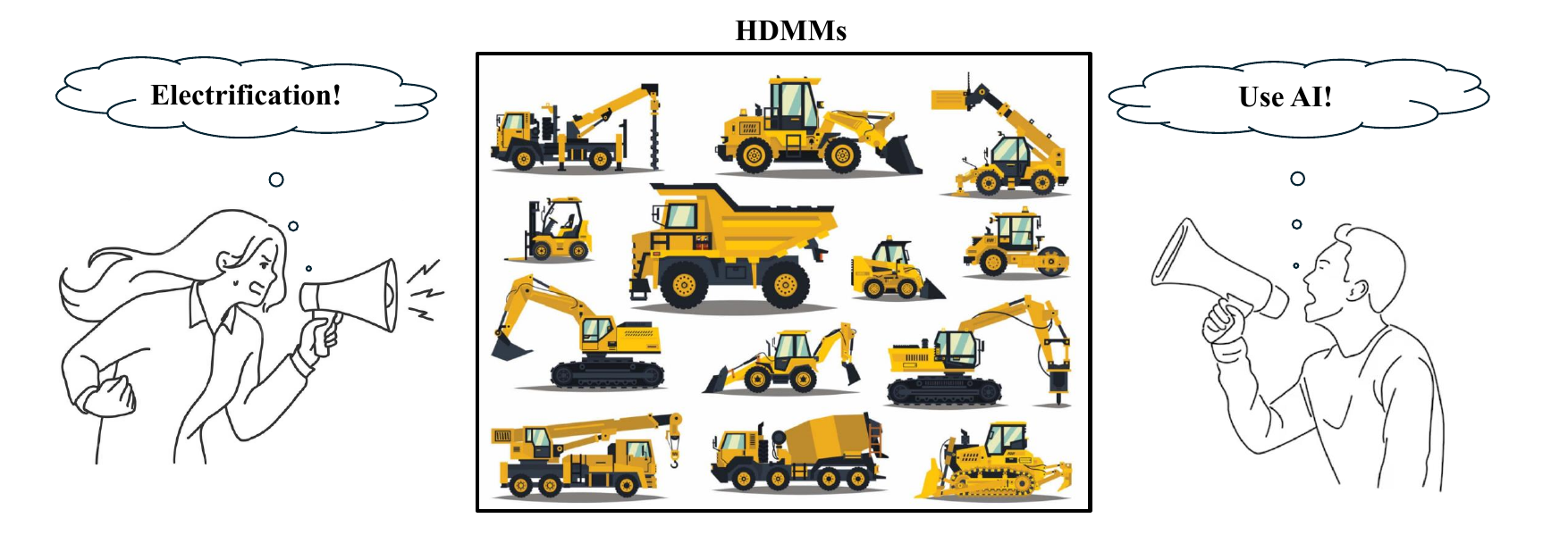}}
  \caption{\textcolor{black}{Inevitable revolution in HDMMs.}}
  \label{asfafadf33}
\end{figure}

Today’s heavy-duty mobile machinery (HDMMs) faces two significant transitions on the path toward adopting cutting-edge technologies, as shown in Fig. \ref{asfafadf33}.

First is the urgent global imperative to combat climate change. The recognized impact of greenhouse gas emissions has led to binding international frameworks, most notably the 2015 Paris Agreement, aimed at significantly reducing carbon dioxide output across sectors. This has created immense pressure for traditionally fossil-fuel-reliant industries to transition toward sustainable and clean energy solutions \cite{agreement2015united}. For decades, diesel-powered hydraulic systems have served as the dominant actuation mechanism in heavy-duty industries. Consequently, transitioning to a fully electric system, whether through direct replacement or system-level redesign, presents substantial technological and economic challenges.
Second is the rapid evolution of artificial intelligence (AI), which is reshaping heavy-duty industries. Machines have evolved from early human-oriented devices to 20th-century industrial robots, and today are transforming heavy-duty machines towards full autonomous HDMMs. They operate in unstructured, high-risk environments, such as tunnels, mines, and construction sites, where hazards and inaccessibility demand operational reliability and safety \cite{herman1998renaissance, romanenko2022robot, adams2025human, klancar2017wheeled, machado2021autonomous, chi2017avoiding, leung2023automation}. Although AI technologies promise benefits for HDMMs—such as continuous operation in hazardous or repetitive tasks, reduced labor dependency, and minimized human error—their adoption is limited by serious safety and reliability concerns, as failures could endanger assets and undermine long-standing industrial trust and investments \cite{katreddi2022review}. Convincing industry stakeholders requires more than demonstrations of robots performing tasks such as playing ping pong or jumping; it demands robust industrial justification grounded in control theory that ensures HDMMs align with international standards, such as ISO/IEC TR 5469:2024: Functional Safety and AI Systems \cite{iso5469}—and is validated through real-world reliability.
Control theory is essential for HDMMs to align with international standards. 

\section{Control Design Challenges for HDMMs}
This section discusses the specific challenges involved in designing intelligent control systems for HDMMs, building on two general challenges previously identified. To address these effectively, a solid foundation in control theory is essential, particularly in evaluating whether model-free or model-based control strategies are more suitable in such applications. The answer to this question is key to tackling the broader challenges of robotizing HDMMs in a reliable and scalable way.

\subsection{Electrification in HDMMs}
\label{electrififcan}
\begin{figure}[h] 
  \centering
\scalebox{1}
    {\includegraphics[trim={0cm 0.0cm 0.0cm
    0cm},clip,width=\columnwidth]{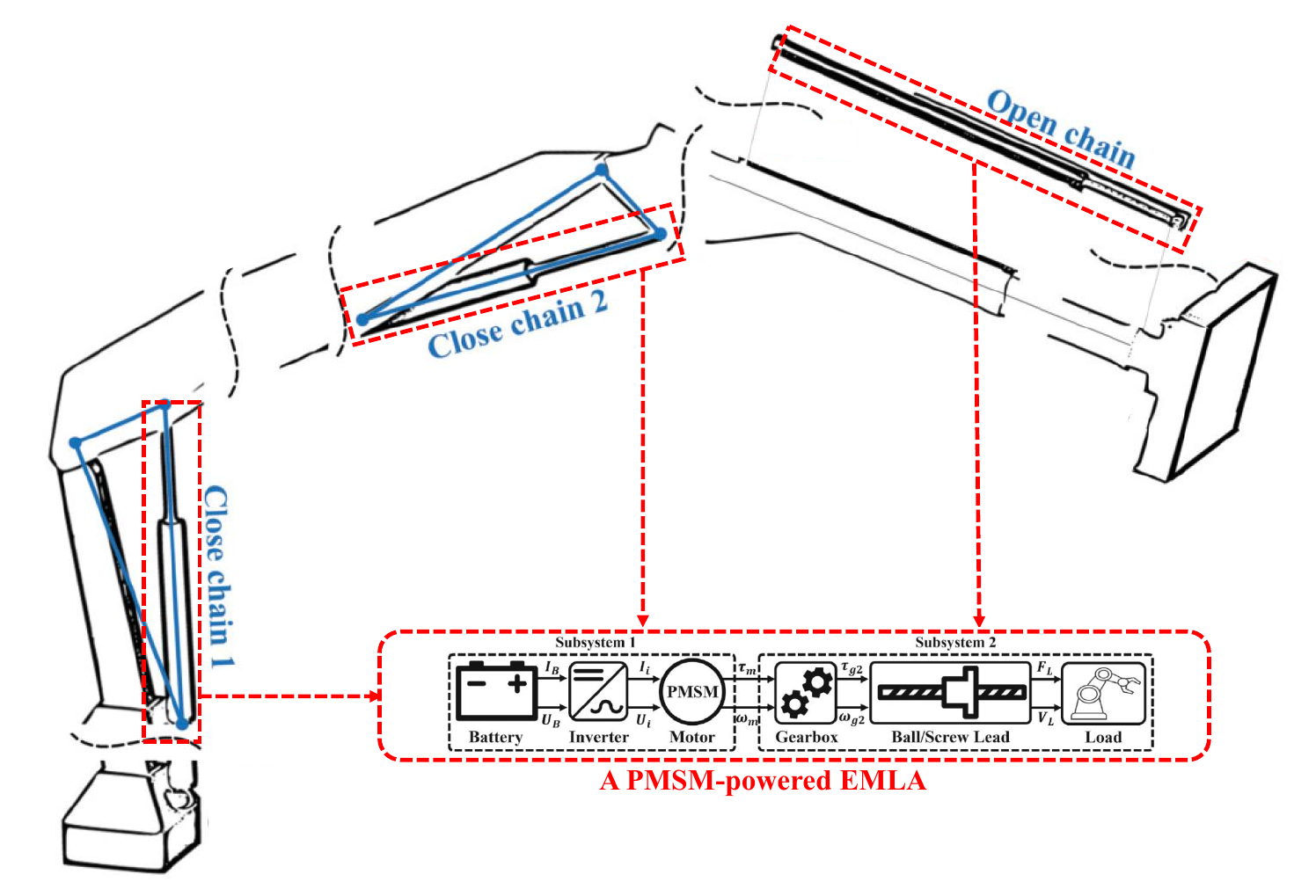}}
  \caption{\textcolor{black}{Fully-electrified manipulator driven by PMSM-powered EMLAs.}}
  \label{asdasaadasdasdassdasd}
\end{figure}

The push for decarbonization, alongside advancements in battery and charging infrastructure, has a profound impact on various sectors, exemplified by the surge in zero-emission battery electric vehicle (BEV) development \cite{bischoff2010strategic, daily2017self, badue2021self}. Extending the concept of BEVs to working machinery, this industry is witnessing the rise of a new class of electric vehicles, known as electrified HDMMs. Heavy-duty electromechanical linear actuators (EMLAs) are emerging as a key enabler of HDMM electrification, offering advantages over traditional electro-hydraulic actuators, including higher energy efficiency, precise motion control, reduced maintenance, and improved safety. EMLAs typically consist of an electric motor, gearbox, screw mechanism, and structural load-bearing components. They efficiently convert electrical energy into linear motion while avoiding common issues associated with hydraulic systems, such as oil leakage and component wear. The integration of sensors and advanced electronic control systems further enhances motion precision. Among available motor types, permanent magnet synchronous motors (PMSMs) are widely favored due to their high efficiency, torque density, and low cogging torque, making EMLAs particularly well-suited for battery-powered mobile machines. Fig. \ref{asdasaadasdasdassdasd} shows a 3-DoF electrified manipulator actuated by different EMLAs.

\begin{figure}[h] 
  \centering
\scalebox{0.85}
    {\includegraphics[trim={0cm 0.0cm 0.0cm
    0cm},clip,width=\columnwidth]{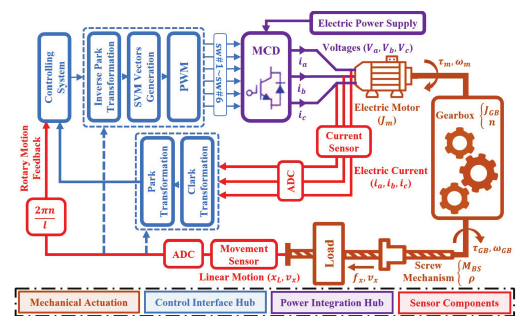}}
  \caption{\textcolor{black}{EMLA schematic \cite{bahari2024system}.}}
  \label{fiasasadfbdfbdfasdg3}
\end{figure}

Incorporating closed-chain structures into the links of robotic manipulators enhances their load-bearing capacity compared to traditional open-chain configurations, due to improved structural rigidity and more efficient force distribution across multiple kinematic loops. However, their implementation introduces several control design challenges. Note that the main scope of this thesis is to design control strategies for electrified HDMMs, aligning with current industrial trends and addressing their specific challenges in depth. To ensure the generality of the intended framework for future alternatives, the design approach also aims to accommodate hydraulic and hybrid HDMMs, which are included in the validation studies. Special challenges in designing control for electrified HDMMs are summarized as follows:

1) Complex actuation mechanisms: actuation systems convert input energy—from electrical servomechanisms—into controlled mechanical motion, enabling precise movements for complex tasks. From a control design perspective, the high-order ordinary differential equations (ODEs) of such systems can be generally decomposed into two sections: the motion dynamics which manage mechanical dynamics and compute the required forces or torques, while the electrical servomechanism governs energy conversion, translating electrical input (e.g., motor voltage) into actuator motion under load via the motor–gearbox assembly. The gearbox, as part of the drivetrain, transmits torque and modulates speed, forming a critical link between electrical energy and mechanical output. When these energy conversion dynamics are fully accounted for, the actuator behaves as a fourth-order system due to the tight coupling between power electronics, drivetrain, and mechanical motion; as shown in Fig, \ref{fiasasadfbdfbdfasdg3}. This complexity, along with the nonlinear and time-varying nature of HDMMs, makes robust model-based control design highly challenging. Ensuring stability and safety under varying loads, disturbances, and actuator uncertainties adds further difficulty, particularly in such applications.

2)  Complex robotic interactions: In large-scale multi-degree-of-freedom (DoF) manipulators, each joint is actuated by a heavy-duty PMSM-driven EMLA to perform demanding robotic tasks. This configuration results in tightly coupled multi-EMLA dynamics, where even small variations in the behavior of a single actuator can impact the performance of the entire manipulator. Consequently, achieving stable and high-performance operation in electrified HDMMs requires careful consideration of the interactions between EMLAs.

3) Intense uncertainties and external disturbances: heavy-duty EMLAs suffer inherent complexities due to motor torque disturbances (induced by magnetic flux harmonics), voltage perturbations (from inverter switching and supply inconsistencies), and load effects. They directly degrade the performance of EMLAs through the actuation chain, ultimately compromising the stability and precision of the overall HDMM system. 

Therefore, replacing mature hydraulic-actuated HDMM control technologies requires significant time and research. In addition, a control system designed solely for meeting current challenges in upcoming electrified HDMMs risks becoming obsolete if future energy architectures change. Thus, developing a generic and adaptable system ensures long-term resilience and compatibility with future technologies.

\subsection{Model-Free or Model-Based Control?}
\label{modelfrrebased}
Before designing a control strategy for HDMMs, it is essential to pause and critically reflect on what the control system is truly expected to achieve. Control design is not merely about applying standard methodologies—it involves making deliberate decisions based on system goals, environmental conditions, and application constraints. 

\textit{``A fundamental and well-recognized trade-off in control theory lies between system robustness and responsiveness \cite{jin2003trade, kool2016does}.''} 

\begin{figure}[h] 
  \centering
\scalebox{0.75}
    {\includegraphics[trim={0cm 0.0cm 0.0cm
    0cm},clip,width=\columnwidth]{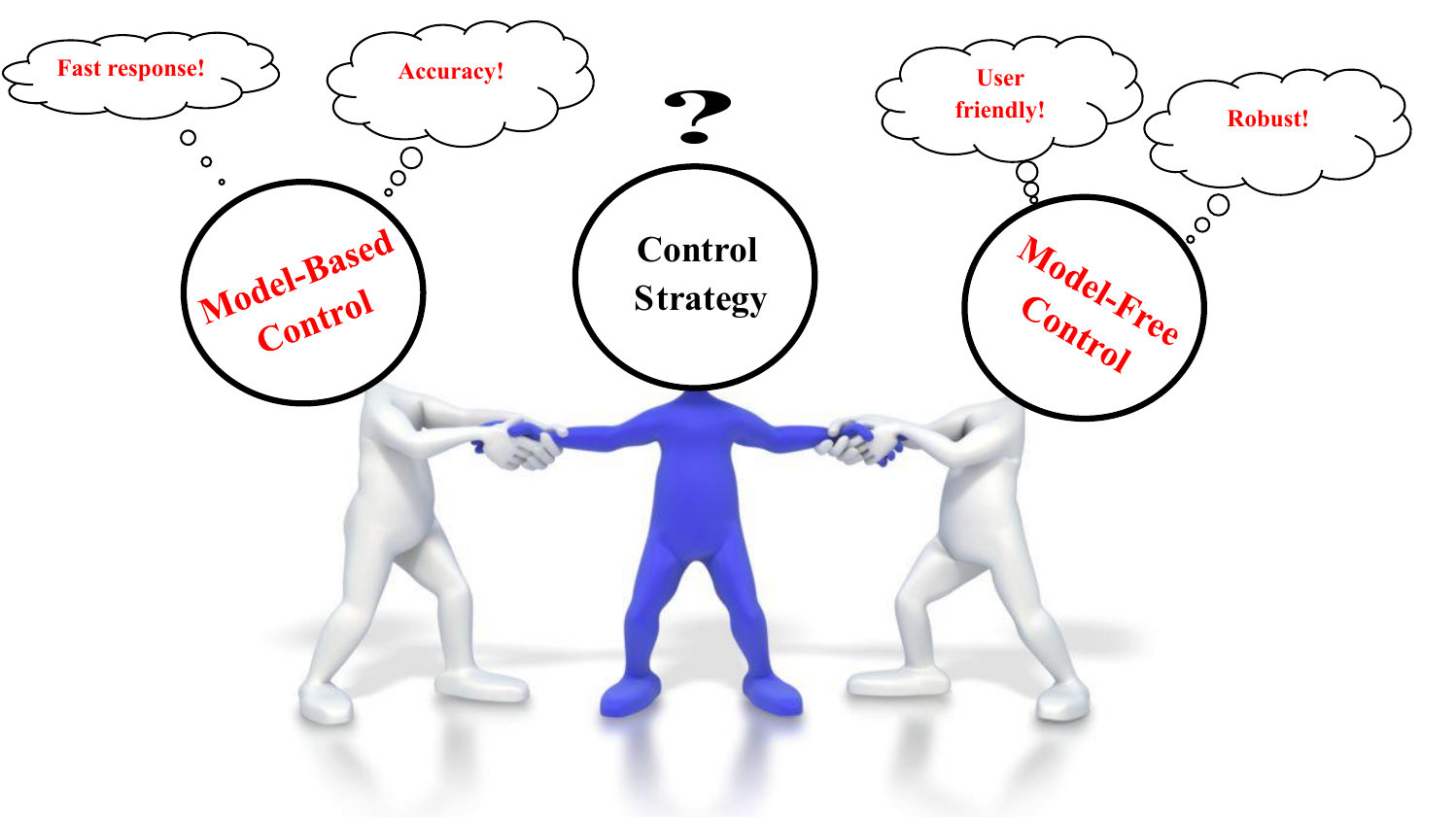}}
  \caption{\textcolor{black}{Trade-off in System Responsiveness and Robustness.}}
  \label{fiasadasdg3}
\end{figure}

The quote clearly expresses that enhancing robustness typically involves designing the system to tolerate a wide range of uncertainties and disturbances; however, this often comes at the expense of system responsiveness, manifesting as slower responses, increased overshoot, or degraded steady-state accuracy; see Fig. \ref{fiasadasdg3}.

Model-based strategies estimate action values by planning ahead using a model of how the system works \cite{kool2016does}. Model-based control strategies, such as optimal or nonlinear controllers, can offer excellent responsiveness and precision when the system model is accurate \cite{brosilow2002techniques}. However, developing accurate models of highly nonlinear HDMMs is a significant challenge. The intricate interactions among mechanical, electrical, and environmental dynamics make the construction and maintenance of such models time-consuming and prone to unavoidable inaccuracies. 

In contrast, model-free control strategies take a generic approach that eliminates the need for complex mathematical modeling \cite{hou2013model}. These methods encompass a wide range of controllers, including intelligent PID controllers \cite{fliess2009model}, sliding mode controllers \cite{ebrahimi2018model}, iterative feedback tuning \cite{baciu2021iterative}, model-free adaptive controllers \cite{hou2013model}, as well as hybrid approaches such as adaptive fuzzy controllers \cite{jalali2013model} and virtual reference feedback tuning \cite{roman2016data}. Additionally, cooperative and hybrid model-free techniques are also employed, offering flexible solutions for a wide range of control applications \cite{precup2021data}. These approaches often enhance the system’s robustness, helping to compensate for uncertainties and external disturbances. 
However, this robustness often comes at the cost of reduced system responsiveness and increased computational burden. Although model-free strategies require minimal modeling effort, they often lack accuracy and demand extensive computational tuning and parameter adjustment \cite{kool2016does}.
While it is often believed that control engineers choose between model-free and model-based strategies based on a trade-off between accuracy and effort, we show that it is possible to leverage both strategies to achieve better performance in less time, effectively managing the trade-off between system robustness and responsiveness.

\subsection{AI Technologies in HDMMs}
\label{abLRdeep}
In recent years, deep learning methods have demonstrated remarkable potential in control and autonomous tasks. This class of control policy can be particularly valuable for HDMMs, which often feature complex mechanical structures and operate under highly variable and demanding conditions. Deep learning-based control has the potential to manage complex nonlinearities and uncertainties more effectively, potentially achieving superior performance compared to conventional control strategies. However, the application of such black-box models to safety-critical systems like HDMMs remains deeply problematic, as their lack of interpretability and formal guarantees makes it difficult to ensure system stability and operational safety. Unlike human-robot interaction (HRI) robotics or small-scale autonomous platforms, especially in controlled research environments, where some degree of experimental risk can be tolerated, there is no room for failures in HDMMs. This is because HDMMs pose significant hazards to human operators, infrastructure, and themselves if control strategies fail or behave unexpectedly. This creates a major barrier to certification under strict international safety standards (e.g., ISO 13849, ISO 26262, and ISO/IEC TR 5469), which require rigorous documentation, failure mode analysis, and verified deterministic behavior under fault conditions.

\section{Motivation and Research Questions}

As discussed in Section~\ref{electrififcan} and \ref{abLRdeep}, the industrial trends toward the electrification of HDMMs and AI technologies offer clear benefits in efficiency, sustainability, and greater autonomy. Rather than simply following the electrification trend, this thesis re-examines the core control problem from a broader perspective, guided by Eisenhower’s insight:

\textit{``plans are nothing; planning is everything.''— Dwight D. Eisenhower}.

While the current focus is on electrified HDMMs, our aim is to develop a generic control strategy that is independent of the underlying actuator energy conversion method, be it hydraulic, electric, hybrid, or any future alternatives. This approach seeks to establish a robust, adaptable foundation for control design that can endure future technological transitions with the main focus on the control design of the emerging EMLAs. Achieving this requires a staged process: understanding conventional hydraulic actuation dynamics, analyzing emerging EMLA systems, and ultimately formulating a unified system model that accommodates multiple energy domains. The complexity lies not only in capturing the unique uncertainties of each actuation type, but also in creating a control framework that remains valid across fundamentally different technologies.

\begin{figure}[h] 
  \centering
\scalebox{0.75}
    {\includegraphics[trim={0cm 0.0cm 0.0cm
    0cm},clip,width=\columnwidth]{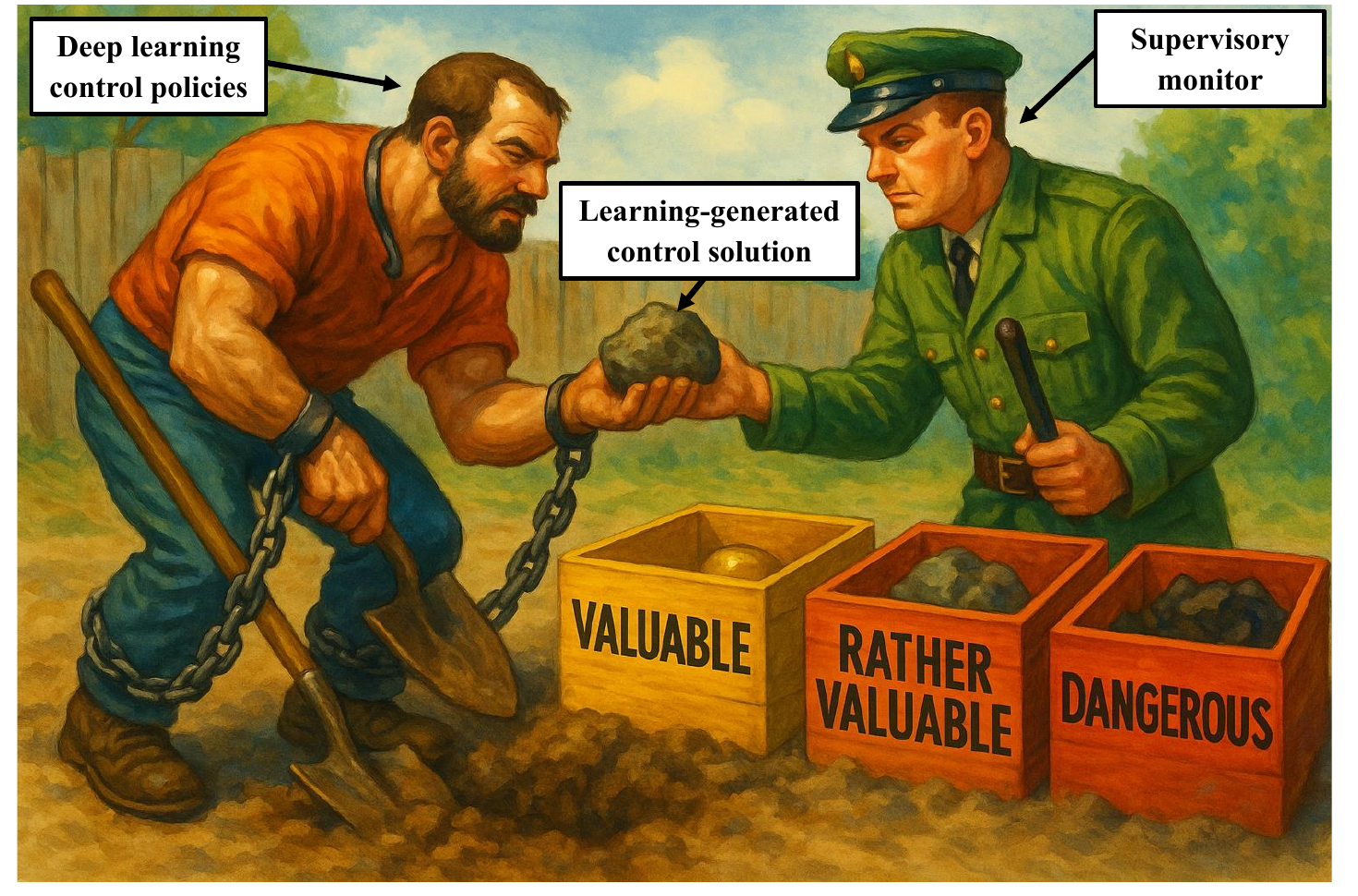}}
  \caption{\textcolor{black}{Motivation for learning-generated control solutions (generated photo by ChatGPT).}}
  \label{figadasfai3}
\end{figure}

Furthermore, rather than attempting to interpret or open AI technologies, we accept the inherent non-transparency of deep learning-based control policies and motivated by the principles outlined in ISO/IEC TR 5469 — Functional Safety and AI Systems \cite{iso5469}, we aim to develop a novel control architecture that can intelligently make decisions on whether to accept, modify, or override learning-generated control actions in compliance with defined safety metrics, thereby ensuring that the overall system remains within safe and stable operational bounds; see Fig. \ref{figadasfai3}.

In addition, as discussed in \ref{modelfrrebased}, control engineering continues to face a polarized debate between model-based and model-free approaches.

\textit{``When the elephants fight, it is the grass that suffers. Be the one who makes them walk together.''— Adapted from an African proverb.} — Adapted from an African proverb

Rather than taking sides, this thesis aims to bridge the gap by integrating both paradigms to balance system responsiveness and robustness—an essential requirement for HDMMs operating under extreme dynamic disturbances.

They lead us to the following research question (RQ):

\begin{publikeenumerate}
    \item [\textbf{RQ1}] How can we achieve a robust control strategy that is \textbf{generic and applicable to any high-order actuation mechanism}, regardless of the type of energy conversion, be it hydromechanical, electromechanical, hybrid, or any emerging future alternative?

    \item [\textbf{RQ2}] How can the control strategy developed in RQ1, originally designed for a single actuation mechanism, be \textbf{extended to multi-body HDMMs}, including both mobile robot platforms and robotic manipulators, that incorporate multiple types of actuation mechanisms, while maintaining its effectiveness and stability in executing robotic tasks without modification?

    \item [\textbf{RQ3}] How can we \textbf{manage the trade-off between system responsiveness and robustness} in a developed robust controller, particularly in HDMMs where intense disturbances amplify the difference between nominal and perturbed conditions?

    \item [\textbf{RQ4}] As HDMMs are inherently prone to faults, degradation, and efficiency losses over time, how can we improve system \textbf{reliability, availability, maintainability, and safety (RAMS)}—under various undesirable conditions in HDMMs?

    \item [\textbf{RQ5}] How can we can design \textbf{an intelligent supervisory component to guarantee system stability and safety-defined metrics when leveraging learning-based policies}, without requiring interpretability or transparency from these complex and stochastic black-box models?
\end{publikeenumerate}

\section{Thesis Contributions}
The scientific contributions (\textbf{C}) of this thesis are as follows:

\begin{publikeenumerate}
    \item [\textbf{C1}] An experimentally validated, stability-guaranteed robust adaptive control (RAC) framework, comprising both model-based and model-free strategies and capable of automatic parameter tuning, has been developed for multi-body HDMMs to ensure compliance with safety-defined performance metrics for both inputs and outputs, with particular focus on managing the trade-off between system robustness and responsiveness.

    \item [\textbf{C2}] An experimentally validated, stability-guaranteed observer framework is developed to ensure accurate estimation of system states in the high-order actuation systems of multi-body HDMMs within a feedback control loop, even in the presence of external disturbances, unmodeled dynamics, and parametric uncertainties.

    \item [\textbf{C3}] To experimentally improve the performance and availability of multi-body HDMMs, the RAC strategy is incorporated as an alternative control approach to a high-performance deep neural network (DNN) policy within a hierarchical control framework. This framework is supervised by two barrier Lyapunov function (BLF)-based safety layers to ensure system stability and compliance with safety-defined performance metrics: 1) the low-level safety layer switches from the DNN to the RAC strategy during severe disturbances to maintain safe operation, albeit at the cost of reduced accuracy and 2) the high-level safety layer evaluates whether continued operation is safe; if not, it initiates a system shutdown to ensure safety.

\end{publikeenumerate}

\section{Outline of the Thesis}
This thesis is organized into six chapters.

\begin{publikeenumerate}
    \item [\textbf{Chapter 1}] Introduced the research topic by presenting the background and general challenges associated with HDMMs. Key challenges in control system design are then discussed, including those related to electrification, the balance between model-free and model-based control strategies, and the integration of AI technologies. The chapter concludes by formulating \textbf{RQs}, outlining the main \textbf{Contributions}, and presenting the overall structure of the thesis.

    \item [\textbf{Chapter 2}] Provides a comprehensive review of the state of the art in the field. It begins with a discussion on actuation mechanisms used in HDMMs and continues with an analysis of subsystem-based control design. The chapter also addresses the representation of system states, input–output constraints, selection of optimal control parameters, and approaches to stability analysis.

    \item [\textbf{Chapter 3}] Introduces the proposed control policies developed in this work, focusing on the experimental validations. The chapter first presents model-free RAC strategies for single actuators and \(n\)-order systems. It then discusses model-based RAC approaches and their extension to model-based RAC methods with safety-defined constraints for various actuation systems of HDMMs, including hydraulic, electrically powered, and hybrid systems.

    \item [\textbf{Chapter 4}] Provides a detailed discussion of the key findings, highlighting the strengths of the work and addressing the formulated \textbf{RQs}.

    \item [\textbf{Chapter 5}] Summarizes the main conclusions drawn from this work and outlines opportunities for future research.
\end{publikeenumerate}

\begin{figure}[h] 
  \centering
\scalebox{1.05}
    {\includegraphics[trim={0cm 0.0cm 0.0cm
    0cm},clip,width=\columnwidth]{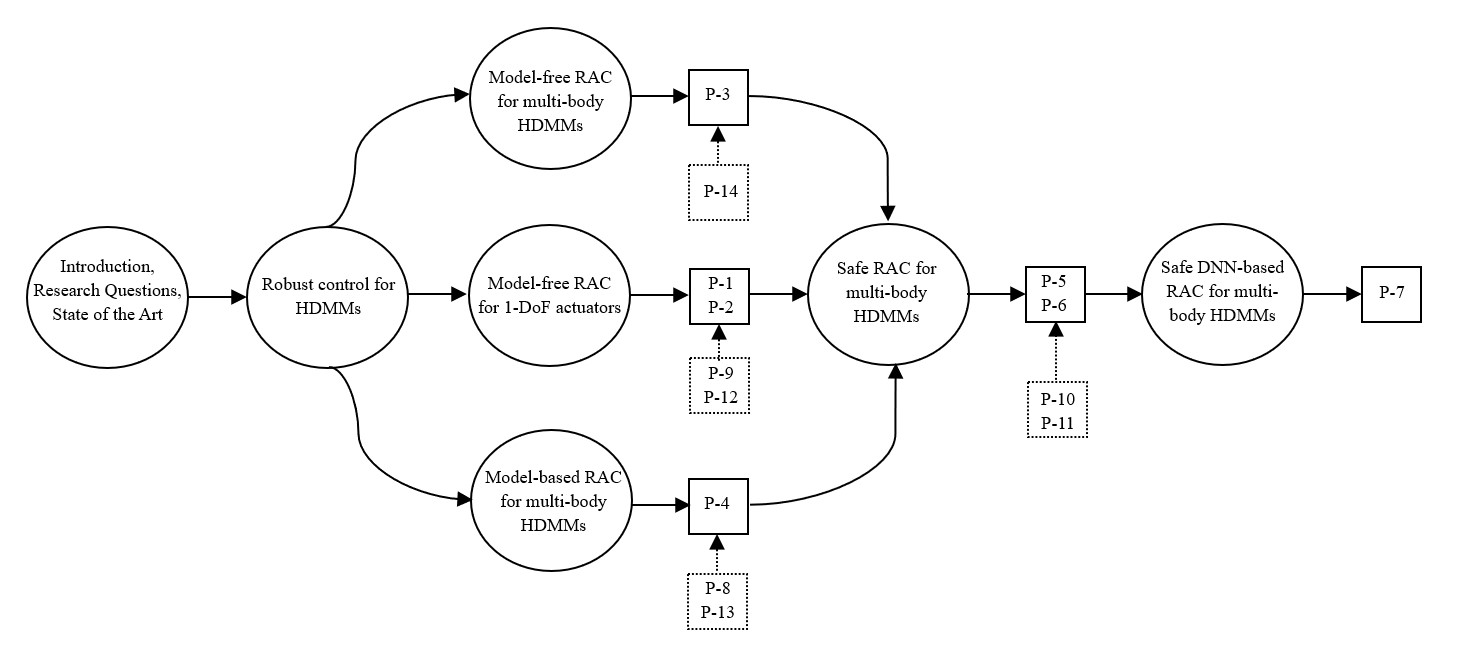}}
  \caption{\textcolor{black}{Outline of the thesis.}}
  \label{figadasfaiadadsadahhhhh3}
\end{figure}

A comprehensive overview of the thesis structure and the connections among the publications is presented in Fig. \ref{figadasfaiadadsadahhhhh3}

\chapter{State-of-the-Art Review}
\label{ch:state_of_the_art}

\section{Heavy-Duty Actuators in SF Form Systems}
\label{Actuation mechanisms in HDMMs}

The strict-feedback (SF) form system structure has drawn considerable attention since the 1990s, particularly in the development of recursive control methodologies \cite{li2022event, wang2015finite}, largely due to its strong capability to model a broad spectrum of real-world systems. These include, for instance, robotic manipulators \cite{doctolero2023neural}, electromechanical actuators \cite{zhang2024adaptive}, DC-DC buck converters \cite{zhang2014robust}, electropneumatic devices \cite{smaoui2006study}, wheeled mobile robots \cite{mathieu2011transformation}, and even neuromuscular stimulation systems \cite{schauer2005online}. Thus, the SF structure is well-suited for representing servomechanism actuators in HDMMs. 

\begin{figure}[h] 
  \centering
\scalebox{1}
    {\includegraphics[trim={0cm 0.0cm 0.0cm
    0cm},clip,width=\columnwidth]{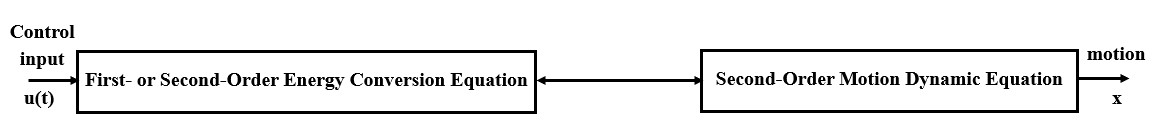}}
  \caption{Servo-drive mechanism decomposition for HDMMs \cite{shahna2024modelacc}.}
  \label{sevgbdzgsfgfj}
\end{figure}

The most widely used servomechanism actuators in industrial applications today are hydraulic systems, which dominate due to their cost-effectiveness and high power-to-weight ratio \cite{yao2017active}; however, there is a growing shift toward emerging EMLAs, despite their higher cost, as industries seek alternatives that offer improved precision, integration, maintenance requirements and energy efficiency \cite{fassbender2024energy}. Ball screws used with EMLAs enable high load capacity by utilizing rolling ball bearings, which reduce friction, distribute forces evenly, and provide high efficiency and stiffness for precise motion in HDMMs \cite{olaru2004new}. Actuators convert energy, whether electrical, hydraulic, or hybrid, into mechanical motion, by enabling precise control. To do so, the control policies are designed for such systems to involve two stages \cite{ren2019adaptive, yang2017position}: the outer loop calculates the required force or torque based on mechanical dynamics to meet position or velocity targets, and the inner loop governs the energy conversion process. This inner stage translates input energy (e.g., oil flow/pressure in hydraulic systems \cite{yao2017active, yang2017position}, or voltage/currents in electrical systems \cite{shahna2024sasPEMCrobust}) into mechanical output. Hence, the state-space models of all actuator systems in HDMMs can be decomposed into one second-order ODE representing the motion dynamics and another second-order ODE capturing the interactive energy conversion processes, such as those involving hydraulic valves and motors, or electric motors, depending on the specific application \cite{shahna2024modelacc}; see Fig. \ref{sevgbdzgsfgfj}. In this formulation, the control input in the actuation mechanism serves as the input to the second-order energy conversion equation, and the system output corresponds to the motion states, such as the position of the actuator.

However, the nonlinear and multi-component nature of these high-order systems makes control challenging. The performance of hydraulic actuators can be influenced by fluctuations in supply pressure, while emerging EMLAs are susceptible to voltage variations \cite{yang2016disturbance, shahna2024robustness}, variability in servo dynamics, sensor inaccuracies, and high load conditions, all of which introduce unmodeled dynamics and disturbances in the SF form. To further clarify this issue, consider a generic SF form system which are cascaded in a triangular structure: each subsystem’s state feeds into the next subsystem as an input, until the last layer where the actual control input appears. This structure essentially radiates out from a core subsystem and allows a recursive stabilization approach. 
This system is expressed as:

\begin{equation}
\label{13}
\begin{aligned}
\dot{x}_1 & =f_1\left(x_1\right)+g_1\left(x_1\right) x_2 \\
\dot{x}_2 & =f_2\left(x_1, x_2\right)+g_2\left(x_1, x_2\right) x_3 \\
& \vdots \\
\dot{x}_{n-1} & =f_{n-1}\left(x_1, \ldots, x_{n-1}\right)+g_{n-1}\left(x_1, \ldots, x_{n-1}\right) x_n \\
\dot{x}_n & =f_n\left(x_1, \ldots, x_n\right)+g_n\left(x_1, \ldots, x_n\right) u
\end{aligned}
\end{equation}

where $x=\left[x_1, \ldots, x_n\right]^{\top} \in \mathbb{R}^n$ denotes the state vector, and $u \in \mathbb{R}$ is the control input. The smooth functions $f_i(\cdot)$ and $g_i(\cdot)$ are assumed to be the modeling term and functional control gain. However, real-world systems are never perfectly known, and their SF models come with various uncertainties, even with a non-triangular form. 
For example, the governing equations of emerging EMLAs at both the electrical and mechanical levels include cross, coupled terms that depend on the entire system state vector \cite{shahna2024sasPEMCrobust}. Consequently, modeling uncertainties in such systems tends to generate non-triangular uncertainties, which violate the structural assumptions required for using the original stability connector.
To manage this complexity, several PMSM control frameworks adopt simplifying assumptions that enable triangular formulations, although such assumptions may not generalize well to all industrial contexts, especially in n-DoF robotic manipulators where multiple EMLAs are supposed to operate in coordination. For instance, studies such as \cite{li2009adaptive} and \cite{zhang2023adaptive} assume the d-axis current component of the PMSM is exactly zero ($i_d=0$), which removes the coupling effects associated with $i_d$ and simplifies the system to a triangular form. However, in practical scenarios where $i_d$ deviates from zero, these coupling effects re-emerge, introducing non-triangular uncertainties that can undermine control performance and robustness.

Overall, uncertainties in HDMMs include parametric uncertainties (unknown or changing parameters in the model) \cite{koivumaki2022subsystem}, unmodeled dynamics (modeling errors or neglected dynamics) \cite{yin2015adaptive}, and disturbances \cite{zerari2021event}. Matched disturbances are uncertainties that enter the system through the same channel as the control input, allowing them to be directly compensated by the controller \cite{fan2023sliding}. In contrast, unmatched disturbances affect the system through different pathways, making them harder to counteract and requiring advanced control strategies like observers or robust adaptive designs \cite{sun2016neural}. Karafyllis, Krstic, and Aslanidis \cite{karafyllis2025deadzone} proposed a triangular SF model for control systems by accounting for both vanishing and non-vanishing disturbances. They argue that this mathematical structure is valid for fully-actuated systems, such as those found in robotics and vehicles; however, as we discussed about the nontriangularity of uncertainties in emerging EMLAs, it does not extend to SF form systems associated with HDMM actuator dynamics. Consequently, a class of SF systems that accounts for unmodeled dynamics, parametric uncertainties, and matched disturbances, more suitable for practical applications, can be expressed as follows \cite{cep2}:

\begin{equation}
\label{1sadasd3}
\begin{aligned}
\dot{x}_1 & =f_1\left(x_1, \textcolor{black}{\ldots, x_n}\right)+\textcolor{purple}{g_1\left(x_1, \ldots, x_n\right)} x_2 +\textcolor{purple}{d_1\left(x_1, \ldots, x_n\right)+\Gamma_1\left(t\right)}\\
\dot{x}_2 & =f_2\left(x_1, x_2, \textcolor{black}{\ldots, x_n}\right)+\textcolor{purple}{g_2\left(x_1, x_2, \ldots, x_n\right)} x_3 +\textcolor{purple}{d_2\left(x_1, \ldots, x_n\right)+\Gamma_2\left(t\right)}\\
& \vdots \\
\dot{x}_{n-1} & =f_{n-1}\left(x_1, \ldots, x_{n-1}, \textcolor{black}{x_n}\right)+\textcolor{purple}{g_{n-1}\left(x_1, \ldots, x_{n-1}, {x_n}\right)} x_{n} + d_{n-1}\left(x_1, \ldots, x_{n_1}, \textcolor{purple}{x_n}\right)+\textcolor{purple}{\Gamma_{n-1}\left(t\right)}\\
\dot{x}_{n} & =f_{n}\left(x_1, \ldots, x_n\right)+\textcolor{purple}{g_{n}\left(x_1, \ldots, x_n\right)} u +\textcolor{purple}{d_{n}\left(x_1, \ldots, x_n\right)+\Gamma_{n}\left(t\right)}\\
\end{aligned}
\end{equation}

where $f_i(.)$ : $\mathbb{R}^n \times \mathbb{R}_{+} \rightarrow \mathbb{R}$ is known modeling terms. The time-varying uncertainty $\Gamma_i: \mathbb{R}_{+} \rightarrow \mathbb{R}$ may represent either vanishing or non-vanishing disturbances with unknown bounds, in contrast to \cite{liu2006global}, which assumes that the upper bound of $\left|\Gamma_i\right|$ is known. The uncertain functions $d_i$ : $\mathbb{R}^n \times \mathbb{R}_{+} \rightarrow \mathbb{R}$ can be non-triangular or triangular unmodeled dynamics (modeling uncertainties) with unknown bounds, in contrast to, \cite{huang2019practical}, which assume the partial knowledge of the $d_i(x_i,t)$ function or its corresponding bounding functions are predefined. $g_i(.): \mathbb{R}^n \times \mathbb{R}_{+} \rightarrow \mathbb{R}$ represents the functional gain for the next state or control input, and it can be an unknown constant (causing parametric uncertainty), an unknown function (causing matched disturbances), or a known modeling parameter or function.  B. Yao in \cite{lai2025indirect} assumed that the functions $g_i$ are given, but their exact values are unknown and may vary over time in HDMMs. This issue can be observed in the numerical example provided in \cite{karafyllis2025deadzone}. In \cite{oliveira2015global}, the upper bound of $g_i$ is assumed to be known. Aside from this structural assumption, the present work does not rely on any further knowledge of the system's nonlinearities.

\textbf{Assumption 1.} \textcolor{black}{In alignment with established adaptive and robust control literature (e.g., \cite{zhang2019low, huang2019practical}), it is assumed that the functional gain $\left|g_i\left(\boldsymbol{x}_i, t\right)\right|$, the modeling uncertainty $\left|d_i\left(\boldsymbol{x}_i, t\right)\right|$, and the external disturbance $\left|\Gamma_i(t)\right|$ are each limited above by unknown values. That is, for all admissible states $\boldsymbol{x}_i$ and for all $t$, these quantities do not grow unbounded.}

\section{Control Design for Uncertain SF Form Systems}
\label{Subsystem-Based Control Design}

Over the past two decades, the backstepping control method has received considerable theoretical attention, particularly for systems that can be modeled in SF form. The method is built on a recursive design framework, where stabilizing controllers and Lyapunov functions are constructed step-by-step from the innermost dynamics toward the actual control input. While the backstepping framework offers a structured approach to stabilization and robustness under uncertainties, its practical applicability remains limited, especially with respect to highly nonlinear systems. A central challenge is the so-called “explosion of complexity,” in which the recursive design process leads to increasingly intricate expressions due to repeated differentiation of intermediate control signals. This issue has been repeatedly reported in \cite{yip1998adaptive, liu2021predefined}. As the system order increases, this results in control laws with deeply nested nonlinear terms, making real-time implementation computationally demanding. Due to this limitation, most practical implementations of backstepping with experimental validation have been limited to relatively low-order systems \cite{yip1998adaptive}. For instance, in \cite{yao2014nonlinear}, a nonlinear adaptive robust backstepping controller was developed for a relatively simple servo valve system with only one DoF. Similarly, in \cite{lai2025indirect}, the authors proposed an indirect adaptive robust backstepping control method for trajectory tracking of a voice coil motor, also a single-DoF actuator, with unknown plant parameters and external disturbances. In \cite{xian2015adaptive}, the authors attempted to design a robust backstepping controller for a six-DoF helicopter system, but their work was limited to simulations and considered only parametric uncertainties, neglecting functional uncertainties arising from modeling errors.

In response to this challenge in higher-order systems, command filters are often used. While command filters help manage the explosion of derivatives in backstepping, they come at a cost: increased computational load and reduced system responsiveness. In this context, \cite{chen2022adaptive} proposes a finite-time command-filtered backstepping control method for an n-order system, addressing only unmodeled dynamics while neglecting matched disturbances and parametric uncertainties. The validation is limited to a second-order numerical simulation. Similarly, \cite{zheng2023practical} proposes a finite-time command-filtered backstepping control approach, with experimental validation on a second-order motor, assuming that the control gain is a parametric uncertainty rather than a time-varying function.

On the other hand, uncertainties in HDMMs are often of large magnitude, and the control input is constrained by actuator limitations, which prevents the use of arbitrarily strong signals to robustify the system. As demonstrated in \cite{karafyllis2025deadzone}, attempts to simplify the problem using dynamic surface control (DSC) have been shown to be not only ineffective but also fundamentally flawed. While DSC mitigates algebraic complexity, it introduces substantial disadvantages. Hence, the authors of \cite{karafyllis2025deadzone} expanded their deadzone-adapted disturbance suppression (DADS) control method from systems with matched uncertainties to a broader class of systems in parametric SF form to effectively prevent both gain escalation and state divergence, independent of the magnitude of disturbances or the extent of parameter uncertainty. However, in contrast to the assumption made in \cite{karafyllis2025deadzone}, it is not practically realistic to assume arbitrarily large uncertainty bounds, because actuation capacities are limited. In real-world systems, excessively large or unbounded uncertainties would demand impractically large, potentially infinite, control efforts (e.g., torques) to preserve robustness and stability, which could damage actuators or activate safety mechanisms that shut the system down. To ensure physical feasibility and actuator protection, HDMM systems must account for both the control input $u$ and the system states remaining within predefined bounded limits when designing control strategies. In addition, the validation of the proposed DADS approach has been limited to simulations on a third-order system, and thus, its reliability in real-world applications remains uncertain.

In addition to backstepping-based control strategies, the virtual decomposition control (VDC) framework \cite{zhu2010virtual}, \cite{zhu1997virtual} has been developed to manage the control of complex robotic systems. A central feature of VDC is its emphasis on modularity, which plays a crucial role in simplifying advanced control system implementations \cite{mastellone2021impact, mattila2017survey}. VDC achieves this by virtually breaking down a robotic system into modular subsystems, typically rigid links and joints, allowing both the control design and stability analysis to be carried out locally at the subsystem level, while still ensuring global asymptotic stability of the overall system.
A distinctive element of VDC is its use of virtual power flows (VPFs) \cite{zhu2010virtual}, which define the dynamic interactions between neighboring subsystems. These VPFs are formulated such that, when subsystems are interconnected, the power exchanges cancel out, maintaining dynamic consistency. However, when VDC principles are extended to non-robotic domains, the interactions among subsystems are no longer governed by VPFs \cite{koivumaki2017adaptive}. To manage the increasing complexity of control design in high-order systems, the work in \cite{koivumaki2022subsystem} extended the subsystem-based approach originally proposed in \cite{zhu2010virtual} and further developed in \cite{koivumaki2017adaptive} for
n-order SF form systems, offering an alternative to conventional backstepping methods. Building upon a generalized representation of subsystems, \cite{koivumaki2017adaptive} introduced a custom stability connector to capture and manage the dynamic interactions between adjacent subsystems. This approach demonstrated that if a particular subsystem includes a stability-preventing connector, the instability can be counteracted by the following subsystem through a stabilizing connector, reaching global asymptotic stability for the whole system. Although the modular control concept introduced in \cite{koivumaki2022subsystem} is both promising and conceptually inspiring, the proposed algorithm is limited to triangular SF systems with only parametric uncertainties while neglecting disturbances and unmodeled dynamics. Furthermore, it assumes that the control gain function is known. Analysis of high-order systems, such as HDMM, is very challenging. The concept of the ``stability connector,'' introduced in \cite{koivumaki2022subsystem}, also offers valuable insights into complex subsystem behavior and facilitates parameter sensitivity analysis. Building on the modularity concept introduced in \cite{koivumaki2022subsystem}, this work extends the approach to a more practically relevant representation of SF systems, as formulated in Eq. \eqref{1sadasd3}. In addition to avoiding the complexity explosion commonly associated with backstepping methods, the proposed approach ensures uniformly exponential stability. Above all, this work aims to go beyond theoretical contributions and low-order experimental validations in state-of-the-art works by focusing on high-order SF form systems, experimentally verified on highly nonlinear HDMMs.

Another major obstacle in control design for high-order systems lies in managing the large number of control gains that must be carefully tuned, as each gain significantly influences both the transient response and steady-state performance of the system, even when applying state-of-the-art control strategies \cite{9161291}. To support this tuning process, population-based optimization algorithms have gained significant traction in recent years due to their effectiveness in handling complex tuning problems \cite{shahna2024integrating}. However, many of these algorithms require careful adjustment of algorithm-specific parameters, which, if not properly selected, can result in high computational demands or convergence to suboptimal local minima \cite{kashani2022population}. Unlike most optimization methods that require parameter tuning, the JAYA algorithm updates solutions without algorithm-specific parameters, using a self-adaptive strategy that moves toward the best candidate and away from the worst \cite{rao2017self}. Hence, this thesis will employ the JAYA optimization algorithms for tuning control parameters.

\section{State Feedback in Uncertain SF Form Systems}
\label{States in HDMM Systems}

Control of multi-body HDMMs requires numerous fault-prone sensors or time derivatives of states. These sensors are also critical for the stabilization of the systems.
Traditional VDC frameworks typically assumed that both position and velocity states of the system are directly measurable for use in control design. While accurate position sensing is generally feasible, measuring rotational velocity is often problematic due to measurement noise. An alternative is to estimate velocity via numerical differentiation of position signals; however, this approach lacks rigorous theoretical support \cite{berghuis2002passivity}. These limitations motivated significant research into controlling robotic manipulators with n DoF without relying on velocity measurements, as seen in works such as \cite{driessen2015observer, malagari2012globally}. Notably, many of these studies omit actuator dynamics in their models. Parallel efforts have addressed similar challenges in systems employing hydraulic actuators, as demonstrated in \cite{bu2000observer} and \cite{sirouspour2002nonlinear}.
Building on these ideas, the controller-observer design in \cite{humaloja2021decentralized} combines insights from the passivity-based control methodology in \cite{berghuis2002passivity} with the subsystem-based VDC framework introduced in \cite{zhu2010virtual}. The authors in \cite{humaloja2021decentralized} developed a control law for an n-DoF open-chain robotic manipulator without access to direct velocity measurements. To overcome this limitation, they implemented a velocity observer that estimates joint velocities using only position measurements and actuator torque inputs. The numerical validation of the high-gain observer proposed in \cite{humaloja2021decentralized} is limited to a second-order system, whereas each actuator in HDMMs exhibits fourth-order SF dynamics. Moreover, as discussed in Section \ref{Actuation mechanisms in HDMMs}, these systems are subject to various uncertainties, highlighting the need for a robust state observer capable of handling such uncertainties in HDMMs. In the second-order motion dynamics domain of each DoF of HDMMs, position sensors are essential components of motion control systems in inverter-driven PMSMs, yet they are highly susceptible to environmental factors such as temperature fluctuations, contamination, external magnetic interference, mechanical shocks, and humidity \cite{bahari2019new}.
A similar challenge arises in the second-order energy conversion system in each DoF of HDMMs, where torque or current/pressure feedback is often required. This information can be obtained either directly through torque or force sensors, or indirectly by inferring actuator behavior from current or pressure measurements \cite{lee2016model}. However, incorporating such sensors increases both hardware and integration costs, while also introducing additional points of failure, particularly in the harsh environments where HDMMs operate under heavy loads \cite{yin2024fault}. To address these issues, \cite{estrada2025hydraulic} proposed a third-order sliding mode controller combined with a finite-time output derivative observer for robust trajectory tracking of hydraulic cylinders. This approach eliminates the need for pressure sensors by relying solely on position measurements, even in the presence of friction, nonlinearities, and disturbances. In a related effort, \cite{estrada2024super} developed an integral sliding surface control strategy that eliminates the need for velocity sensors in noisy hydraulic systems, enabling accurate tracking of sufficiently smooth reference trajectories. These examples underscore the growing importance of sensorless or sensor-minimized control strategies that improve the reliability of sensor-derived information and reduce dependence on vulnerable hardware components in demanding operational conditions. Lacking or inaccurate sensory information of high-order states in the SF representation of HDMMs increases the complexity and difficulty of control. This thesis aims to develop state feedback observers for multi-body HDMMs, ensuring uniform exponential convergence to the true system states even in the presence of uncertainties, as characterized in Eq. \eqref{1sadasd3}.

\section{Input-Output Constraints in Control Design}
\label{Input-Output Constraints}
When input-output behavior is considered in control theory, safety characteristics become a primary concern. While it is theoretically possible to design control strategies for uncertain SF form systems in Eq. \ref{1sadasd3}, especially when idealized assumptions are made, the practical implementation in real-time environments poses significant challenges \cite{tarbouriech2007advanced}. In practical control systems are almost always subject to input and output constraints, which may stem from actuator limitations, performance objectives, or motion safety requirements \cite{shahna2021design, gaagai2023constrained}. For example, actuator limitation is a frequent issue in SF-based systems and can degrade performance or even destabilize the system, as:

\begin{equation}
\label{1sdssd2}
\begin{aligned}
&&&\text{Sat}(u)= \begin{cases}{u}_{max}, & u \geq {u}_{max} \\
0 & {u}_{min} \leq u \leq {u}_{max} \\
{u}_{min} & u \leq {u}_{min}\end{cases}
\end{aligned}
\end{equation}

To counteract this, Min et al. \cite{min2018output} proposed a combined observer and recursive backstepping method to probabilistically ensure global boundedness of the closed-loop response. Despite notable progress, handling multiple layers of uncertainty, including external disturbances, parametric variation, modeling inaccuracies, and unknown input gains, remains a key challenge in SF system control. Recent strategies such as prescribed performance control (PPC) have been introduced to improve both transient and steady-state behavior while reducing the overall computational burden \cite{bikas2018combining, theodorakopoulos2015low}, though they typically rely on a priori knowledge of the control direction. Xie et al. \cite{xie2024low} addressed this by employing smooth orientation functions, although their approach did not consider the critical issue of control input peaking and saturation, which can be detrimental in real-time applications.
Most PPC methods employed BLFs. A limitation of the BLF concept is that it terminates system operation whenever PPC is violated, regardless of whether the condition is truly dangerous or not. However, these approaches tend to be overly conservative because they do not actively steer the system back toward safer conditions once the boundary of the safe region is reached \cite{liu2019barrier}. This thesis also aims to define safety-critical input-output constraints for HDMMs modeled as n-order SF, as described in Eq. \eqref{1sadasd3}, while ensuring PPC. To enhance system availability, this thesis introduces an intelligent supervisory safety component that can automatically manage the trade-off between robustness and system responsiveness.

\chapter{Proposed Control Policies}
\label{ch:controller}

HDMMs are inherently complex systems, consisting of multiple interdependent subsystems operating in coordination. Hence, as discussed in Section \ref{ch:state_of_the_art}, the modularity concept in control theory looks promising for HDMM control system designs, as it allows complex HDMMs to be decomposed into smaller subsystems, making them easier to analyze, develop, and deploy. Although a subsystem-based design simplifies the local understanding of each module, managing the interactions between subsystems remains highly challenging, particularly given the various types of uncertainties HDMMs typically face.
Overall, modularity offers significant advantages: it simplifies system-level analysis, implementation and facilitates future development. In particular, its modular nature allows specific subsystems to be modified, upgraded, or redesigned without the need to rework the entire control system.
In this section, building on the modularity concept proposed in \cite{koivumaki2022subsystem}, we aim to extend it by addressing practical challenges that go beyond theoretical assumptions. Accordingly, the control approach proposed in this thesis is presented in four main stages:

In Section 3.1, a model-free RAC is designed that goes beyond the traditional model-free PID, which is limited to second-order systems \cite{astrom2006advanced}, to address heavy-duty actuators. This design considers the ODEs governing both energy conversion and motion dynamics. The purpose of the model-free RAC design is to experimentally validate its robustness in handling heavy loads in HDMMs. To demonstrate this, an emerging EMLA and a hydraulic in-wheel-drive (IWD) actuator are studied, with the assumption that their models are unknown to the model-free RAC.

In Section 3.2, the robustness- and stability-guaranteed model-free RAC introduced in Section 3.1 is extended to multi-body HDMM systems (both mobile robots and multi-link robots), which exhibit greater complexity, involve numerous interacting uncertainties, and are subject to various fault scenarios.

In Section 3.3, the model-free RAC is extended by introducing a functional term that can be populated with available subsystem-level modeling information of multi-body HDMMs. In this way, the robustness-guaranteed model-free RAC is enhanced in control performance by transitioning into a model-based RAC, tailored for HDMMs operating under non-triangular uncertainties, both parametric and structural. Each additional accurate modeling term naturally improves responsiveness.

Finally, in Section 3.4, safety-defined metrics, developed step by step, are integrated into the model-based RAC. As a result, the performance of the HDMM is significantly enhanced through the incorporation of learning-based policies within the RAC framework.

\section{Model-Free RAC for 1-DoF Actuators of HDMMs}
\label{Model-Free RAC for Single Actuators}

\begin{figure}[h] 
  \centering
\scalebox{1}
    {\includegraphics[trim={0cm 0.0cm 0.0cm
    0cm},clip,width=\columnwidth]{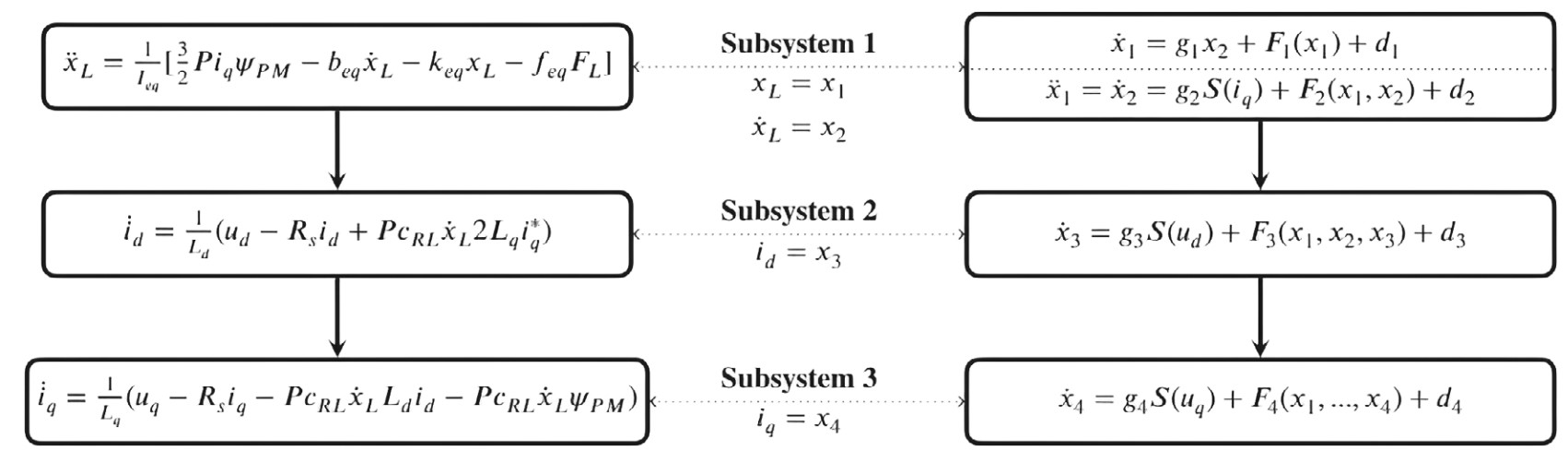}}
  \caption{Decomposing the PMSM-powered EMLA system into a four-order SF form system (\textbf{P-1}).}
  \label{emlasubs}
\end{figure}

Building upon the subsystem-based framework in \cite{koivumaki2022subsystem}, a model-free RAC strategy in \textbf{P-1} was developed to address input constraints, model uncertainties, and external disturbances in a PMSM-driven EMLA—an emerging heavy-duty actuator that is increasingly replacing hydraulic actuators in HDMMs. The study first demonstrated that the state-space representation of the complex multi-stage gearbox-based EMLA, which incorporates the non-idealities of the ball screw mechanism, could be reformulated as a fourth-order uncertain SF system, Eq. \eqref{1sadasd3}, into three subsystems; as shown in Fig. \ref{emlasubs}. Subsystem 1 is based on motion dynamics, while Subsystems 2 and 3 are based on energy conversion (motor current and voltage in d-q frame), modeled using the Park transformation.

\begin{figure}[h] 
  \centering
\scalebox{1}
    {\includegraphics[trim={0cm 0.0cm 0.0cm
    0cm},clip,width=\columnwidth]{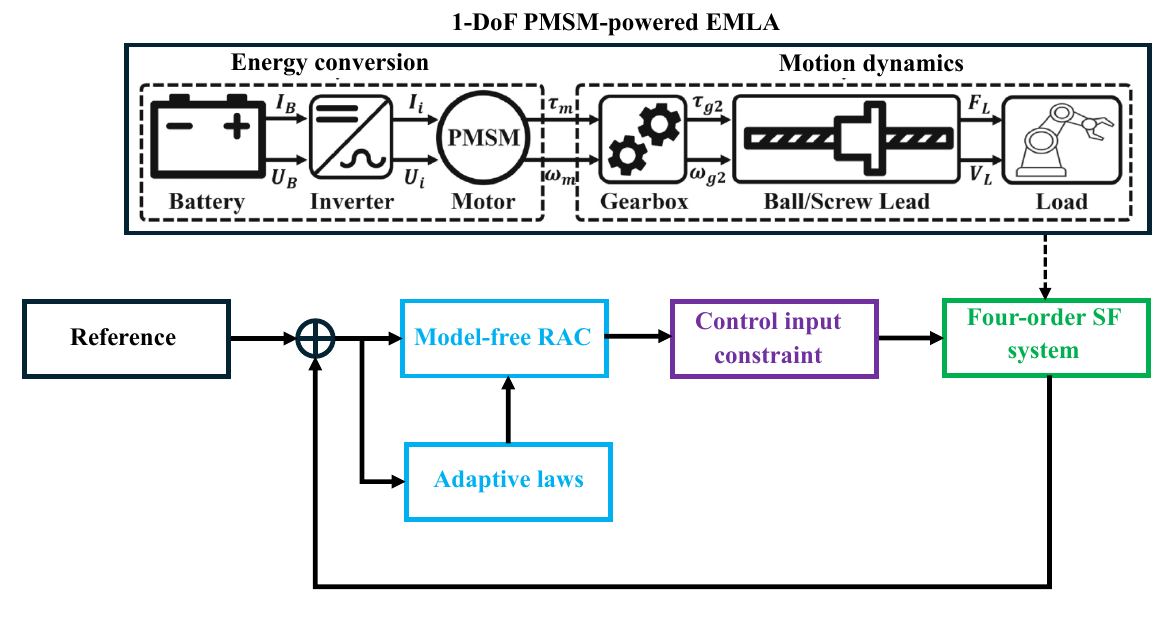}}
  \caption{Model-free RAC for a single PMSM-powered actuator in \textbf{P-1}.}
  \label{P-1_schematica}
\end{figure}

For such a novel SF system of the PMSM-powered EMLA, three adaptive laws were proposed within the RAC framework in \textbf{P-1} to estimate both model uncertainties and matched disturbances (load-induced forces), while ensuring the actuator followed a desired trajectory.
A notable feature of the proposed RAC is its ability to avoid the traditional “explosion of complexity” associated with backstepping methods by treating the time derivative of the virtual control input as part of the system’s uncertainty. In addition, control inputs must remain within the actuator’s nominal limits (e.g., torque capacity) to prevent overheating, mechanical damage, or instability due to actuator saturation or nonlinearity (see Section \ref{Input-Output Constraints}). Exceeding these limits can compromise safety, reduce system reliability, and degrade long-term performance. 
Although the use of the traditional constraint on the system input signal Sat($u(t)$) expressed in Eq. \eqref{1sdssd2} can effectively address this issue, it does so at the expense of system stability. To mathematically represent saturation function and further analysis of $\text{Sat}(u(t))$ into the SF, in \textbf{P-1}, Sat($u$) is defined as a linear function as $\text { Sat }(u(t))=\lambda_1 u(t)+\lambda_2$, where

\begin{equation}
\label{12}
\begin{aligned}
&&&\lambda_1= \begin{cases}\frac{1}{\left|u\right|+1}, & u \geq {u}_{max} \hspace{0.1cm} \text {or } \hspace{0.1cm} u \leq {u}_{min} \\
1 & {u}_{min} \leq u \leq {u}_{max}\end{cases},
&&&\lambda_2= \begin{cases}{u}_{max}-\frac{u}{\left|u\right|+1}, & u \geq {u}_{max} \\
0 & {u}_{min} \leq u \leq {u}_{max} \\
{u}_{min}-\frac{u}{\left|u\right|+1} & u \leq {u}_{min}\end{cases}
\end{aligned}
\end{equation}
$u_{min}$ and $u_{max}$ are defined as the lower and upper bounds, respectively. Thus, this model was incorporated as an additional term to the conventional SF system.
The resulting model-free RAC design guarantees robustness and uniform exponential stability of the closed-loop system. The controller’s performance was evaluated through simulations using two distinct duty cycles, each designed to emulate varying operational demands and load disturbances close to the rated performance limits of the PMSM-powered actuator. The proposed model-free RAC extended the subsystem-based control presented in \cite{koivumaki2022subsystem}, which did not account for modeling uncertainties and disturbances. Moreover, while the approach in \cite{koivumaki2022subsystem} achieved only asymptotic stability, the proposed RAC guarantees uniform exponential stability. Fig. \ref{P-1_schematica} schematically summarizes the development in \textbf{P-1}, with the individual novel design blocks illustrated in different colors.

\subsection{Experimental Validity}

The model-free RAC proposed in \textbf{P-1} is not limited to control of EMLAs alone. In \textbf{P-2}, the control approach was extended to encompass all common actuators used in industry. To achieve this, all actuator models were transformed into third- or fourth-order systems in SF form. Furthermore, the model-free control strategy presented in \textbf{P-2} was validated using two different heavy-duty actuators: an EMLA and a hydraulic IWD actuator, with the assumption that their models are unknown to the model-free RAC.

\subsubsection{Case Study of a PMSM-Powered EMLA}

\begin{figure}[h] 
  \centering
\scalebox{1.3}
    {\includegraphics[trim={0cm 0.0cm 0.0cm
    0cm},clip,width=10.5cm]{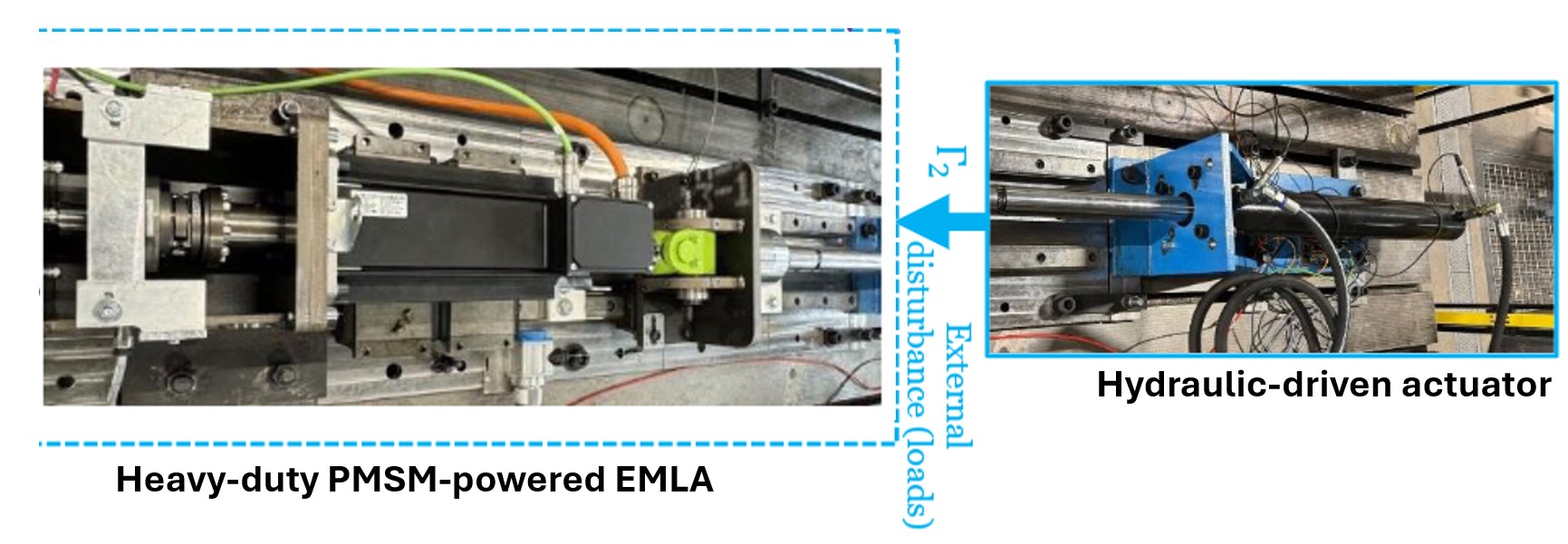}}
  \caption{Heavy-duty PMSM-powered EMLA (\textbf{P-2}).}
  \label{figjkassdlkjldfskasffjksdhn5}
\end{figure}

The first experimental scenario associated with \textbf{P-2} was conducted on a heavy-duty EMLA driven by a PMSM. Under defined technical specifications and safety standards, the EMLA is capable of actuating nominal loads up to 75 kN. The experimental setup used to evaluate the controller is shown in Fig. \ref{figjkassdlkjldfskasffjksdhn5}. To simulate variable external loading conditions, a hydraulic actuator was mechanically coupled to the load side of the EMLA, enabling its use as a load emulator on the actuator under test. The magnitude of the hydraulic load force was regulated via the hydraulic valve system.
As this study assumes that the model of the EMLA system is unknown to the model-free RAC, no further modeling details are provided here.
The controller generates motor torque commands to track a reference position trajectory, which was generated using quintic polynomials, as described in Chapter 13 of \cite{jazar2010theory}. The applied load force—treated as a system disturbance—ranged between 65 and 76 kN. The results demonstrated that the generic model-free RAC proposed in \textbf{P-2} achieved a position tracking error of less than 2 mm under these conditions.

\subsubsection{Hydraulic IWD in Mobile Robotics}

\begin{figure}[h] 
  \centering
\scalebox{1.2}
    {\includegraphics[trim={0cm 0.0cm 0.0cm
    0cm},clip,width=10.5cm]{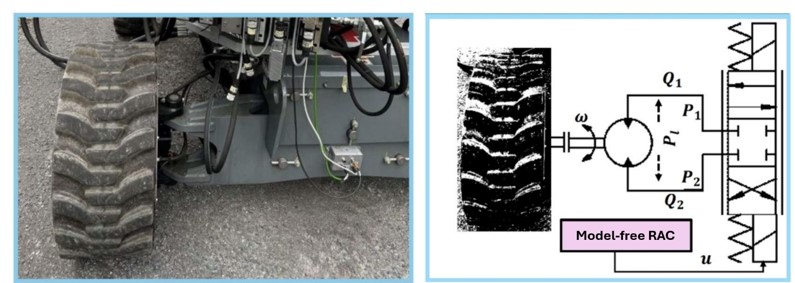}}
  \caption{Hydraulic IWD actuator (\textbf{P-2}).}
  \label{becadadk}
\end{figure}

In the second experimental scenario presented in \textbf{P-2}, the proposed model-free RAC was validated using a heavy-duty hydraulic IWD, under the assumption that the system model was unknown. The experimental setup featured the wheel with a diameter of 0.854 m and a reducer transmission with a gear ratio of 17.7 between the in-wheel motor and the wheel; see Fig. \ref{becadadk}. The model-free RAC generated valve control signals based on feedback from wheel angular velocity measurements, which were obtained using electromagnetic speed sensors integrated into the hydraulic IWD.
As the model-free RAC operates without requiring knowledge of the system dynamics, further system modeling details are omitted here.
The proposed controller demonstrated strong performance in tracking desired velocity references, which were applied to the wheel via a joystick interface. Compared to similar model-free control approaches, model-free adaptive control (MFAC) \cite{wang2022research} and backstepping sliding mode control (BSMC) \cite{truong2022backstepping}, the RAC achieved significantly improved tracking accuracy in this case study.

\textbf{Remark 1:} Both experiments validate that the proposed approach offers a generalizable solution capable of handling arbitrarily bounded load disturbances and input constraints, making it applicable to a wide range of actuator mechanisms. Consequently, the model-free RAC introduced in \textbf{P-1} was further developed in \textbf{P-2} and \textbf{P-9} to control all types of actuation mechanisms commonly used in HDMMs (\textbf{RQ1}). Both publications \textbf{P-1}, \textbf{P-2}, supported by \textbf{P-9}, partially fulfilled \textbf{C1}.

\section{Model-Free RAC for Multi-Body HDMMs}
\label{Model-Free RAC for n-order System}

\begin{figure}[h] 
  \centering
\scalebox{1}
    {\includegraphics[trim={0cm 0.0cm 0.0cm
    0cm},clip,width=\columnwidth]{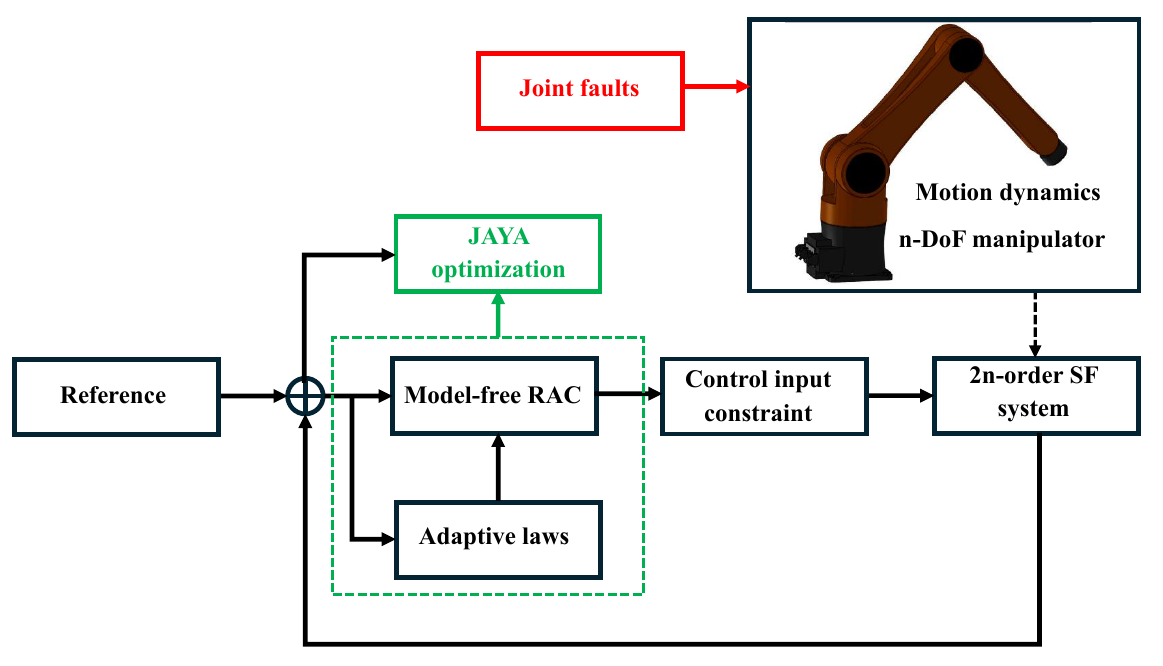}}
  \caption{Model-free RAC for an n-DoF manipulator system in \textbf{P-2}.}
  \label{P-2_schematica}
\end{figure}

As discussed in Section \ref{electrififcan}, multi-body HDMMs exhibit complex robotic interactions. In Section \ref{Actuation mechanisms in HDMMs}, it was shown that an $n$-DoF heavy-duty robotic manipulator can contain up to $4\times n$ subsystems, making the control design highly challenging. Building upon Section \ref{Model-Free RAC for Single Actuators}, \textbf{P-3} introduced a novel model-free RAC approach for an
n-DoF robotic manipulator that focuses exclusively on motion dynamics, specifically, the relationship between joint torques and accelerations, while deliberately ignoring the underlying energy conversion mechanisms responsible for torque generation. This intentional simplification was made to avoid excessive control complexity, particularly since one level of complexity had already been added by extending the model-free RAC from a single-DoF actuator to a multi-DoF manipulator. The energy conversion aspects, which were omitted in this stage, are reintroduced and integrated into the control design for multi-body HDMMs in the following section. As a result, the n-DoF manipulator can be represented as a second-order SF system in Eq. \eqref{1sdssd2}, and the control design focuses solely on the closed-loop torque control for motion tracking, see Fig. \ref{P-2_schematica}.
To further challenge the robustness of the proposed model-free RAC, the joints are subjected to various time-dependent faults that degrade their efficiency. Despite these conditions, the proposed RAC guarantees uniformly exponential stability of the entire robotic system. Due to the large number of subsystems, each with its own control design parameters, \textbf{P-3} also employs a population-based metaheuristic optimization algorithm, the JAYA algorithm, to automatically tune the control gains. Unlike most optimization methods that require parameter tuning, the JAYA algorithm updates solutions without algorithm-specific parameters, using a self-adaptive strategy that moves toward the best candidate and away from the worst \cite{rao2017self}. The optimization objective is defined as the minimization of a cost function representing the system's position and velocity tracking errors. The JAYA algorithm begins by initializing a population of candidate solutions, where each candidate is a vector of control gains. The cost function is evaluated for each candidate, identifying the best candidate (lowest cost) and the worst candidate (highest cost). In each iteration, every candidate is updated using the following rule: the updated candidate is accepted if it yields a lower cost function value; otherwise, the previous candidate is retained. After each iteration, the global best and worst candidates are updated. This iterative process continues until convergence or a predefined iteration limit is reached. The best candidate for termination is selected as the optimal set of control gains.

\subsection{Simulation Validity}

To evaluate the effectiveness of the proposed model-free RAC, fault scenarios were introduced by inducing actuator faults in two joints, occurring 15 seconds after the start of operation. The same fault conditions were applied to two other benchmark fault-tolerant control methods, both of which maintained asymptotic stability while executing the same reference tracking tasks under uncertainty. To ensure a fair comparison, suboptimal control parameters for all three methods were obtained using two metaheuristic optimization techniques: Teaching–Learning (TL) and the JAYA algorithm. Neither optimization method requires algorithm-specific tuning parameters, but JAYA is simpler to implement due to its single-phase structure, unlike TL, which operates in two phases. In addition to the model-free RAC demonstrating uniformly exponential stability, the results also showed that the JAYA-tuned controller achieved superior tracking performance for the tested robotic manipulator.
Hence, \textbf{P-3} extends \textbf{P-1} and \textbf{P-2}, and together they partially meet \textbf{C1}. In addition, \textbf{P-3} addresses \textbf{RQ2} to some extent, which will be fully addressed in \textbf{P-4}. The proposed model-free RAC for the n-DoF manipulator system was later extended to a PMSM-powered, EMLA-driven manipulator in \cite{bahari2024system}, and further integrated with collision avoidance through learning-based methods in \cite{shahna2024integrating} and \cite{kolagar2024combining}.

\section{Model-based RAC for Multi-Body HDMMs}
\label{Model-based RAC for n-order System}

As discussed in Section \ref{modelfrrebased}, when more accurate system models are integrated into control strategies, the system responsiveness, and consequently the overall control performance, tends to improve. This enhancement arises from the well-known trade-off in control theory between robustness and responsiveness. In the model-free RAC design, the controller must simultaneously compensate for system uncertainties and external disturbances through adaptive algorithms, while also addressing internal forces such as friction, mechanical dynamics, and inertia. When a complete or partial functional model of an HDMM is available, the controller can achieve smoother and more precise performance. Conversely, in the absence of accurate modeling, the estimation of large internal forces may lead to high-frequency corrective actions, or “spiking” control signals, which ultimately degrade control performance \cite{sharun2012adaptive}.  

\begin{figure}[h] 
  \centering
\scalebox{0.75}
    {\includegraphics[trim={0cm 0.0cm 0.0cm
    0cm},clip,width=\columnwidth]{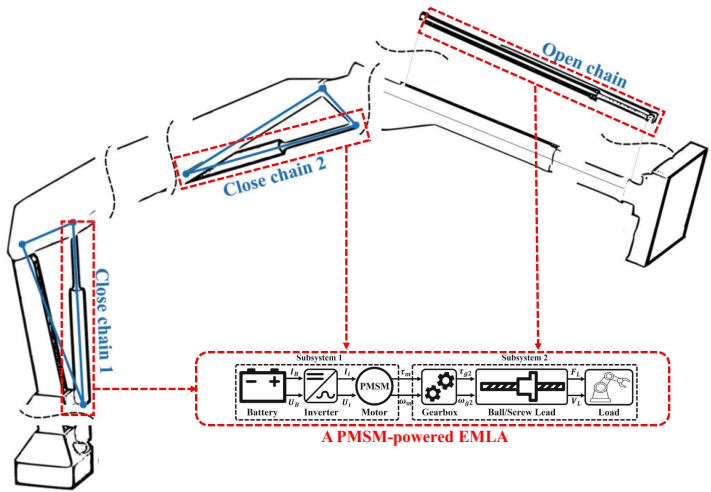}}
  \caption{\textcolor{black}{Fully-electrified HDMM driven by PMSM-powered EMLAs.}}
  \label{asdasasdasd}
\end{figure}

Incorporating closed-chain structures into the links of robotic manipulators enhances their load-bearing capacity compared to traditional open-chain configurations, due to improved structural rigidity and more efficient force distribution across multiple kinematic loops \ref{asdasasdasd}. Building upon Sections \ref{Model-Free RAC for Single Actuators} and \ref{Model-Free RAC for n-order System}, this section advances the approach by incorporating EMLA-driven actuation modeling into the RAC framework for n-DoF robotic manipulators, thereby transitioning to a model-based RAC.
As discussed in Section \ref{electrififcan}, developing a control strategy for an EMLA-driven multi-body HDMM is highly challenging due to the complex and tightly coupled dynamics of its mechanical subsystems. This implies that each joint of the manipulator is driven by a PMSM-powered EMLA, which may differ in model and power rating, to facilitate motion; see Fig. \ref{asdasasdasd}.

The interdependence of multiple components results in integrated dynamics where even small variations often necessitate a complete redesign of the control architecture. Achieving stable, high-performance operation while efficiently managing these interactions—without frequent control system reconfiguration—remains a significant difficulty.

Building on the subsystem modularity concept introduced in \cite{koivumaki2022subsystem}, \textbf{P-4} presents a generalized model-based RAC framework that can be applied uniformly to all PMSM-driven, EMLA-actuated joints of the HDMM. In this extended control approach, a functional term is introduced
that can be populated with available subsystem-level modeling information.
Each additional accurate modeling term naturally improves responsiveness. Hence, joints are modeled as two interconnected subsystems (energy conversion and motion dynamics), overall $4n$-order SF system provided in \eqref{1sdssd2}. This modular representation of PMSM-powered EMLA joints allows the unified controller to remain unchanged even when the system undergoes dynamic modifications, such as replacing motors, altering other dynamic components, or adding and removing joints. Therefore, the focus of this section is to integrate the control strategies developed for single-DoF PMSM-powered EMLAs in \textbf{P-1} and \textbf{P-2} with those for the n-DoF manipulator in \textbf{P-3}, ultimately yielding a developed model-based RAC strategy for a fully electrified n-DoF EMLA-driven manipulator that accounts for both actuation mechanisms and robotic motion. However, as demonstrated in \textbf{P-4} and discussed in Section \ref{Actuation mechanisms in HDMMs}, non-triangular uncertainties are inherent in such applications and pose a significant control challenge.
Moreover, as highlighted in Section \ref{States in HDMM Systems}, state information obtained from sensors in HDMMs is often heavily corrupted by noise, or some system states may even be unavailable.

\begin{figure}[h] 
  \centering
\scalebox{1}
    {\includegraphics[trim={0cm 0.0cm 0.0cm
    0cm},clip,width=\columnwidth]{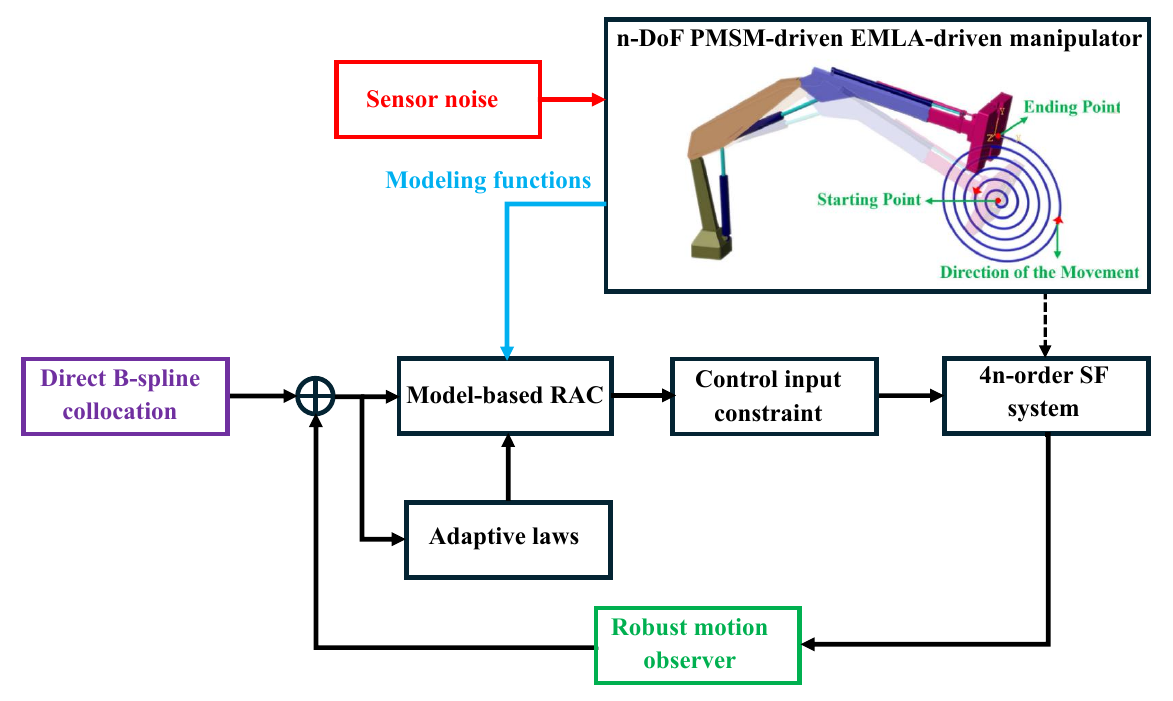}}
  \caption{Model-based RAC for an n-DoF EMLA-driven manipulator in \textbf{P-3}.}
  \label{P-3_schematica}
\end{figure}

To address these challenges, the unified model-based RAC is augmented with two adaptive networks for the fully electrified n-DoF EMLA-driven manipulator in the uncertain SF \eqref{1sdssd2}. The first compensates for load disturbances and non-triangular uncertainties, while the second functions as a motion-state observer, filtering noisy encoder position measurements and estimating velocity states for integration into the control framework. The proposed control framework in \textbf{P-3} is summarized in Fig. \ref{P-3_schematica}. As observed, to define a meaningful control objective at the end-effector level and to analyze the motion trajectories of the EMLAs, we employ a trajectory optimization technique, direct collocation with B-spline curves, which transcribes the problem into a finite-dimensional nonlinear programming problem.
Leveraging the proposed model-based RAC framework, the uniformly exponential stability of the EMLA-driven HDMMs is established, while carefully balancing noise rejection and system responsiveness through appropriate tuning of the observer parameters.

\subsection{Experimental Validity}

\begin{figure}[h] 
  \centering
\scalebox{1.2}
    {\includegraphics[trim={0cm 0.0cm 0.0cm
    0cm},clip,width=10.5cm]{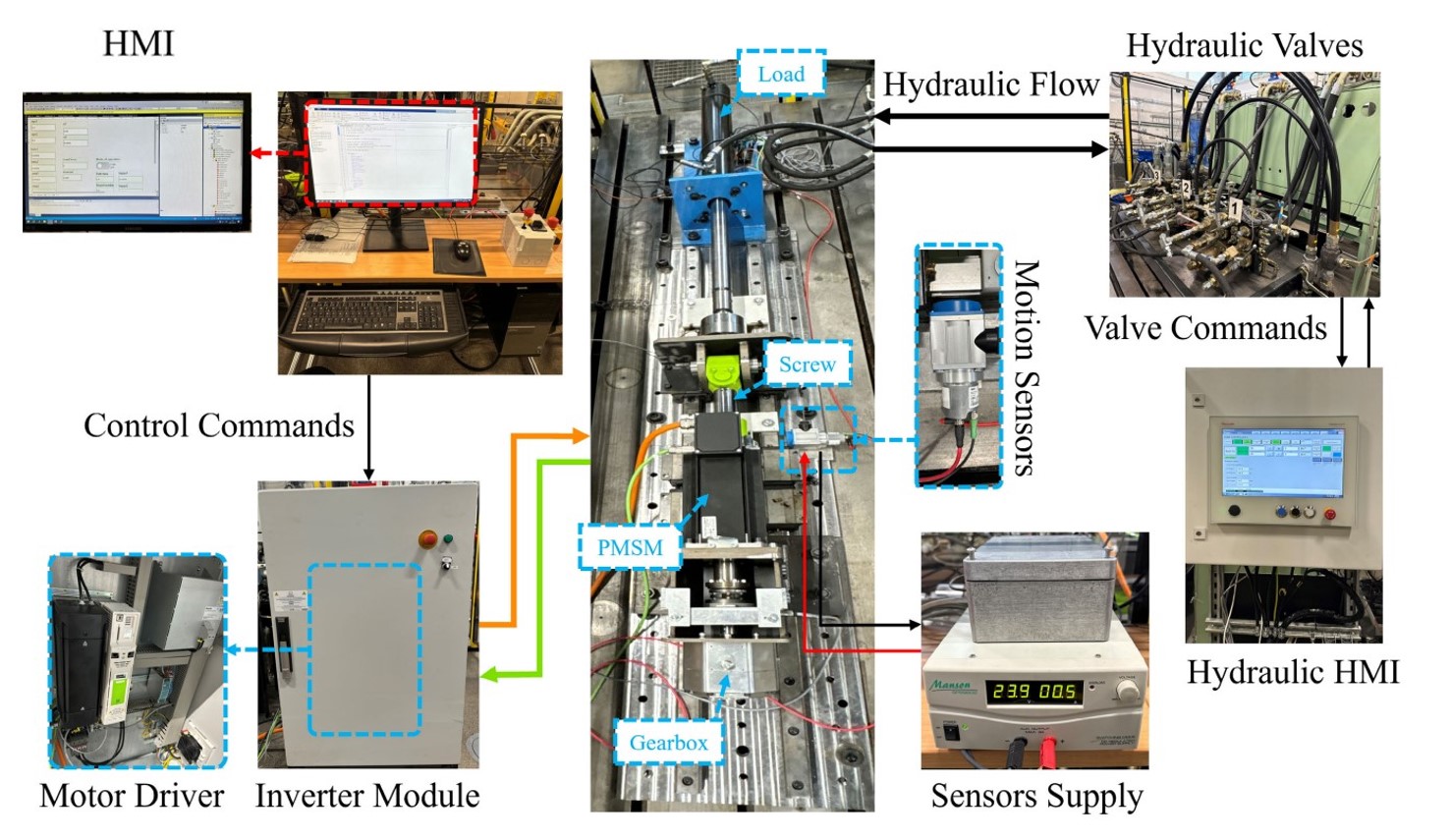}}
  \caption{The EMLA prototype (\textbf{P-4} and \textbf{P-6}). The relevant video is available at: \href{https://youtu.be/udFBJ0T9d8A}{https://youtu.be/udFBJ0T9d8A}.}
  \label{figjkassdlkjlkasffjksdhn5}
\end{figure}

Experiments associated with \textbf{P-4} were performed on a PMSM-driven EMLA, which functions as a prototype for actuating one joint of a forthcoming 3-DoF fully electrified manipulator. In accordance with technical specifications and safety standards, the EMLA is capable of handling and actuating loads up to approximately $75$ kN. The experimental setup used for controller evaluation is depicted in Fig. \ref{figjkassdlkjlkasffjksdhn5}.
To simulate variable external loading conditions, a hydraulic cylinder was mechanically coupled to the load side of the EMLA. This configuration enabled the application of controllable linear forces on the actuator under test. The magnitude of the force generated by the hydraulic cylinder was regulated through an HMI, which controlled the hydraulic valve system.
The proposed control algorithms generated motor torque commands, which were transmitted to a Unidrive M700-064 00350 A controller. This controller, operating with its factory-configured internal control system (which was not accessible or modified), interfaced with the EMLA’s dedicated HMI for command execution and monitoring. The Unidrive then delivered voltage signals to an inverter module that powered the PMSM. The motor used in the EMLA test platform was a three-phase Nidec PMSM rated at 380/480 VAC and 11.6 kW. Detailed specifications of the EMLA components—including the electric motor, gearbox, and ball screw—are summarized in Table \ref{tab:emla_specsads}.

Communication and signal exchange among the controller, inverter module, and other subsystems were implemented via an EtherCAT network, ensuring high-speed, real-time control and monitoring. The control system operated at a sampling rate of 1,000 Hz, and the load force was measured using a 16-bit resolution message protocol. 

Two experimental scenarios using a prototype EMLA powered by a PMSM were conducted to assess the robustness and responsiveness of the proposed model-based RAC in practical settings: (1) operation under upper-moderate velocity with a gradually increasing load from 7 kN to 75 kN, and (2) operation at near-maximum velocity under a constant high load of 75 kN. Comparative evaluations against two state-of-the-art control strategies— command-filter-approximator-based adaptive control (CAC) in \cite{liu2023command} and an adaptive neural asymptotic tracking control (ANATC) in \cite{zhang2023adaptive}—revealed that the proposed model-based RAC outperformed both in all key performance metrics, including tracking accuracy, torque efficiency, and convergence time.
The experimental implementation also highlighted limitations in CAC and ANATC, particularly in position tracking performance. These limitations are expected to become more pronounced in multi-body HDMM systems utilizing multiple PMSM-powered EMLAs, as increasing the number of actuators can compound errors at the manipulator’s end-effector in the task space. Overall, the experimental results confirm the superior robustness and control performance of the model-based RAC framework, underscoring its suitability for high-load, high-performance applications in HDMMs.
Hence, building on \textbf{P-1}, \textbf{P-2}, and \textbf{P-3}, \textbf{P-4} fully addresses \textbf{RQ2} and partially addresses \textbf{RQ3}, as outlined in Section \ref{ch:Introduction}. It also satisfies \textbf{C1} and \textbf{C2} to some extent.

\begin{table}[ht]
\centering
\caption{Specifications of the EMLA system components (\textbf{P-4}).}
\small
\begin{tabular}{lcc}
\hline\hline
\textbf{Parameter} & \textbf{Value} & \textbf{Unit} \\
\hline
PM Magnetic Flux     & 0.134  & Wb \\
Number of Pole Pairs & 4      & - \\
Rated Power          & 11.6   & kW \\
Rated Current        & 23.1   & A \\
Peak Current         & 48.2   & A \\
Rated Torque         & 37     & N·m \\
Peak Torque          & 77     & N·m \\
Rated Speed          & 3000   & rpm \\
Maximum Speed        & 3877   & rpm \\
Phase Resistance     & 0.08   & $\Omega$ \\
Phase Inductance     & 2.42   & mH \\
Gear Ratio           & 7.7    & - \\
Screw Lead           & 16     & mm \\
Screw Diameter       & 63     & mm \\
Screw Lead Angle     & 4.55   & Degree\textdegree \\
\hline\hline
\end{tabular}
\label{tab:emla_specsads}
\end{table}

\section{Safe Model-based RAC for Multi-Body HDMMs}
In this section, we extend the work of the previous section by focusing on input–output control constraints, which are critical as discussed in Section \ref{Input-Output Constraints}. Although the designed RAC accounts for control signal constraints in Section \ref{Model-Free RAC for Single Actuators}, it lacks a supervisory component capable of automatically monitoring the control performance of HDMMs and shutting down the system when it exhibits unstable or unreasonable behavior that could endanger the environment or itself. In the context of control theory, safety constraints for dynamic and hybrid systems are often enforced using techniques such as barrier Lyapunov functions (BLFs) \cite{wang2023concurrent, liang2023adaptive} and control barrier functions (CBFs) \cite{ames2019control, xiao2021high},
which maintaining system states within a designated safe region. 
By employing logarithmic BLFs, state or tracking error limits can be embedded directly into the RAC framework. As the tracking error or state approaches the predefined bound, the logarithmic function becomes undefined, which can be used to automatically trigger system termination if necessary. In this section, we not only extend the previous RAC strategies by incorporating safety constraints (built on \cite{shahna2024sasPEMCrobust}) but, inspired by \cite{shahna2024modelacc}, also take a further step by generalizing the proposed control framework for HDMMs with any type of actuation mechanism. Accordingly, this section is divided into two parts: hydraulic-powered HDMMs and electrically powered HDMMs, which represent the most widely used actuation mechanisms in industry.

\subsection{Hydraulic HDMMs}

\begin{figure}[h] 
  \centering
\scalebox{0.45}
    {\includegraphics[trim={0cm 0.0cm 0.0cm
    0cm},clip,width=\columnwidth]{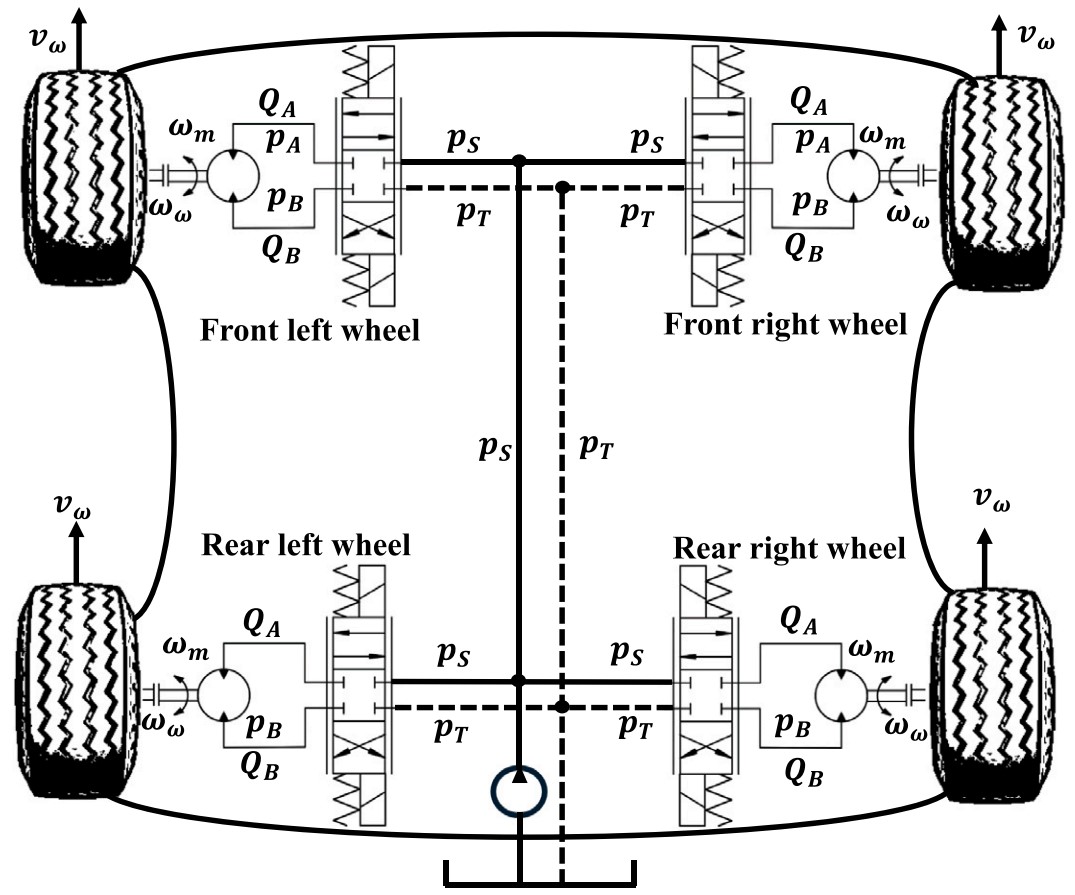}}
  \caption{\textcolor{black}{A Hydraulic-powered wheeled HDMM system (\textbf{P-5}).}}
  \label{haulfhfdsdfsdfpub1}
\end{figure}

In this section, the case study focuses on a 6000 kg hydraulic-powered WMR. The HDMM has four independently driven wheels, each actuated by its own hydraulic system. The hydraulic system is controlled by adjusting the control valve signals, which regulate the hydraulic flow to the motors. This generates sufficient torque, which, after gear reduction, drives the wheels to produce the required motion; as illustrated in Fig. \ref{haulfhfdsdfsdfpub1}.
In this context, the energy conversion stage includes the hydraulic system that generates torque, while the ODEs correspond to the HDMM’s overall movement. Each hydraulic IWD can be modeled as a third-order SF (12-order for the whole system). There are several challenges in designing a control strategy for this case study. First, torque sensors are prone to faults and often produce noisy measurements, while it is an important intermediary parameter between the hydraulic system and the wheel dynamics. Second, wheel slip and rough terrain conditions further complicate the control of this HDMM system.

\begin{figure}[h] 
  \centering
\scalebox{1}
    {\includegraphics[trim={0cm 0.0cm 0.0cm
    0cm},clip,width=\columnwidth]{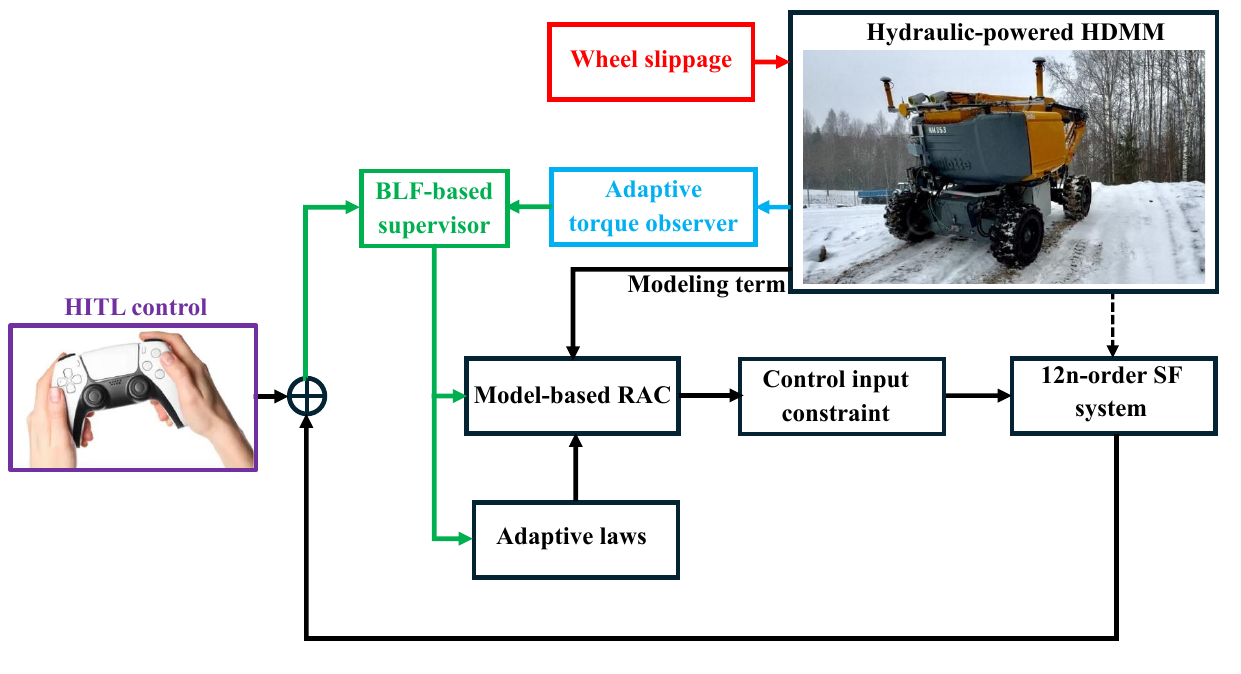}}
  \caption{Model-based RAC for a four wheeled HDMM in \textbf{P-4}.}
  \label{P-4_schematica}
\end{figure}

In \textbf{P-5}, the model-based RAC framework is developed for hydraulic-actuated HDMMs by utilizing the assigned model function proposed in \textbf{P-4}, which is employed here to incorporate modeling terms specific to the hydraulic system. To eliminate the closed-loop dependency on fault-prone torque
(or hydraulic pressure) sensors, a novel adaptive observer network is proposed within the RAC framework. This torque estimation aligns the wheel velocity with its reference value while mitigating disturbances caused by rough terrain, wheel slip, and other wheel-related effects. Then, the estimated torque is used as the reference input for the hydraulic actuation system. Within this system, another adaptive network adjusts the valve control signals, thereby modulating hydraulic pressure to either increase or decrease the force delivered by the hydraulic fluid. This regulation ensures that each actuator generates the necessary torque. 
Importantly, logarithmic BLF functions are employed as supervisory elements within the RAC architecture, and they trigger system shutdown whenever predefined performance bounds are violated. Therefore, the proposed RAC operates only after the logarithmic BLF has monitored the system, as it does not receive any direct input signals; see Fig. \ref{P-4_schematica}. As observed, in experimental tests, a human-in-the-loop (HITL) block was used as the operator’s joystick commands, while observing the robot’s behavior. These real-time inputs allow the operator to directly control the system in a remote-control mode.

\subsubsection{Experimental Validity}

\begin{figure}[h] 
  \centering
\scalebox{0.65}
    {\includegraphics[trim={0cm 0.0cm 0.0cm
    0cm},clip,width=10.5cm]{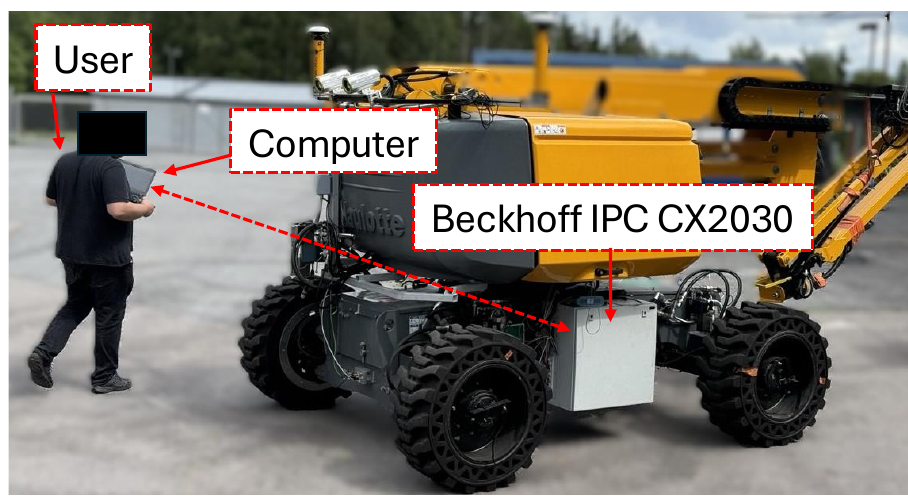}}
  \caption{Communication of the setup (\textbf{P-5}).}
  \label{beck}
\end{figure}

The experimental evaluation presented in Publication \textbf{P-5} was carried out using a hydraulic wheeled HDMM, namely the Haulotte 16RTJ PRO—an articulated boom lift equipped with four-wheel drive and a total mass of 6,650 kg. Each of its four wheels is driven by a hydraulic-powered IWD unit, controlled via high-bandwidth valve systems, and all wheels have an identical diameter of 0.854 m. The system's communication architecture, including its interface with the host computer, is depicted in Fig.~\ref{beck}.

\begin{table}[ht]
\centering
\caption{Instrumentation and hardware configuration of the hydraulic WMR \cite{shahna2025fault}.}
\begin{tabular}{l l}
\hline\hline
\textbf{Component} & \textbf{Description} \\
\hline
Kubota Diesel Engine            & 26.5 kW @ 2,800 rpm \\
Bosch Rexroth Pump              & 63 l/min \\
Danfoss OMSS Motors             & 100 cm\textsuperscript{3}/rev \\
Bosch Rexroth valves            & 40 l/min @ $\Delta p$ = 3.5 MPa \\
IFM PA3521 transducers          & Sensor range: 25 MPa \\
Danfoss EMD Speed Sensor        & 0–2500 rpm \\
Beckhoff IPC CX2030            & 1,000-Hz sample rate \\
\hline\hline
\end{tabular}
\label{tab:hardwsadaare1}
\end{table}

\begin{figure}[h] 
  \centering
\scalebox{0.65}
    {\includegraphics[trim={0cm 0.0cm 0.0cm
    0cm},clip,width=10.5cm]{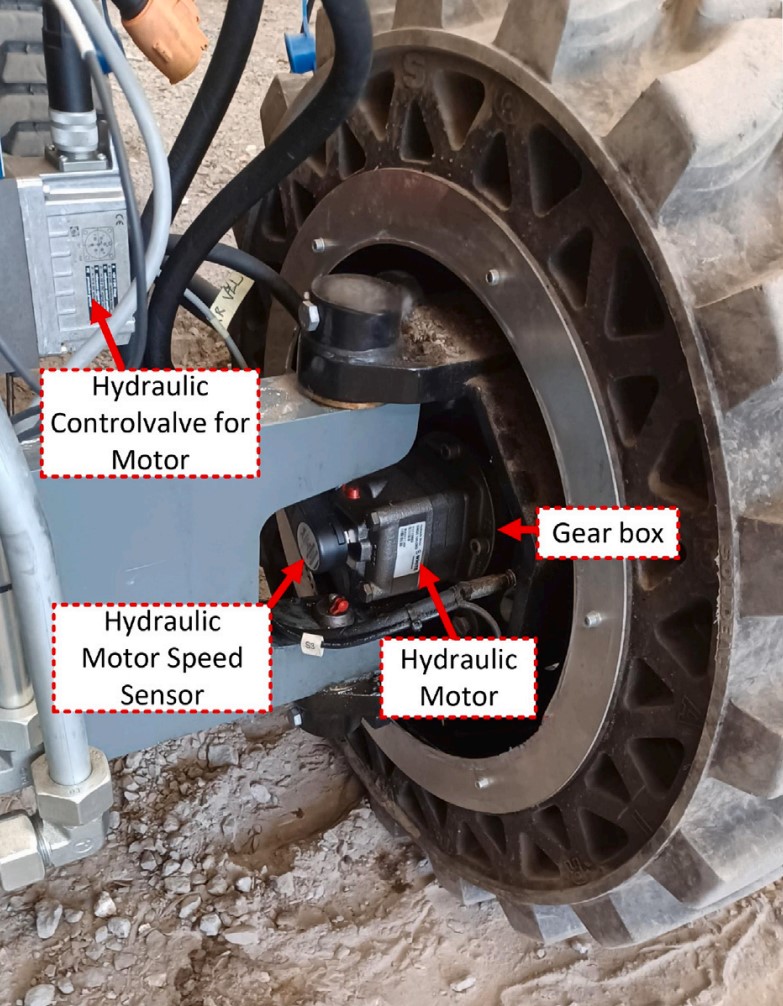}}
  \caption{Experimental setup of the case study (\textbf{P-5}).}
  \label{expasdasdp1}
\end{figure}

The drivetrain includes a reducer transmission with a gear ratio of 17.7 between the in-wheel motors and the wheels. Previous closed-loop control implementations for this system, such as the model-based PD controller in \cite{hulttinen2021model}, did not address robustness nor offer formal stability guarantees. The hardware configuration and instrumentation used in the experiments are expressed in Table \ref{tab:hardwsadaare1} and Fig. \ref{expasdasdp1}.

Wheel angular velocities were captured using electromagnetic speed sensors integrated into each in-wheel motor. These sensors enhance navigation precision by accurately measuring both rotational speed and direction. They are mounted on the end cover of each Danfoss motor and function by detecting a magnet rotating inside the motor shaft. 

To evaluate the robustness of the model-based RAC in realistic and demanding conditions, two experimental scenarios were designed involving an HDMM equipped with four hydraulically powered in-wheel drive (IWD) actuators:

Scenario 1: The robot was deployed on a snow-covered gravel surface characterized by soft soil and irregular terrain. This environment posed a high risk of wheel slippage, primarily due to the uneven distribution of the HDMM's weight at the wheel–ground contact points. This setup effectively simulates real-world off-road conditions typically encountered by IWD-actuated HDMMs.

Scenario 2: The robot traversed a steep, ice-covered, rocky incline. During uphill movement, the increased gravitational load challenged the drive system, while downhill movement placed significant demand on the braking mechanism. Combined with the low-friction icy surface, these factors created a high-risk scenario where the 6,650-kg heavy-wheeled mobile robot (HWMR) was susceptible to skidding or tipping.

\begin{figure}[h!]
\centering
\scalebox{0.55}{\includegraphics[trim={0cm 0.0cm 0.0cm 0cm},clip,width=\columnwidth]{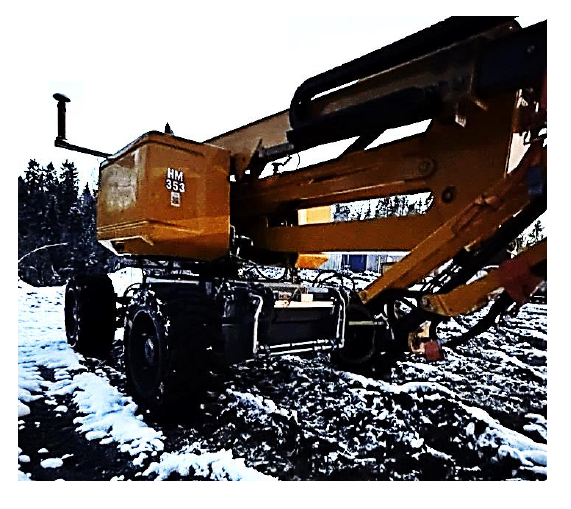}}
\caption{Experiment on a rough terrain. The relevant video is available at: {\href{https://youtu.be/4vSnfQ9PQAA}{https://youtu.be/4vSnfQ9PQAA}} (\textbf{P-5}).}
\label{expp1}
\end{figure}

\begin{figure}[h!]
\centering
\scalebox{0.55}{\includegraphics[trim={0cm 0.0cm 0.0cm 0cm},clip,width=\columnwidth]{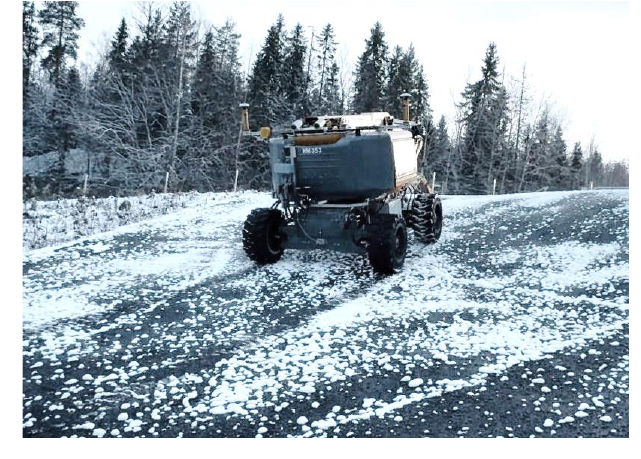}}
\caption{Experiment on a slippery terrain. The relevant video is available at: {\href{https://youtu.be/AWnTkHOndLU}{https://youtu.be/AWnTkHOndLU}} (\textbf{P-5}).}
\label{expp2}
\end{figure}

Images related to realistic challenging scenarios in \textbf{P-5} are shown in Figs.  \ref{expp1} and \ref{expp2} .

The performance of the proposed model-based RAC was evaluated against two benchmark controllers under identical experimental conditions: a model-free robust backstepping-based adaptive control (BAC) introduced in \cite{zhang2017backstepping} and a model-based robust decentralized valve-structure control (DVSC), originally introduced in \cite{schwarz2024robust}. The comparison focused on two key tracking metrics: velocity error and torque effort, quantified using the root mean squared velocity tracking error (in m/s) and the peak motor torque demand (in N·m), respectively, to ensure a fair and consistent evaluation.
Unlike the proposed RAC, which integrates a torque observer and enforces safety-defined constraints, both BAC and DVSC relied on direct torque feedback from the HWMR’s onboard pressure sensors and lacked built-in safety mechanisms, requiring the use of an emergency stop when critical thresholds were exceeded.
In addition to demonstrating uniformly exponential stability, the RAC framework provided improved torque estimation and incorporated safety-aware control, leading to superior tracking performance in the control of the four hydraulic IWDs. For comparison, a PID controller was also tested. However, it failed to effectively manage the HWMR, likely because PID control is more suited for systems with predominantly second-order dynamics. In contrast, the hydraulic-powered IWD-actuated HDMM exhibits higher-order dynamics and is subject to significant disturbances from terrain-induced slippage and uncertainties inherent in hydraulic actuation systems.

Hence, \textbf{P-5} provides further evidence for addressing \textbf{RQ1} and \textbf{RQ2}. This work was later extended in \cite{shahna2025fault} to handle both sensor and actuator faults, and in \cite{shahna2025anti} to explicitly model wheel slippage. Together, \cite{shahna2025fault} and \cite{shahna2025anti} support \textbf{P-5} in addressing \textbf{RQ4} to a significant extent. Supported by \textbf{P-4}, \textbf{P-5} also meets both \textbf{C1} and \textbf{C2} to a significant extent.

\subsection{Electrical HDMMs}
Inspired by PPC \cite{zhang2019low}, this section develops the supervisory safety component proposed in \textbf{P-5}. Rather than employing constant performance bounds, \textbf{P-6} introduces an exponential functional bound for the tracking error within the logarithmic framework. This formulation allows the definition of performance limitations based on control metrics such as overshoot bounds and steady-state error, as follows:

\begin{equation}
\begin{aligned}
\small
\label{33}
o(t)=\left(o^{shoot}-o^{bound}\right) e^{-o^* t}+o^{bound}
\end{aligned}
\end{equation}

Given that $o^{\text {shoot }}>o^{\text {bound }}>0$ and $o^*>0$, the overshoot of tracking error is constrained to remain below $o^{\text {shoot }}$. The parameters $o^{\text {bound }}$ and $o^*$ define the steady-state error bound and the convergence rate of the RAC performance, respectively. As the tracking error or state approaches the predefined metric threshold, the logarithmic function becomes undefined, which can be used to automatically trigger system termination if necessary.

\begin{figure}[h] 
  \centering
\scalebox{1}
    {\includegraphics[trim={0cm 0.0cm 0.0cm
    0cm},clip,width=\columnwidth]{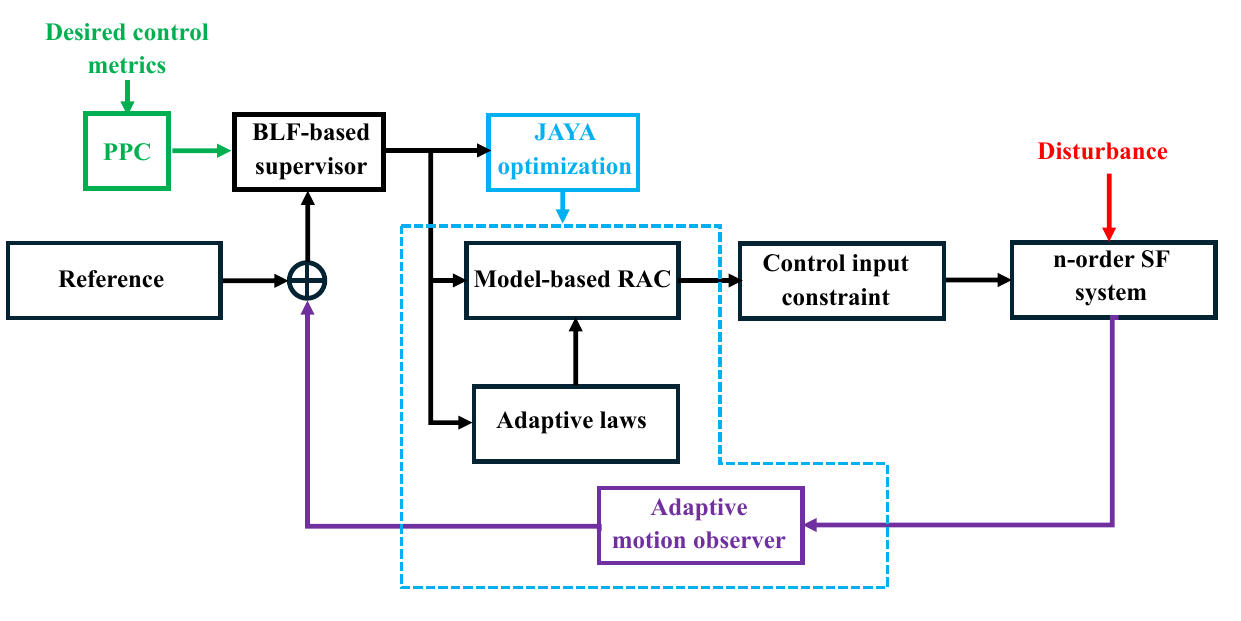}}
  \caption{Model-based RAC with PPC for an n-order SF system \textbf{P-6}.}
  \label{P-5_schematica}
\end{figure}

Hence, \textbf{P-6} proposes a novel model-based RAC for an n-order SF system, guaranteeing control performance in terms of both control metrics and control signals; see Fig. \ref{P-5_schematica}. In addition, novel adaptive networks are developed to estimate all states of the SF systems. The proposed RAC ensures uniform exponential stability, and the trade-off between system responsiveness and robustness is deeply analyzed. Furthermore, the JAYA algorithm, previously employed in \textbf{P-3}, is utilized to optimize the control and observer parameters, thereby avoiding the complex manual tuning typically required in high-order systems. 

\subsubsection{Experimental Validity}
\begin{figure}[h] 
  \centering
\scalebox{1}
    {\includegraphics[trim={0cm 0.0cm 0.0cm
    0cm},clip,width=\columnwidth]{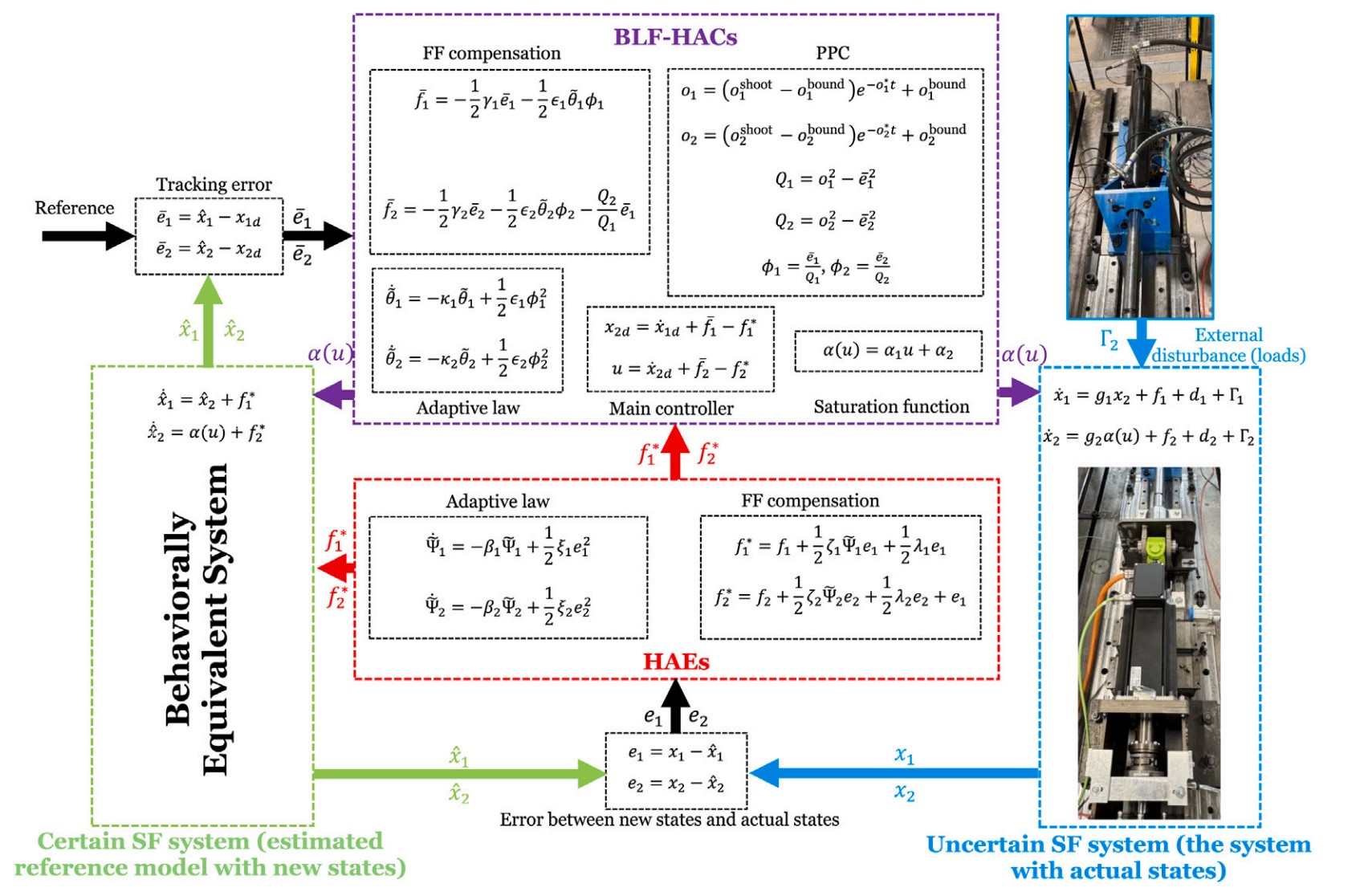}}
  \caption{Schematic of the safe model-based RAC framework applied to the studied EMLA (\textbf{P-6}).}
  \label{P-5_schadffematica}
\end{figure}

The proposed control policy is validated through extensive experiments on the studied PMSM-powered EMLA, tested under load conditions ranging from 0–95$\%$ of its full capacity; see \ref{P-5_schadffematica}. In this experiment, the proposed RAC utilizes two adaptive networks. The first network estimates the states and modeling terms of the PMSM-powered EMLA by constructing an equivalent reference system, to which the actual system states are driven to converge. The second network, enhanced by PPC, compensates for the estimated model dynamics to ensure accurate trajectory tracking while maintaining stability and meeting control performance criteria within a BLF-based PPC framework. A comprehensive comparison was conducted between the proposed safe model-based RAC and two benchmark control strategies—Adaptive Active Torque Control (AATC) \cite{zhao2022dual} and Adaptive Fault-Tolerant Control (AFTC)\cite{zhang2024robust}—under identical experimental conditions. The evaluation focused on tracking accuracy, response speed, control effort, and robustness for the PMSM-powered EMLA operating under uncertain SF conditions with disturbances.
The results showed that the safe RAC achieved the highest overall accuracy, exhibiting the lowest mean squared tracking errors in both position and velocity subsystems. AATC, in comparison, demonstrated larger tracking errors—particularly in the velocity subsystem—indicating reduced effectiveness in velocity regulation. AFTC delivered moderately improved accuracy over AATC but still fell short of the precision achieved by RAC.
In terms of dynamic response, the RAC also outperformed both alternatives by delivering the fastest transient performance, with settling times of 0.42 seconds for position tracking and 0.43 seconds for velocity tracking. AATC showed the slowest convergence, while AFTC responded faster than AATC but remained slower than RAC.
Regarding control effort, the torque amplitudes generated by RAC and AATC were notably lower than those of AFTC, which peaked at 36.5 Nm (highlighted in red). This indicates that AFTC required higher actuation effort during operation.
The RAC also successfully satisfied all performance constraints across both load-increasing (50–70 kN) and load-decreasing (70–50 kN) scenarios. In contrast, AATC violated position constraints under high-load conditions, while AFTC exceeded torque constraints during load reduction—both violations are marked in red in the results summary. Although AATC and AFTC demonstrated a degree of robustness under varying load conditions, their ability to consistently meet all prescribed constraints was inferior to that of RAC.

While the performance of AATC and AFTC came close to RAC in some aspects, it is important to highlight the context: the tested EMLA system represents a single joint prototype of a larger electrically actuated heavy-duty manipulator, which will ultimately incorporate two additional EMLA-driven joints. In such multi-joint systems, even small tracking deviations—on the order of millimeters—can accumulate and lead to substantial end-effector errors during complex operations. As such, the superior precision, stability, and constraint satisfaction achieved by the proposed RAC make it a more reliable and appropriate solution for high-performance robotic applications in heavy-duty environments.

Thus, building on \textbf{P-1}–\textbf{P-5}, \textbf{P-6} fulfills both \textbf{C1} and \textbf{C2}, fully addresses \textbf{RQ3}, and partially addresses \textbf{RQ4}. In \textbf{P-6}, to monitor system safety, we considered logarithmic BLFs enhanced by PPC. A limitation of the BLF concept is that it terminates system operation whenever PPC is violated, regardless of whether the condition is truly dangerous or not. However, these approaches tend to be overly conservative because they do not actively steer the system back toward safer conditions once the boundary of the safe region is reached \cite{liu2019barrier}. To enhance system availability, the next section introduces an intelligent supervisory safety component that can automatically manage the trade-off between robustness and system responsiveness. Thus, the final missing piece for fully addressing \textbf{RQ4} (system maintainability and reliability, and finally improving availability) is presented in the next section.

\subsection{Hybrid HDMMs}
Build on publications \textbf{P-1} to \textbf{P-6}, \textbf{P-7} seeks to answer a core research question: How can learning-based control strategies—intended to avoid complex analytical modeling while delivering strong nominal performance—be reliably implemented on HDMMs with intricate actuation architectures, while ensuring compliance with international safety standards for both system safety and stability? 

\begin{figure}[h] 
  \centering
\scalebox{0.75}
    {\includegraphics[trim={0cm 0.0cm 0.0cm
    0cm},clip,width=\columnwidth]{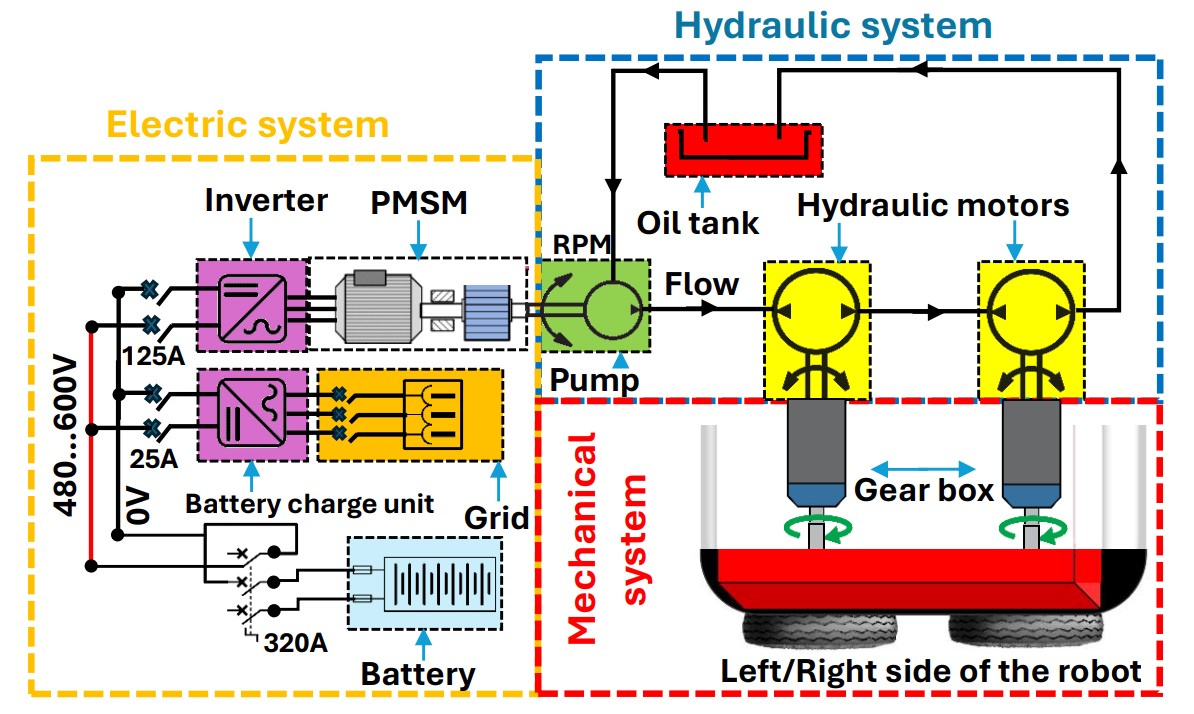}}
  \caption{Mechanical structure of each side of the studied HDMM \cite{shahna2025lidar}.}
  \label{P-5_schemataadica}
\end{figure}

Drawing inspiration from Section 10 of ISO/IEC TR 5469 on AI and functional safety \cite{iso5469}, supervisory barrier functions can be used as monitoring tools to identify unsafe behaviors in AI-driven systems, whether they arise from internal malfunctions or external disturbances. Once such conditions are detected, conventional stabilizing controllers can be activated to preserve safety. Following this principle, \textbf{P-6} proposes a hierarchical control framework for a skid-steer HDMM with hybrid actuation systems. The mechanical structure of the HDMMs is shown in Fig. \ref{P-5_schemataadica}.

To achieve high performance and operational availability in the studied HDMM, \textbf{P-6} introduces a stability-guaranteed hierarchical control framework, supervised by two safety layers with different levels of authority, as illustrated in Fig. \ref{i9sasddadado_synthes}. This framework is composed of three main components:

\begin{figure}[h!]
\hspace*{-0.0cm} 
\centering
\includegraphics[width=1\textwidth, height=8cm]{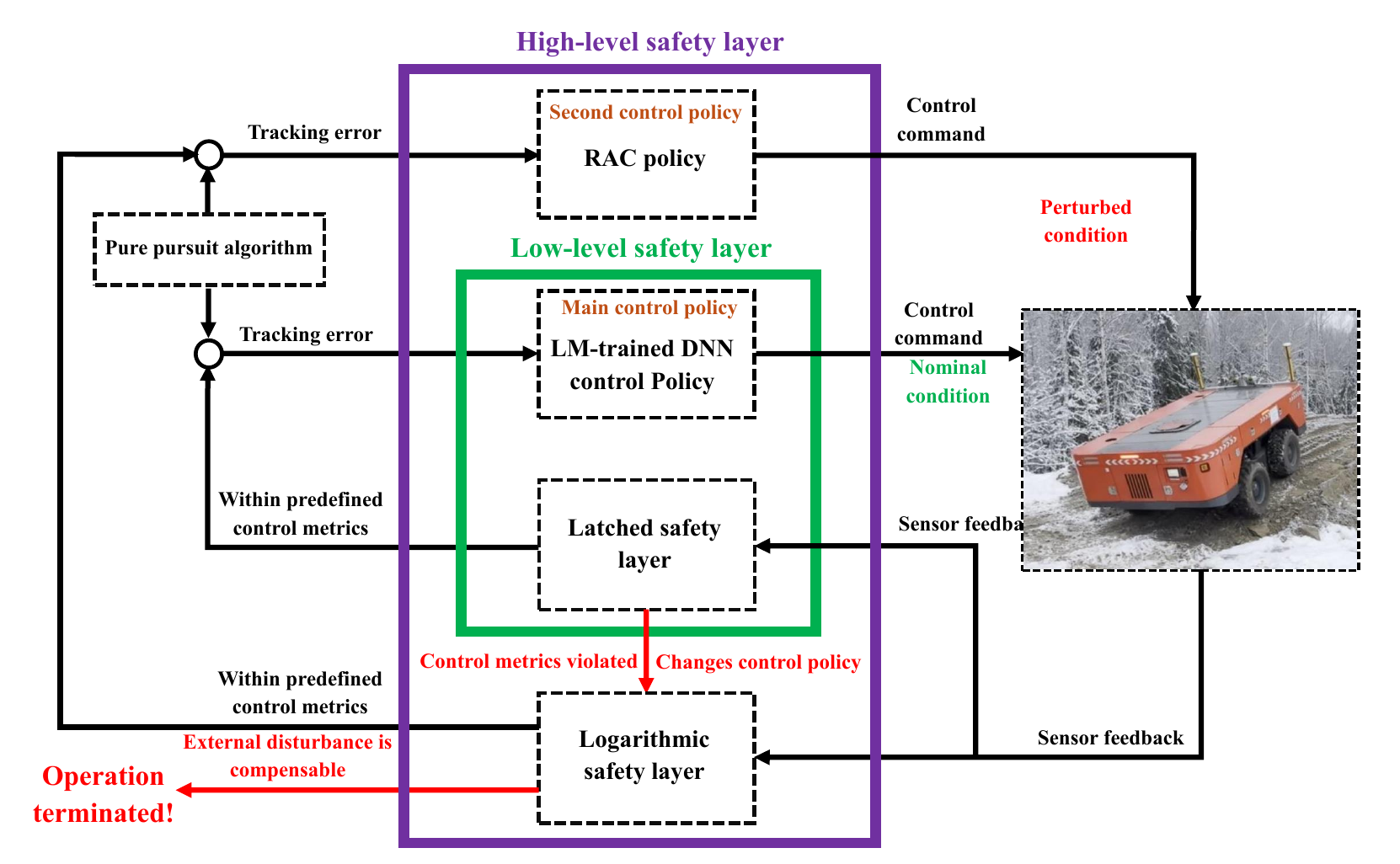}
\caption{Synthesis of LM-trained DNN-based Control policies with the RAC policy, governed by two safety layers with different authorities for a hybrid HDMM in \textbf{P-6}.}
\label{i9sasddadado_synthes}
\end{figure}

1) A geometric path-tracking method, the pure pursuit algorithm, is used for wheel-level task commands. In \textbf{P-6}, it generates two types of reference trajectories: a circular path and an S-shaped trajectory.

2) At the core of the proposed control framework is a high-performance deep neural network (DNN) controller, trained on data collected from onboard sensors to capture the relationship between control inputs and system motion. Training is performed using the second-order Levenberg–Marquardt (LM) backpropagation method. The training dataset is generated in a controlled environment with a simple open-loop strategy, where input commands are gradually ramped to their maximum limits under human supervision to safely record system responses. The trained DNN controller is then deployed as the primary control policy during nominal operations, delivering precise and efficient performance.

3) A lower-level safety layer, implemented via a latched function, continuously monitors the system and intervenes when disturbances degrade performance beyond the predefined PPC threshold. Upon detection, it disables the DNN controller and switches to the RAC policy, enhancing system robustness and preserving safe operation, albeit at the cost of reduced responsiveness for the remainder of the operational cycle.

4) A high-level safety layer developed by BLF-based PPC continuously supervises system behavior regardless of the active control policy. This layer initiates a complete system shutdown only if disturbances reach a level where compensation is no longer possible, thereby preventing unsafe operation and mitigating risks to the system and its environment.

\subsubsection{Experimental Validity}

\begin{figure}[h] 
  \centering
\scalebox{0.55}
    {\includegraphics[trim={0cm 0.0cm 0.0cm
    0cm},clip,width=\columnwidth]{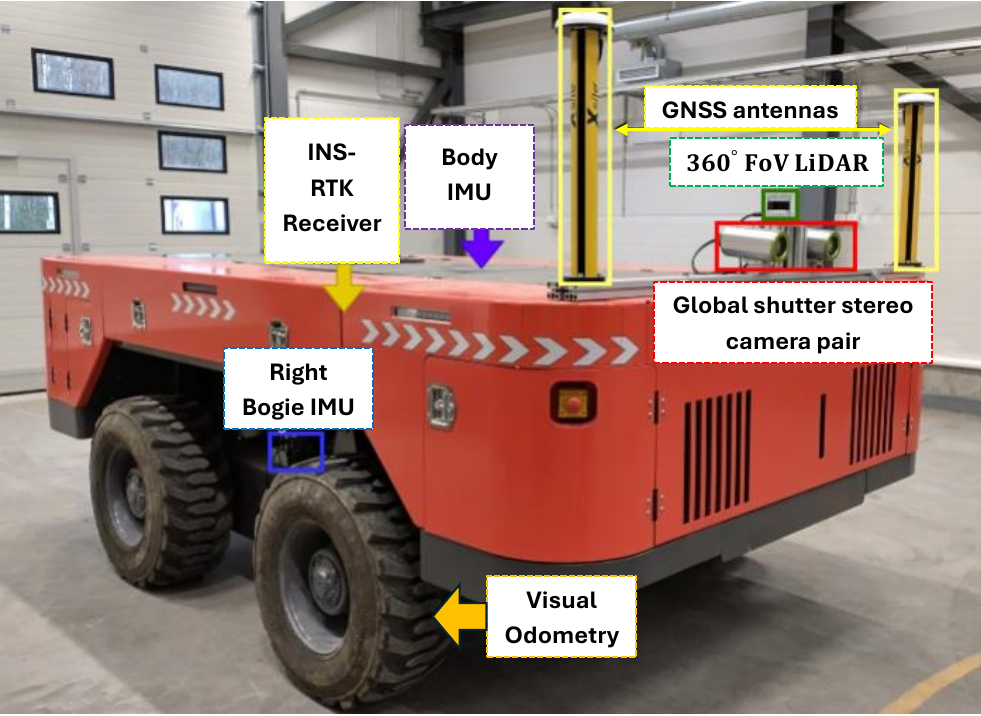}}
  \caption{The dimensions of the MPD are $2 m \times 4 m \times 1.5 \mathrm{~m}$ \cite{shahna2025lidar}.}
  \label{MPD_virtuall}
\end{figure}

The validation of the algorithms proposed in \textbf{P-7} was conducted using a 6,000 kg skid-steered wheeled HDMM, known as the Multi-Purpose Deployer (MPD), whose actuation system employs a hybrid of electric motors and hydraulic components. This platform includes two identical active suspension bogies, one on each side, employing a serial-parallel actuation architecture that integrates actuator chains. Each side of the vehicle is powered by a PMSM, which acts as the primary force input to the system and is controlled through its rotational speed (rpm) signal. The electric drive system includes a battery-powered inverter that regulates both speed and torque of the PMSMs, while the hydraulic circuit transfers power from the pump to the motors. The PMSM drives a hydraulic pump that supplies fluid power to two serially connected hydraulic motors. These motors, mechanically coupled and typically rotating at the same speed, transmit torque through a high-ratio gearbox to the wheels on the corresponding side. Angular velocity is measured using a speed sensor placed either on one of the hydraulic motors or directly on the wheels, providing real-time feedback for control.
Each hydraulic motor is equipped with an odometry sensor, and the velocities from the two wheels on each side are averaged to compute the effective linear velocity. This integrated system handles energy conversion, torque amplification, and mechanical transmission for wheel actuation. Fig. \ref{MPD_virtuall} shows the image of the MPD and installed sensors on it. 

\begin{figure}[h!]
\hspace*{-0.0cm} 
\centering
\includegraphics[width=0.7\textwidth, height=6cm]{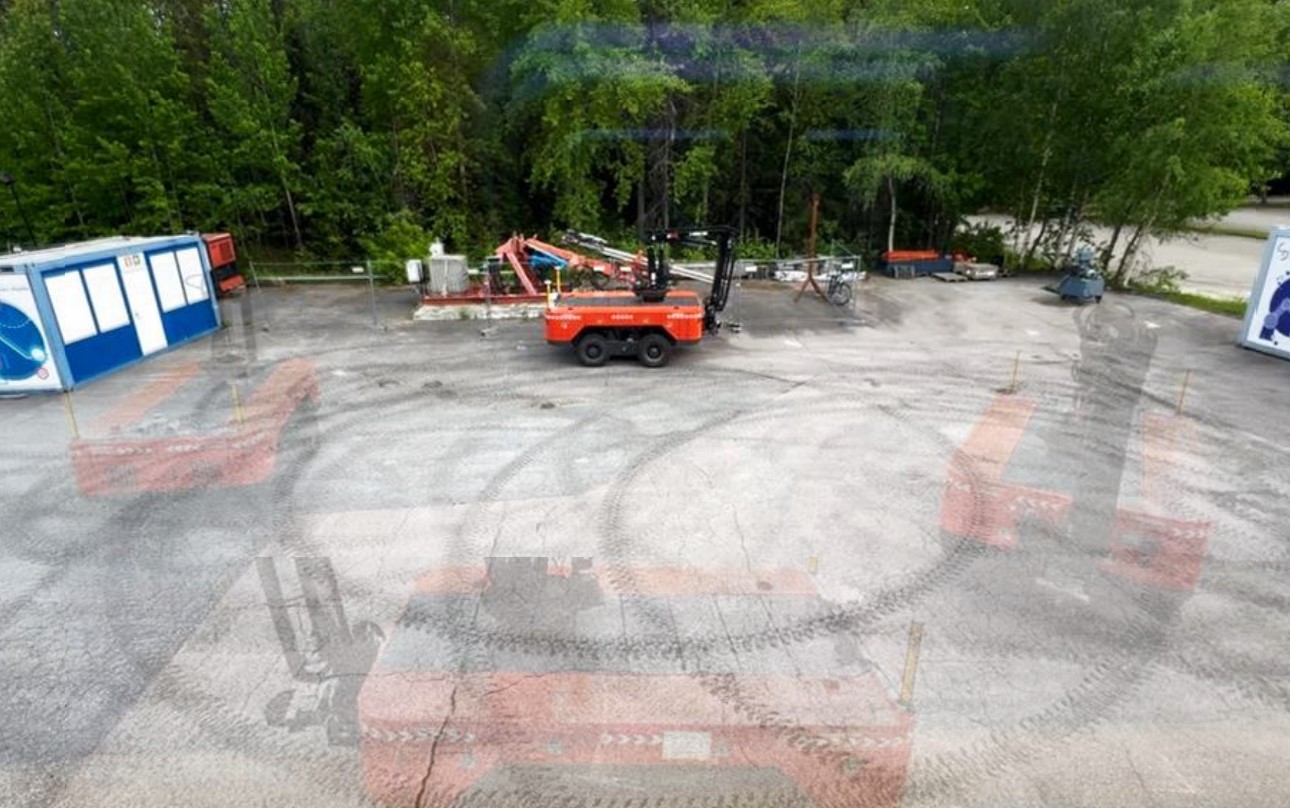}
\caption{Circular trajectory tracked by the SSWMR in \textbf{P-7}. The relevant video is available at: \href{https://youtu.be/VbGeSchr9ek}{https://youtu.be/VbGeSchr9ek}.}
\label{circcccular}
\end{figure}

\begin{figure}[h!]
\hspace*{-0.0cm} 
\centering
\includegraphics[width=0.7\textwidth, height=6cm]{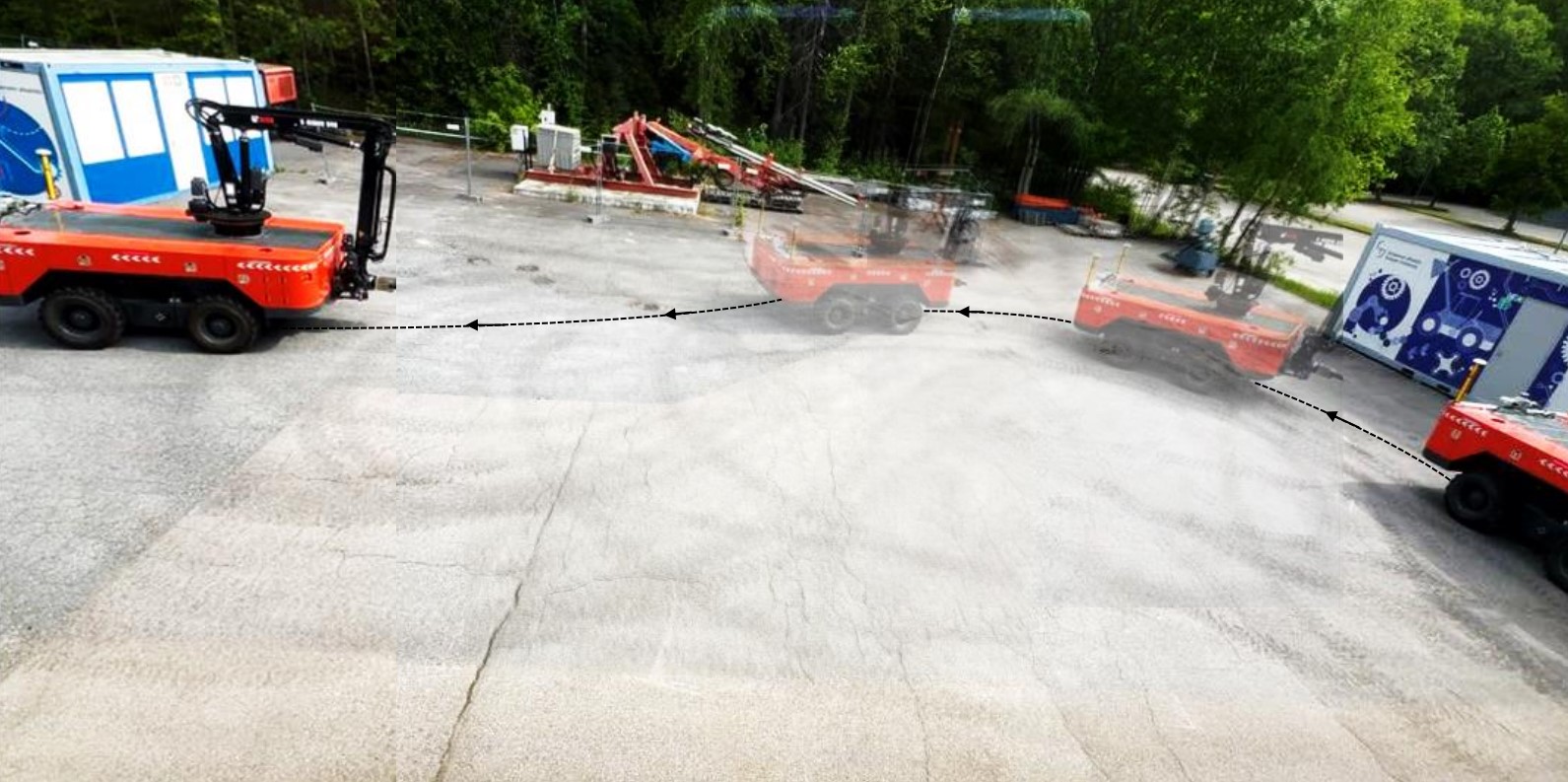}
\caption{S-shape trajectory tracked by the SSWMR in \textbf{P-7}. The relevant video is available at: \href{https://youtu.be/cRXNikD8vd0}{https://youtu.be/cRXNikD8vd0}.}
\label{control.s}
\end{figure}

Sensor data acquisition is managed by an embedded Beckhoff industrial PC using the EtherCAT communication protocol. Figs. \ref{circcccular} and \ref{control.s} show the test operations which have been fulfilled in \textbf{P-7}.

The proposed framework guarantees uniformly exponential stability and has been experimentally validated on an HDMM under realistic operational conditions across three scenarios: DNN-based control under nominal conditions, safe model-based RAC operation under perturbed conditions, and the synthesis of DNN and model-based RAC policies under perturbations. The results demonstrate an improvement in overall system availability.

Hence, \textbf{P-7} completes the final pieces required to fully address \textbf{RQ3}, \textbf{RQ4}, and \textbf{RQ5}, as outlined in Section \ref{ch:Introduction}. It also fulfilled \textbf{C3}.

\chapter{Discussion}
\label{ch:Discussion}

This section summarizes how each of the \textbf{RQ}s outlined in Section 1.6 has been addressed.

\section{\textbf{RQ1}: Generic RAC for High-Order Actuators}

\textit{How can we achieve a robust control strategy that is generic and
applicable to any high-order actuation mechanism, regardless of
the type of energy conversion, be it hydromechanical, electromechanical, hybrid, or any emerging future alternative?}

The foundation of the current research was laid in \textbf{P-1}, where a model-free RAC approach was proposed for EMLAs in HDMMs, aligning with the broader trend of electrification. This control strategy not only provides exponential stability guarantees but also surpasses traditional methods in terms of robustness to intense modeling and parametric uncertainties and load disturbances, particularly in rough-terrain operations.
Crucially, the model-free nature of the proposed RAC makes it inherently adaptable to various actuation mechanisms, regardless of the underlying energy conversion method—be it hydraulic, electric, or future technologies. Building on this, the methodology was generalized in \textbf{P-2}, where the same control approach was extended to all common actuator types used in HDMMs.
Therefore, the contributions in \textbf{P-1}, supported and extended by \textbf{P-2}, experimentally addressed \textbf{RQ1}, demonstrating the feasibility and generalizability of a unified model-free RAC framework across different actuator technologies in HDMMs.

\section{\textbf{RQ2}: Development from High-Order Actuator Control to Multi-Body HDMMs}

\textit{How can the control strategy developed in RQ1, originally
designed for a single actuation mechanism, be extended to multi-body HDMMs, including both mobile robot platforms and robotic manipulators, that incorporate multiple types of actuation mechanisms, while maintaining its effectiveness and stability in executing robotic tasks without modification?}

As discussed in Section \ref{electrififcan}, multi-body HDMMs exhibit complex robotic interactions due to their multi-body structures and the nature of their operational environments. Section \ref{Actuation mechanisms in HDMMs} further elaborated that an $n$-DoF robotic manipulator can comprise up to $4\times n$ interconnected subsystems, significantly complicating the control design process.
Building on \textbf{P-1} and \textbf{P-2}:

 1) For validation on multi-link robotic systems, the model-free RAC was extended in \textbf{P-3} to an n-DoF robotic manipulator, focusing exclusively on second-order motion dynamics. This controller demonstrated the ability to execute robotic tasks despite the presence of joint faults, while maintaining overall system stability with exponential convergence and accounting for the dynamic interactions among all n joints. Expanding on this, and leveraging the subsystem modularity concept introduced in \cite{koivumaki2022subsystem}, \textbf{P-4} extended the RAC framework to a full-scale heavy-duty manipulator actuated by emerging heavy-duty EMLAs. The modular representation of PMSM-powered EMLA joints allowed the unified control structure to remain unchanged, even when the system underwent dynamic modifications such as motor replacements, changes in component dynamics, or structural alterations involving the addition or removal of joints.

2) For validation on mobile robot systems and building on the generality of this RAC policy, \textbf{P-5} applied the same modularity-based approach to a $6000$-kg four-wheeled HDMM actuated by a valve-controlled hydraulic system, enabling it to perform complex robotic tasks on snowy, rough terrain.

Therefore, the work presented in \textbf{P-4} and \textbf{P-5}, supported by the theoretical foundation in \textbf{P-3}, experimentally addressed \textbf{RQ2}. The contributions were further validated and reinforced by the results presented in related publications \textbf{P-8} \cite{shahna2024integrating}, \textbf{P-13} \cite{kolagar2024combining}, and \textbf{P-14} \cite{bahari2024system}.

\section{\textbf{RQ3}: Trade-Off between System Responsiveness and Robustness}

\textit{How can we manage the trade-off between system responsiveness and robustness in a developed robust controller, particularly in HDMMs where intense disturbances amplify the difference between nominal and perturbed conditions?}

As discussed in Section \ref{modelfrrebased}, a fundamental and well-recognized trade-off in control theory exists between system robustness and responsiveness. To address noise-corrupted state measurements, \textbf{P-4} experimentally developed an adaptive observer network, which was integrated into the control framework to filter noisy encoder position signals and estimate velocity states. During the stability analysis, the observer parameters were carefully tuned to balance noise rejection with system responsiveness.
Building on this balancing concept and a deeper analysis of the stability proofs, \textbf{P-6} conducted a comprehensive investigation into the trade-off between responsiveness and robustness in HDMMs. It highlighted how these characteristics influence the tuning of all design parameters within the proposed RAC framework, while maintaining exponential stability of the system. This balance is particularly critical for HDMMs, which are subject to high dynamic disturbances that can significantly affect control performance.

\section{\textbf{RQ4}: Improvement of System RAMS}

\textit{As HDMMs are inherently prone to faults, degradation, and
efficiency losses over time, how can we improve system RAMS, under various undesirable conditions in HDMMs?}

In addition to handling control input constraints, \textbf{P-5} developed a logarithmic BLF framework for a hydraulically powered wheeled HDMM operating under wheel slippage conditions. This framework functioned as a supervisory element within the model-based RAC architecture, triggering system shutdown whenever predefined performance bounds were violated. Consequently, the proposed RAC was activated only after the logarithmic BLF had verified safe operating conditions, as the controller did not receive direct access to raw system output signals.
Inspired by the PPC approach in \cite{xie2024low}, \textbf{P-6} extended the supervisory safety mechanism introduced in \textbf{P-5}. Instead of relying on constant performance bounds, \textbf{P-6} introduced dynamic bounds for control performance metrics—such as overshoot and steady-state error—within the logarithmic BLF framework.
It is important to note that the proposed RAC's fault tolerance capability for HDMMs was experimentally validated in \textbf{P-12} \cite{shahna2025fault} and \textbf{P-11} \cite{shahna2025anti}, which supports the main findings in \textbf{P-3}. To enhance system availability, \textbf{P-7} developed an intelligent supervisory safety element that enabled switching between primary and backup control policies, thereby enhancing system availability under severe disturbances. Thus, \textbf{RQ4} was experimentally addressed through contributions in \textbf{P-3}, \textbf{P-5}, and \textbf{P-6}, with additional support from \textbf{P-12} and \textbf{P-11}, and \textbf{P-7} completed the final pieces required to fully addressed \textbf{RQ4}.

\section{\textbf{RQ5:} Stability-Guaranteed Learning-based Technologies with Safety-Defined Metrics}

\textit{How can we can design an intelligent supervisory component
to guarantee system stability and safety-defined metrics when leveraging learning-based technologies policies, without requiring interpretability or transparency from these complex and stochastic black-box models?}

\textbf{P-5} and \textbf{P-6} progressively established the foundation for a supervisory control framework in HDMMs. Specifically, \textbf{P-5} introduced constant constraints based on BLFs into a stability-guaranteed model-based RAC strategy, laying the groundwork for integrating safety boundaries into control design. Building on this, \textbf{P-6} extended the approach by incorporating control performance metrics derived from PPC, enabling the enforcement of predefined system behavior.
This line of development culminated in Publication \textbf{P-7}, supported by additional insights from \textbf{P-10} and \textbf{P-11}, which presented a complete and experimentally validated high-performance control architecture. In this work, the RAC strategy was synthesized with DNN policies to form a robust switching controller structured around a two-layer PPC-based supervisory mechanism. The lower-level safety layer, implemented using a latched function, continuously monitored the system and intervened when performance degraded beyond predefined PPC thresholds. In such cases, it automatically disabled the DNN controller and switched to the RAC policy, thereby reinforcing system robustness and preserving safe operation. In parallel, the high-level safety layer, constructed using a BLF-based PPC formulation, supervised overall system behavior irrespective of the active control policy. This layer was responsible for initiating a complete system shutdown when disturbances exceeded the system's compensatory capabilities, thus ensuring safe operation and mitigating risks to both the machine and its environment.

\chapter{Conclusions}
\label{ch:Conclusions}

This thesis has developed significant new methods for increasing the autonomy, robustness, and safety of HDMMs by advancing a unified RAC framework applicable across various actuation mechanisms and operating conditions. The conducted work culminates in a modular and experimentally validated control architecture that combines the RAC, safety-guaranteed supervisory elements, and learning-based policies. As summarized in Chapter 4, all five \textbf{RQ}s were experimentally addressed through the accompanying publications:

\textbf{P-1} and \textbf{P-2} experimentally addressed \textbf{RQ1} by proposing and generalizing a model-free RAC strategy for diverse actuator types, including hydraulic, electric, hybrid, and possible future alternatives. The approach offers strong stability guarantees and robustness under high-dynamic disturbances, making it well-suited for HDMMs operating in uncertain environments.

\textbf{P-3}, \textbf{P-4}, and \textbf{P-5} experimentally extended this foundation to multi-body HDMMs, effectively answering \textbf{RQ2}. The proposed control strategy retained its general form across system reconfigurations, such as actuator replacements or structural changes. These contributions were further extended in the related publications \textbf{P-8}, \textbf{P-13}, and \textbf{P-14}, which demonstrated the scalability and applicability of the modular control framework in complex robotic tasks.

The fundamental trade-off between system robustness and responsiveness, a key challenge in HDMM control, was examined in \textbf{P-4} and \textbf{P-6}, directly addressing \textbf{RQ3}. To manage this, \textbf{P-4} introduced an adaptive observer network to filter noisy sensors, while \textbf{P-6} further analyzed how this trade-off impacts the tuning of all control design parameters, emphasizing its importance for maintaining stability under high dynamic disturbances.
The experimental results showed that tuning RAC parameters allows for effective disturbance filtering without sacrificing responsiveness, thus ensuring safe and stable operation.

\textbf{RQ4} was answered through a series of developments from \textbf{P-5}, \textbf{P-6}, and \textbf{P-7}, where supervisory safety mechanisms were implemented using BLF- and PPC-based control structures. These were capable of enforcing dynamic performance limits and triggering safe fallback behavior. Additional support from \textbf{P-11} and \textbf{P-12} confirmed the system’s ability to tolerate faults and ensure availability under severe disturbances.

Finally, \textbf{P-7}, supported by \textbf{P-5}, \textbf{P-10}, and \textbf{P-11}, experimentally addressed \textbf{RQ5} by presenting a two-layer supervisory architecture that integrates DNNs with RAC policies. This hybrid system ensures that learning-based control methods operate safely and the system stability, even without requiring interpretability of the neural models, in conjunction with enforcing predefined performance and safety constraints through robust fallback mechanisms.

\begin{figure}[!h] 
    \centering
    \scalebox{1.00}
    {\includegraphics[trim={0cm 0.0cm 0.0cm
    0cm},clip,width=\columnwidth]{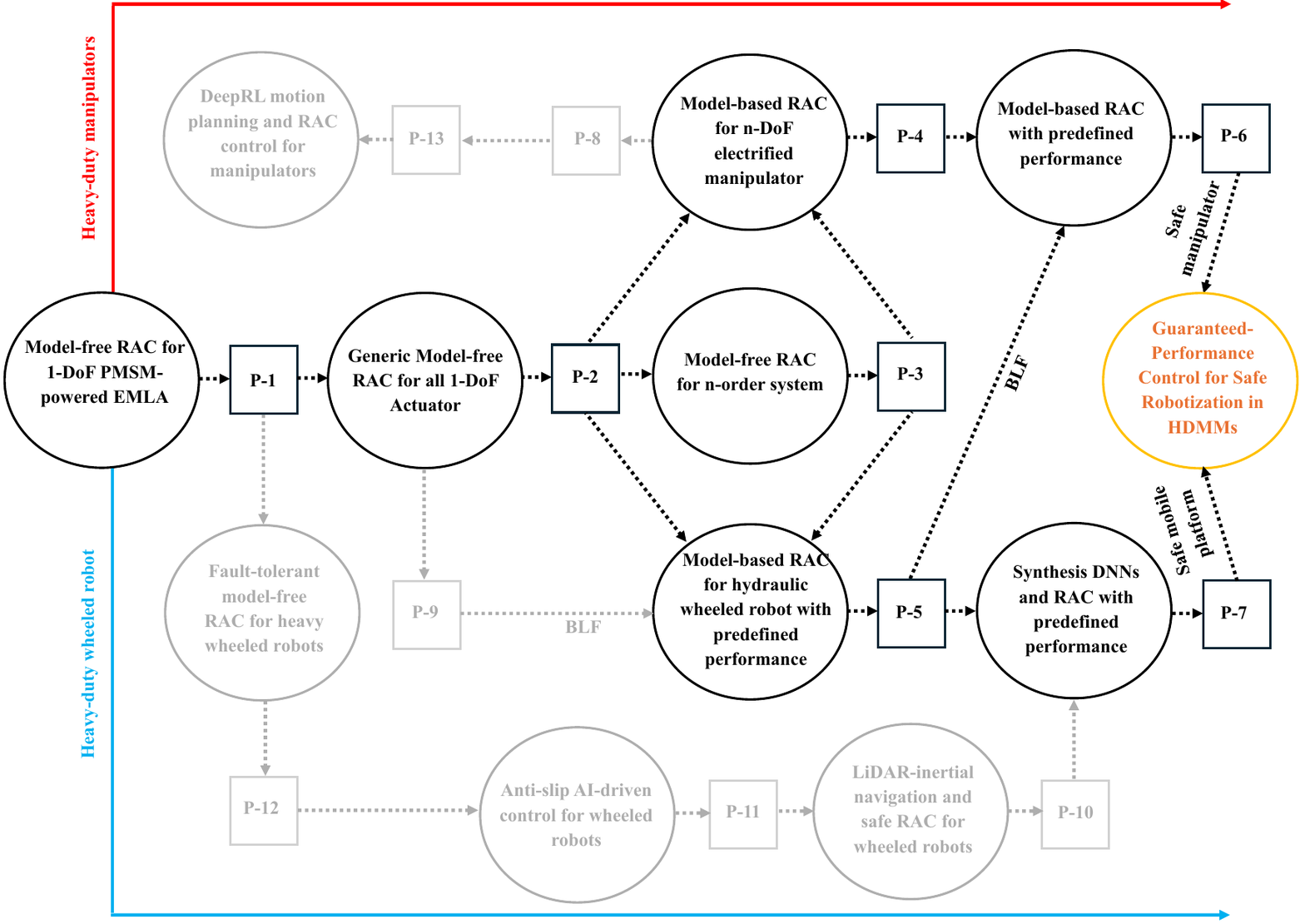}}
    \captionsetup{justification=centering}
    \caption{Overview of the articles contributing to this compendium thesis.}
    \label{sadstory}
\end{figure}

Fig. \ref{sadstory} summarizes the research line of this thesis: the core line (\textbf{P-1} to \textbf{P-7}) is shown with bold nodes and dark lines. Related articles (\textbf{P-8} to \textbf{P-14}), which were not discussed in detail in this thesis, appear in grey as they support and complement the main research narrative.

Looking toward the future of the developed control framework in emerging HDMM applications, several important challenges remain:

1) One key direction involves unifying manipulator (multi-link robot) control strategies with wheeled platform (mobile robot) control within a system to perform complex robotic tasks such as carrying heavy loads. While this thesis addressed both domains separately, their seamless integration remains an open challenge.

2) This thesis presents a generic control framework intended for a wide range of actuation systems. However, implementations were developed and validated using a SF model representing the current actuation mechanism studied. For potential future use with alternative actuation mechanisms, especially those resulting in higher-order system models, careful consideration will be needed to appropriately extend the current control design to accommodate the increased system complexity.

3) As the power stage was not available during this study, the focus was placed primarily on the PMSM-powered EMLAs, specifically from the current control loop to the mechanical load. The system was decomposed into subsystems starting from the motor's electrical dynamics onward. However, the upstream components, such as the battery, high-frequency (e.g., 20 kHz) DC-DC conversion stage, and their interactions with the control loop, were not explored in detail. Therefore, the author recommends that future work should extend the modeling and control framework to include the full power delivery path, from the energy source through the conversion stages to the actuator.

4) Additionally, the role of learning-based technologies in HDMMs is still in its early stages; their broader adoption is expected to accelerate progress in intelligent automation for these systems significantly.

5) Another critical area for advancement lies in improving the system’s RAMS. While this thesis proposed solutions to enhance these aspects from a dynamics and control perspective, further work is needed to reach the levels demanded by industrial deployment and certification.

6) Furthermore, although the thesis focused predominantly on the dynamics of HDMMs, the author recognizes that kinematic-level modeling and control should also be further developed. The proposed methods offer a solid foundation upon which such advancements can be built.

7) Finally, achieving full autonomy in HDMMs will require the integration of additional knowledge-driven frameworks for perception and localization, alongside the continued incorporation of advanced learning-based methods.

The author is optimistic that this thesis contributes meaningfully to the long-term vision of intelligent, autonomous, and standard-compliant HDMM systems, and looks forward with enthusiasm to the advancements the future holds.

\chapter{Summary of Publications}
\label{ch:Summary}
\section{\textbf{P-1} \textit{Robust decomposed system control for an electro‐mechanical linear actuator mechanism under input constraints}}

The paper develops a robust control strategy for an EMLA with a multi-stage gearbox, subject to input limitations, model uncertainties, and external disturbances. A detailed dynamic model is constructed that captures the nonlinear characteristics of the system, including the non-ideal behavior of the ball screw mechanism. The system is reformulated into a four-order nonlinear SF structure and divided into three interconnected parts. A novel model-free RAC method is introduced that guarantees uniform exponential stability and avoids the typical complexity growth of conventional backstepping approaches by managing the derivative of the virtual control input as an uncertain element. The approach remains effective even when disturbances and constraints are only loosely bound, making it suitable for broader applications. Simulation studies under varying operational demands confirm the reliability and effectiveness of the proposed controller.

\section{\textbf{P-2} \textit{Model-Free generic robust control for servo-driven actuation mechanisms with layered insight into energy conversions}}

Conventional model-free control is often limited in handling higher-order systems due to multi-modal oscillations, nonlinearities, and complex dynamics, which can lead to poor performance or instability. These challenges are especially prominent in primary actuation systems like hydraulically, pneumatically, and electrically driven actuators, which are widely used in industrial applications. To address these issues, this paper presents a model-free, generic RAC framework designed for servo-driven actuators. The RAC ensures strong stability and adaptability in the presence of uncertainties and load disturbances by first modeling the common dynamics, then decomposing their state-space models into smaller, manageable subsystems. Control is applied separately to each subsystem, effectively handling motion dynamics and energy conversion processes while maintaining input constraints, regardless of the actuator type. The proposed RAC method is entirely model-free, ensuring robustness and uniformly exponential stability in trajectory tracking. Moreover, all control signals remain within predefined limits, and the framework’s effectiveness is demonstrated experimentally on two complex servo-driven actuators.

\section{\textbf{P-3} \textit{Exponential auto-tuning fault-tolerant control of n degrees-of-freedom manipulators subject to torque constraints}}

The paper addresses the impact of actuator faults on the stability and tracking performance of n-DoF robotic manipulators by proposing a model-free RAC strategy capable of tolerating a wide range of failures and modeling errors. A mathematical formulation is developed that models different actuator conditions, including normal operation, stuck failure, performance degradation, and excessive torque, while also accounting for over-generated torques by faulty actuators. Building on this model, a subsystem-based fault-tolerant RAC method is designed to keep joint states close to desired trajectories despite uncertainties. The control gains are adjusted using a modified swarm intelligence algorithm that can progressively reach near-optimal levels without requiring manual parameter tuning. The method guarantees uniform exponential stability and is validated through simulations that demonstrate improved tracking accuracy and faster convergence.

\section{\textbf{P-4} \textit{Robustness-guaranteed observer-based control strategy with modularity for cleantech EMLA-driven heavy-duty robotic manipulator}}

The paper proposes a modular observer-based control strategy for fully electrified heavy-duty robotic manipulators to ensure robust performance under uncertainties, disturbances, and complex subsystem interactions. A detailed dynamic model of the PMSM-powered EMLAs, is developed to capture the mechanisms’ behavior. Reference trajectories for each joint are generated using direct collocation with B-spline curves, forming the basis of the control task. A subsystem-based RAC method is then enhanced with an adaptive state observer that accurately estimates linear position and velocity at the actuator load side, compensating for measurement inaccuracies. The approach is formulated in a unified generic structure applicable to all joints, preserving modularity even when the system configuration changes. It addresses non-triangular uncertainties as well as torque and voltage disturbances while ensuring exponential stability through Lyapunov-based analysis. Simulation and experimental results demonstrate improved tracking accuracy, faster convergence, and reduced torque effort compared to recent methods.

\section{\textbf{P-5} \textit{Robust torque-observed control with safe input–output constraints for hydraulic in-wheel drive systems in mobile robots}}

The paper introduces a robust torque-observer-based valve control framework for independently actuated hydraulic-powered IWD systems in wheeled HDMMs. A torque observer network is developed using an adaptive BLF to estimate the required torque for aligning wheel velocities with reference values while rejecting disturbances such as wheel slippage and uneven terrain effects. This estimated torque is then used as a reference for a second RAC network that regulates valve signals in the hydraulic actuation mechanism to generate the necessary torque without relying on error-prone torque or pressure sensors. The framework employs logarithmic BLFs to constrain key system signals, including valve commands, wheel velocities, and tracking errors, ensuring safe operation and preventing excessive torque generation. Experimental validation on a large-scale hydraulic-powered platform demonstrates that the approach enhances robustness, maintains motion synchronization, and guarantees exponential convergence under uncertain and harsh operating conditions, outperforming existing control methods.

\section{\textbf{P-6} \textit{Model reference-based control with guaranteed predefined performance for uncertain strict-feedback systems}}

The paper presents a reference-based RAC framework for uncertain nonlinear SF systems to ensure reliable tracking performance under time-varying disturbances and modeling uncertainties. The method operates at the subsystem level, beginning with homogeneous adaptive estimators that transform the uncertain system into a reference model by adaptively estimating its dynamics. This reference model is then regulated using homogeneous adaptive controllers enhanced by logarithmic BLFs, which enforce user-defined transient and steady-state performance requirements while addressing control input saturation. A generic stability connector manages dynamic interactions between subsystems, guaranteeing uniform exponential stability without the need to compute analytic derivatives of virtual controls. The sensitivity of the framework's parameters is analyzed to balance robustness and responsiveness. Experimental results on an EMLA demonstrate improved disturbance rejection and tracking accuracy compared with existing adaptive control methods.

\section{\textbf{P-7} \textit{Synthesis of deep neural networks with safe, robust adaptive control for reliable operation of wheeled mobile robots} (\textbf{Unpublished})}

The paper proposes a hierarchical control framework for wheeled HDMMs that integrates DNN control with RAC policies to balance accuracy, robustness, and safety. The DNN serves as the primary control policy, providing high-precision performance during nominal conditions. A low-level safety layer monitors disturbances and switches to a RAC strategy when system performance deteriorates beyond a threshold, maintaining stability at the expense of reduced accuracy. A high-level safety layer continuously supervises overall system behavior and triggers a shutdown only when disturbances exceed a level that makes compensation infeasible, ensuring compliance with safety standards. This architecture improves interpretability and fault tolerance while guaranteeing uniform exponential stability. Experimental validations on large-scale robotic platforms demonstrate the effectiveness of the combined control strategy compared with standalone approaches.



\printbibliography[heading=bibintoc, notkeyword={thisdissertation}]

@article{agreement2015united,
  title={United nations framework convention on climate change},
  author={Agreement, Paris},
  journal={Paris, France},
  volume={25},
  year={2015}
}

@article{bischoff2010strategic,
  title={The strategic research agenda for robotics in europe [industrial activities]},
  author={Bischoff, Rainer and Guhl, Tim},
  journal={IEEE Robotics \& Automation Magazine},
  volume={17},
  number={1},
  pages={15--16},
  year={2010},
  publisher={IEEE}
}

@article{daily2017self,
  title={Self-driving cars},
  author={Daily, Mike and Medasani, Swarup and Behringer, Reinhold and Trivedi, Mohan},
  journal={Computer},
  volume={50},
  number={12},
  pages={18--23},
  year={2017},
  publisher={IEEE}
}

@article{romanenko2022robot,
  title={Robot vs worker},
  author={Romanenko, Taras and Shcherbinina, Polina},
  journal={Technology and language},
  volume={3},
  number={1},
  pages={17--28},
  year={2022}
}

@article{yao2017active,
  title={Active disturbance rejection adaptive control of hydraulic servo systems},
  author={Yao, Jianyong and Deng, Wenxiang},
  journal={IEEE Transactions on Industrial Electronics},
  volume={64},
  number={10},
  pages={8023--8032},
  year={2017},
  publisher={IEEE}
}

@article{ren2019adaptive,
  title={Adaptive backstepping control of a pneumatic system with unknown model parameters and control direction},
  author={Ren, Hai-Peng and Wang, Xuan and Fan, Jun-Tao and Kaynak, Okyay},
  journal={IEEE Access},
  volume={7},
  pages={64471--64482},
  year={2019},
  publisher={IEEE}
}

@article{yang2017position,
  title={Position tracking control law for an electro-hydraulic servo system based on backstepping and extended differentiator},
  author={Yang, Xuebo and Zheng, Xiaolong and Chen, Yuhong},
  journal={IEEE/ASME Transactions on Mechatronics},
  volume={23},
  number={1},
  pages={132--140},
  year={2017},
  publisher={IEEE}
}

@article{fassbender2024energy,
  title={Energy-efficiency comparison of different implement powertrain concepts to each other and between different heavy-duty mobile machines},
  author={Fassbender, David and Brach, Christine and Minav, Tatiana},
  journal={International Journal of Fluid Power},
  pages={127--144},
  year={2024}
}

@inproceedings{lee2016model,
  title={Model-free joint torque control strategy for hydraulic robots},
  author={Lee, Woongyong and Kim, Min Jun and Chung, Wan Kyun},
  booktitle={2016 IEEE International Conference on Robotics and Automation (ICRA)},
  pages={2408--2415},
  year={2016},
  organization={IEEE}
}

@article{yao2014nonlinear,
  title={Nonlinear adaptive robust backstepping force control of hydraulic load simulator: Theory and experiments},
  author={Yao, Jianyong and Jiao, Zongxia and Yao, Bin},
  journal={Journal of Mechanical Science and Technology},
  volume={28},
  number={4},
  pages={1499--1507},
  year={2014},
  publisher={Springer}
}

@article{xian2015adaptive,
  title={Adaptive backstepping tracking control of a 6-DOF unmanned helicopter},
  author={Xian, Bin and Guo, Jianchuan and Zhang, Yao},
  journal={IEEE/CAA Journal of Automatica Sinica},
  volume={2},
  number={1},
  pages={19--24},
  year={2015},
  publisher={IEEE}
}

@article{lai2025indirect,
  title={Indirect Adaptive Robust Backstepping Control for Input Delay Systems With Unknown Periodic Disturbances: Theory and Experiments},
  author={Lai, Han and Hu, Jinfei and Xia, Kaiyang and Yao, Bin},
  journal={IEEE/ASME Transactions on Mechatronics},
  year={2025},
  publisher={IEEE}
}

@article{estrada2024super,
  title={Super-twisting based sliding mode control of hydraulic actuator without velocity state},
  author={Estrada, Manuel A and Ruderman, Michael and Fridman, Leonid},
  journal={Control Engineering Practice},
  volume={142},
  pages={105739},
  year={2024},
  publisher={Elsevier}
}

@article{yip1998adaptive,
  author    = {P. P. Yip and J. K. Hedrick},
  title     = {Adaptive dynamic surface control: A simplified algorithm for adaptive backstepping control of nonlinear systems},
  journal   = {International Journal of Control},
  volume    = {71},
  number    = {5},
  pages     = {959--979},
  year      = {1998},
}

@article{liu2021predefined,
  author    = {B. Liu and M. Hou and C. Wu and W. Wang and Z. Wu and B. Huang},
  title     = {Predefined-time backstepping control for a nonlinear strict-feedback system},
  journal   = {International Journal of Robust and Nonlinear Control},
  volume    = {31},
  number    = {8},
  pages     = {3354--3372},
  year      = {2021},
}

@inproceedings{zhang2024adaptive,
  title={Adaptive predefined time control for strict-feedback systems with actuator quantization},
  author={Zhang, Wentong and Yu, Bo},
  booktitle={Actuators},
  volume={13},
  number={9},
  pages={366},
  year={2024},
  organization={MDPI}
}

@article{doctolero2023neural,
  title={Neural-Adaptive Switching Control of Task-Space Objectives on Strict-Feedback Robot},
  author={Doctolero, Samuel C and Westwick, David T},
  journal={IFAC-PapersOnLine},
  volume={56},
  number={2},
  pages={9215--9220},
  year={2023},
  publisher={Elsevier}
}

@article{gaagai2023constrained,
  title={Constrained distributed consensus control of homogeneous vehicle platoons with bidirectional communication},
  author={Gaagai, Ramzi and Horn, Joachim},
  journal={Control Eng. Pract.},
  volume={140},
  pages={105690},
  year={2023},
  publisher={Elsevier}
}

@article{mathieu2011transformation,
  title={Transformation of a mismatched nonlinear dynamic system into strict feedback form},
  author={Mathieu, Johanna L and Hedrick, J Karl},
  year={2011}
}

@article{shahna2021design,
  title={Design of a finite time passivity based adaptive sliding mode control implementing on a spacecraft attitude dynamic simulator},
  author={Shahna, Mehdi Heydari and Abedi, Mostafa},
  journal={Control Eng. Pract.},
  volume={114},
  pages={104866},
  year={2021},
  publisher={Elsevier}
}

@article{zhang2014robust,
  title={Robust control for PWM-based DC--DC buck power converters with uncertainty via sampled-data output feedback},
  author={Zhang, Chuanlin and Wang, Junxiao and Li, Shihua and Wu, Bin and Qian, Chunjiang},
  journal={IEEE Trans. Power Electron.},
  volume={30},
  number={1},
  pages={504--515},
  year={January 2014},
  publisher={IEEE}
}

@article{huang2019practical,
  title={Practical tracking via adaptive event-triggered feedback for uncertain nonlinear systems},
  author={Huang, Yaxin and Liu, Yungang},
  journal={IEEE Trans. Autom. Control},
  volume={64},
  number={9},
  pages={3920--3927},
  year={January 2019},
  publisher={IEEE}
}

@article{zhang2019low,
  title={Low-complexity tracking control of strict-feedback systems with unknown control directions},
  author={Zhang, Jin-Xi and Yang, Guang-Hong},
  journal={IEEE Trans. Autom. Control},
  volume={64},
  number={12},
  pages={5175--5182},
  year={December 2019},
  publisher={IEEE}
}

@article{koivumaki2022subsystem,
  title={Subsystem-based control with modularity for strict-feedback form nonlinear systems},
  author={Koivum{\"a}ki, Janne and Humaloja, Jukka-Pekka and Paunonen, Lassi and Zhu, Wen-Hong and Mattila, Jouni},
  journal={IEEE Trans. Autom. Control},
  volume={68},
  number={7},
  pages={4336--4343},
  year={September 2022},
  publisher={IEEE}
}

@book{zhu2010virtual,
  title={Virtual Decomposition Control: Toward Hyper Degrees of Freedom Robots},
  author={Zhu, Wen-Hong},
  volume={60},
  year={2010},
  publisher={Springer Science \& Business Media}
}

@article{xie2024low,
  title={Low-complexity tracking control of unknown strict-feedback systems with quantitative performance guarantees},
  author={Xie, Haixiu and Jing, Yuanwei and Zhang, Jin-Xi and Dimirovski, Georgi M},
  journal={IEEE Trans. Cybern.},
  volume={54},
  number={10},
  pages={5887-5900},
  year={May 2024},
  publisher={IEEE}
}

@article{oliveira2015global,
  title={Global tracking for a class of uncertain nonlinear systems with unknown sign-switching control direction by output feedback},
  author={Oliveira, Tiago Roux and Peixoto, Alessandro Jacoud and Hsu, Liu},
  journal={Int. J. Control},
  volume={88},
  number={9},
  pages={1895--1910},
  year={April 2015},
  publisher={Taylor \& Francis}
}

@article{liu2006global,
  title={Global robust stabilization of cascade-connected systems with dynamic uncertainties without knowing the control direction},
  author={Liu, Lu and Huang, Jie},
  journal={IEEE Trans. Autom. Control},
  volume={51},
  number={10},
  pages={1693--1699},
  year={October 2006},
  publisher={IEEE}
}

@book{jazar2010theory,
  title={Theory of Applied Robotics},
  author={Jazar, Reza N},
  year={2010},
  publisher={Springer}
}

@article{min2018output,
  title={Output-feedback control for stochastic nonlinear systems subject to input saturation and time-varying delay},
  author={Min, Huifang and Xu, Shengyuan and Zhang, Baoyong and Ma, Qian},
  journal={IEEE Trans. Autom. Control},
  volume={64},
  number={1},
  pages={359--364},
  year={April 2018},
  publisher={IEEE}
}

@article{theodorakopoulos2015low,
  title={Low-complexity prescribed performance control of uncertain MIMO feedback linearizable systems},
  author={Theodorakopoulos, Achilles and Rovithakis, George A},
  journal={IEEE Trans. Autom. Control},
  volume={61},
  number={7},
  pages={1946--1952},
  year={September 2015},
  publisher={IEEE}
}

@article{bikas2018combining,
  title={Combining prescribed tracking performance and controller simplicity for a class of uncertain MIMO nonlinear systems with input quantization},
  author={Bikas, Lampros N and Rovithakis, George A},
  journal={IEEE Trans. Autom. Control},
  volume={64},
  number={3},
  pages={1228--1235},
  year={June 2018},
  publisher={IEEE}
}

@article{wang2015finite,
  title={Finite-time stabilization of high-order stochastic nonlinear systems in strict-feedback form},
  author={Wang, Hui and Zhu, Quanxin},
  journal={Automatica},
  volume={54},
  pages={284--291},
  year={April 2015},
  publisher={Elsevier}
}

@article{li2022event,
  title={Event-based finite-time control for nonlinear multiagent systems with asymptotic tracking},
  author={Li, Yongming and Li, Yuan-Xin and Tong, Shaocheng},
  journal={IEEE Trans. Autom. Control},
  volume={68},
  number={6},
  pages={3790--3797},
  year={June 2022},
  publisher={IEEE}
}

@article{smaoui2006study,
  title={A study on tracking position control of an electropneumatic system using backstepping design},
  author={Smaoui, Mohamed and Brun, Xavier and Thomasset, Daniel},
  journal={Control Eng. Pract.},
  volume={14},
  number={8},
  pages={923--933},
  year={2006},
  publisher={Elsevier}
}

@article{schauer2005online,
  title={Online identification and nonlinear control of the electrically stimulated quadriceps muscle},
  author={Schauer, T and Neg{\aa}rd, N-O and Previdi, Fabio and Hunt, KJ and Fraser, MH and Ferchland, E and Raisch, J},
  journal={Control Eng. Pract.},
  volume={13},
  number={9},
  pages={1207--1219},
  year={2005},
  publisher={Elsevier}
}

@article{humaloja2021decentralized,
  title={Decentralized observer design for virtual decomposition control},
  author={Humaloja, Jukka-Pekka and Koivum{\"a}ki, Janne and Paunonen, Lassi and Mattila, Jouni},
  journal={Trans. Automat. Contr.},
  volume={67},
  number={5},
  pages={2529--2536},
  year={2021},
  publisher={IEEE}
}

@ARTICLE{9161291,
  author={Shan, Yunxiao and Zheng, Boli and Chen, Longsheng and Chen, Long and Chen, De},
  journal={TVT}, 
  title={A Reinforcement Learning-Based Adaptive Path Tracking Approach for Autonomous Driving}, 
  year={2020},
  volume={69},
  number={10},
  pages={10581-10595},
  doi={10.1109/TVT.2020.3014628}}

@inproceedings{ames2019control,
  title={Control barrier functions: Theory and applications},
  author={Ames, Aaron D and Coogan, Samuel and Egerstedt, Magnus and Notomista, Gennaro and Sreenath, Koushil and Tabuada, Paulo},
  booktitle={2019 18th European control conference (ECC)},
  pages={3420--3431},
  year={2019},
  organization={IEEE}
}

@article{rao2017self,
  title={A self-adaptive multi-population based Jaya algorithm for engineering optimization},
  author={Rao, R Venkata and Saroj, Ankit},
  journal={Swarm Evol. Comput.},
  volume={37},
  pages={1--26},
  year={2017},
  publisher={Elsevier}
}

@article{kashani2022population,
  title={Population-based optimization in structural engineering: a review},
  author={Kashani, Ali R and Camp, Charles V and Rostamian, Mehdi and Azizi, Koorosh and Gandomi, Amir H},
  journal={Artif. Intell. Rev.},
  pages={1--108},
  year={2022},
  publisher={Springer}
}

@article{liang2023adaptive,
  title={Adaptive force tracking impedance control for aerial interaction in uncertain contact environment using barrier function},
  author={Liang, Jiacheng and Zhong, Hang and Wang, Yaonan and Chen, Yanjie and Zeng, Junhao and Mao, Jianxu},
  journal={IEEE Transactions on Automation Science and Engineering},
  year={2023},
  publisher={IEEE}
}

@techreport{iso5469,
  title        = {Artificial intelligence — Functional safety and AI systems},
  institution  = {International Organization for Standardization},
  number       = {ISO/IEC TR 5469:2024},
  year         = {2024},
  url          = {https://www.iso.org/standard/81283.html}
}

@article{liu2019barrier,
  title={Barrier {L}yapunov function-based adaptive fuzzy FTC for switched systems and its applications to resistance--inductance--capacitance circuit system},
  author={Liu, Lei and Liu, Yan-Jun and Li, Dapeng and Tong, Shaocheng and Wang, Zhanshan},
  journal={IEEE Transactions on Cybernetics},
  volume={50},
  number={8},
  pages={3491--3502},
  year={2019},
  publisher={IEEE}
}

@article{shahna2025lidar,
  title={LiDAR-Inertial SLAM-Based Navigation and Safety-Oriented AI-Driven Control System for Skid-Steer Robots},
  author={Shahna, Mehdi Heydari and Haaparanta, Eemil and Mustalahti, Pauli and Mattila, Jouni},
  journal={IEEE Conference on Decision and Control (CDC)},
  year={2025}
}

@article{hulttinen2021model,
  title={Model-Based Control of a Mobile Platform With Independently Controlled In-Wheel Hydraulic Motors},
  author={Hulttinen, Lionel and Mattila, Jouni},
  journal={Journal of Fluid Power Systems Technology},
  volume={85239},
  pages={V001T01A001},
  year={2021},
  publisher={American Society of Mechanical Engineers}
}

@article{shahna2025fault,
  title={Fault-Tolerant Control for System Availability and Continuous Operation in Heavy-Duty Wheeled Mobile Robots},
  author={Shahna, Mehdi Heydari and Mustalahti, Pauli and Mattila, Jouni},
  journal={The IEEE/ASME International Conference on Advanced Intelligent Mechatronics (AIM)},
  year={2025}
}

@inproceedings{shahna2024integrating,
  title={Integrating deeprl with robust low-level control in robotic manipulators for non-repetitive reaching tasks},
  author={Shahna, Mehdi Heydari and Kolagar, Seyed Adel Alizadeh and Mattila, Jouni},
  booktitle={2024 IEEE International Conference on Mechatronics and Automation (ICMA)},
  pages={329--336},
  year={2024},
  organization={IEEE}
}

@article{kolagar2024combining,
  title={Combining Deep Reinforcement Learning with a Jerk-Bounded Trajectory Generator for Kinematically Constrained Motion Planning},
  author={Kolagar, Seyed Adel Alizadeh and Shahna, Mehdi Heydari and Mattila, Jouni},
  journal={European Control Conference (ECC)},
  year={2025}
}

@inproceedings{shahna2024sasPEMCrobust,
  title={Robust model-free control framework with safety constraints for a fully electric linear actuator system},
  author={Shahna, Mehdi Heydari and Mustalahti, Pauli and Mattila, Jouni},
  booktitle={2024 IEEE 21st International Power Electronics and Motion Control Conference (PEMC)},
  pages={1--10},
  year={2024},
  organization={IEEE}
}

@article{shahna2024modelacc,
  title={Model-free generic robust control for servo-driven actuation mechanisms with layered insight into energy conversions},
  author={Shahna, Mehdi Heydari and Mattila, Jouni},
  journal={American Control Conference (ACC)},
  year={2024}
}

@article{li2009adaptive,
  title={Adaptive speed control for permanent-magnet synchronous motor system with variations of load inertia},
  author={Li, Shihua and Liu, Zhigang},
  journal={ IEEE Trans. Ind. Electron.},
  volume={56},
  number={8},
  pages={3050--3059},
  year={2009},
  publisher={IEEE}
}

@article{olaru2004new,
  title={A new model to estimate friction torque in a ball screw system},
  author={Olaru, Dumitru and Puiu, George C and Balan, Liviu C and Puiu, Vasile},
  journal={Product engineering},
  pages={333--346},
  year={2004},
  publisher={Springer}
}

@article{sirouspour2002nonlinear,
  title={Nonlinear control of hydraulic robots},
  author={Sirouspour, Mohammad Reza and Salcudean, Septimiu E},
  journal={IEEE Transactions on Robotics and Automation},
  volume={17},
  number={2},
  pages={173--182},
  year={2002},
  publisher={IEEE}
}

@article{malagari2012globally,
  title={Globally exponential controller/observer for tracking in robots without velocity measurement},
  author={Malagari, Srinivasulu and Driessen, Brian J},
  journal={Asian Journal of Control},
  volume={14},
  number={2},
  pages={309--319},
  year={2012},
  publisher={Wiley Online Library}
}

@article{berghuis2002passivity,
  title={A passivity approach to controller-observer design for robots},
  author={Berghuis, Harry and Nijmeijer, Henk},
  journal={IEEE Transactions on robotics and automation},
  volume={9},
  number={6},
  pages={740--754},
  year={2002},
  publisher={IEEE}
}

@inproceedings{bu2000observer,
  title={Observer based coordinated adaptive robust control of robot manipulators driven by single-rod hydraulic actuators},
  author={Bu, Fanping and Yao, Bin},
  booktitle={Proceedings 2000 ICRA. Millennium Conference. IEEE International Conference on Robotics and Automation. Symposia Proceedings (Cat. No. 00CH37065)},
  volume={3},
  pages={3034--3039},
  year={2000},
  organization={IEEE}
}

@article{zhang2017backstepping,
  title={Backstepping based adaptive region tracking fault tolerant control for autonomous underwater vehicles},
  author={Zhang, Mingjun and Liu, Xing and Wang, Fei},
  journal={The Journal of Navigation},
  volume={70},
  number={1},
  pages={184--204},
  year={2017},
  publisher={Cambridge University Press}
}

@article{truong2022backstepping,
  title={Backstepping sliding mode-based model-free control of electro-hydraulic systems},
  author={Truong, Hoai-Vu-Anh and Trinh, Hoai-An and Ahn, Kyoung-Kwan},
  journal={Journal of Drive and Control},
  volume={19},
  number={1},
  pages={51--61},
  year={2022},
  publisher={The Korean Society for Fluid Power and Construction Equipment}
}

@article{ebrahimi2018model,
  title={Model-free sliding mode control, theory and application},
  author={Ebrahimi, Nahid and Ozgoli, Sadjaad and Ramezani, Amin},
  journal={Proceedings of the Institution of Mechanical Engineers, Part I: Journal of Systems and Control Engineering},
  volume={232},
  number={10},
  pages={1292--1301},
  year={2018},
  publisher={SAGE Publications Sage UK: London, England}
}

@article{roman2016data,
  title={Data-driven model-free adaptive control tuned by virtual reference feedback tuning},
  author={Roman, Raul-Cristian and Radac, Mircea-Bogdan and Precup, Radu-Emil and Petriu, Emil M},
  journal={Acta Polytechnica Hungarica},
  volume={13},
  number={1},
  pages={83--96},
  year={2016}
}

@article{jalali2013model,
  title={Model-free adaptive fuzzy sliding mode controller optimized by particle swarm for robot manipulator},
  author={Jalali, Amin and Piltan, Farzin and Gavahian, Atefeh and Jalali, Meysam and others},
  journal={International Journal of Information Engineering and Electronic Business},
  volume={5},
  number={1},
  pages={68},
  year={2013},
  publisher={Modern Education and Computer Science Press}
}

@inproceedings{baciu2021iterative,
  title={Iterative feedback tuning of model-free controllers},
  author={Baciu, Andrei and Lazar, Corneliu and Caruntu, Constantin-Florin},
  booktitle={2021 25th International Conference on System Theory, Control and Computing (ICSTCC)},
  pages={467--472},
  year={2021},
  organization={IEEE}
}

@article{fliess2009model,
  title={Model-free control and intelligent PID controllers: towards a possible trivialization of nonlinear control?},
  author={Fliess, Michel},
  journal={IFAC Proceedings Volumes},
  volume={42},
  number={10},
  pages={1531--1550},
  year={2009},
  publisher={Elsevier}
}

@book{brosilow2002techniques,
  title={Techniques of model-based control},
  author={Brosilow, Coleman and Joseph, Babu},
  year={2002},
  publisher={Prentice Hall Professional}
}

@article{kool2016does,
  title={When does model-based control pay off?},
  author={Kool, Wouter and Cushman, Fiery A and Gershman, Samuel J},
  journal={PLoS computational biology},
  volume={12},
  number={8},
  pages={e1005090},
  year={2016},
  publisher={Public Library of Science San Francisco, CA USA}
}

@book{hou2013model,
  title={Model free adaptive control},
  author={Hou, Zhongsheng and Jin, Shangtai},
  year={2013},
  publisher={CRC press Boca Raton, FL, USA}
}

@book{precup2021data,
  title={Data-driven model-free controllers},
  author={Precup, Radu-Emil and Roman, Raul-Cristian and Safaei, Ali},
  year={2021},
  publisher={CRC Press}
}

@article{wang2022research,
  title={Research on model-free adaptive control of electro-hydraulic servo system of continuous rotary motor},
  author={Wang, Xiaojing and Zhang, Yang and Li, Chunhui},
  journal={IEEE Access},
  volume={10},
  pages={31165--31174},
  year={2022},
  publisher={IEEE}
}

@article{zhang2024robust,
  title={Robust fixed-time adaptive fault-tolerant control for dynamic positioning of ships with thruster faults},
  author={Zhang, Yuanyuan and Zhang, Jianqiang and Sui, Bowen},
  journal={Applied Sciences},
  volume={14},
  number={13},
  pages={5738},
  year={2024},
  publisher={MDPI}
}

@article{zhao2022dual,
  title={Dual-triggered adaptive torque control strategy for variable-speed wind turbine against denial-of-service attacks},
  author={Zhao, Shiyi and Xia, Jinhui and Deng, Ruilong and Cheng, Peng and Yang, Qinmin and Jiao, Xuguo},
  journal={IEEE Transactions on Smart Grid},
  volume={14},
  number={4},
  pages={3072--3084},
  year={2022},
  publisher={IEEE}
}

@article{schwarz2024robust,
  title={Robust identification and control of mobile hydraulic systems using a decentralized valve structure},
  author={Schwarz, Johannes and Lohmann, Boris},
  journal={Control Engineering Practice},
  volume={151},
  pages={106030},
  year={2024},
  publisher={Elsevier}
}

@article{xiao2021high,
  title={High-order control barrier functions},
  author={Xiao, Wei and Belta, Calin},
  journal={IEEE Transactions on Automatic Control},
  volume={67},
  number={7},
  pages={3655--3662},
  year={2021},
  publisher={IEEE}
}

@article{liu2023command,
  title={Command-filter-approximator-based adaptive control for uncertain nonlinear systems and its application in PMSMs},
  author={Liu, Jiapeng and Wang, Qing-Guo and Yu, Jinpeng},
  journal={IEEE Transactions on Systems, Man, and Cybernetics: Systems},
  volume={53},
  number={11},
  pages={6828--6835},
  year={2023},
  publisher={IEEE}
}

@article{zhang2023adaptive,
  title={Adaptive neural asymptotic tracking control for PMSM systems under current constraints and unknown dynamics},
  author={Zhang, Jianyi and Ren, Wei and Li, Jingjie and Sun, Xi-Ming},
  journal={IEEE Transactions on Circuits and Systems II: Express Briefs},
  volume={71},
  number={2},
  pages={777--781},
  year={2023},
  publisher={IEEE}
}

@article{sharun2012adaptive,
  title={Adaptive neuro-controller based on hybrid multi layered perceptron network for dynamic systems},
  author={Sharun, SM and Mashor, MY and Jaafar, WNH Wan and Nazid, N Mohd and Yaacob, S},
  journal={International Journal of Control Science and Engineering},
  volume={2},
  number={3},
  pages={34--41},
  year={2012}
}

@article{astrom2006advanced,
  title={Advanced PID control},
  author={Astrom, Karl J and H{\"a}gglund, Tore},
  journal={IEEE Control Systems},
  volume={26},
  number={1},
  pages={98--101},
  year={2006}
}

@article{driessen2015observer,
  title={Observer/controller with global practical stability for tracking in robots without velocity measurement},
  author={Driessen, Brian J},
  journal={Asian Journal of Control},
  volume={17},
  number={5},
  pages={1898--1913},
  year={2015},
  publisher={Wiley Online Library}
}

@article{zhu1997virtual,
  title={Virtual decomposition based control for generalized high dimensional robotic systems with complicated structure},
  author={Zhu, Wen-Hong and Xi, Yu-Geng and Zhang, Zhong-Jun and Bien, Zeungnam and De Schutter, Joris},
  journal={IEEE Transactions on Robotics and Automation},
  volume={13},
  number={3},
  pages={411--436},
  year={1997},
  publisher={IEEE}
}

@article{koivumaki2017adaptive,
  title={Adaptive and nonlinear control of discharge pressure for variable displacement axial piston pumps},
  author={Koivum{\"a}ki, Janne and Mattila, Jouni},
  journal={Journal of Dynamic Systems, Measurement, and Control},
  volume={139},
  number={10},
  pages={101008},
  year={2017},
  publisher={American Society of Mechanical Engineers}
}

@article{mattila2017survey,
  title={A survey on control of hydraulic robotic manipulators with projection to future trends},
  author={Mattila, Jouni and Koivum{\"a}ki, Janne and Caldwell, Darwin G and Semini, Claudio},
  journal={iEeE/ASME Transactions on Mechatronics},
  volume={22},
  number={2},
  pages={669--680},
  year={2017},
  publisher={IEEE}
}

@article{mastellone2021impact,
  title={The impact of control research on industrial innovation: What would it take to make it happen?},
  author={Mastellone, Silvia and van Delft, Alex},
  journal={Control Engineering Practice},
  volume={111},
  pages={104737},
  year={2021},
  publisher={Elsevier}
}

@article{chen2022adaptive,
  title={Adaptive fixed-time tracking control for nonlinear systems based on finite-time command-filtered backstepping},
  author={Chen, Ming and Li, Yingsen and Wang, Huanqing and Peng, Kaixiang and Wu, Libing},
  journal={IEEE Transactions on Fuzzy Systems},
  volume={31},
  number={5},
  pages={1604--1613},
  year={2022},
  publisher={IEEE}
}

@article{zheng2023practical,
  title={Practical finite-time command filtered backstepping with its application to DC motor control systems},
  author={Zheng, Xiaolong and Yu, Xinghu and Jiang, Jiaxu and Yang, Xuebo},
  journal={IEEE Transactions on Industrial Electronics},
  volume={71},
  number={3},
  pages={2955--2964},
  year={2023},
  publisher={IEEE}
}

@article{sun2016neural,
  title={Neural network-based DOBC for a class of nonlinear systems with unmatched disturbances},
  author={Sun, Haibin and Guo, Lei},
  journal={IEEE transactions on neural networks and learning systems},
  volume={28},
  number={2},
  pages={482--489},
  year={2016},
  publisher={IEEE}
}

@article{fan2023sliding,
  title={Sliding mode control for strict-feedback nonlinear impulsive systems with matched disturbances},
  author={Fan, Debao and Zhang, Xianfu and Pan, Weihao},
  journal={IEEE Transactions on Circuits and Systems II: Express Briefs},
  volume={71},
  number={2},
  pages={787--791},
  year={2023},
  publisher={IEEE}
}

@article{zerari2021event,
  title={Event-triggered adaptive output-feedback neural-networks control for saturated strict-feedback nonlinear systems in the presence of external disturbance},
  author={Zerari, Nassira and Chemachema, Mohamed},
  journal={Nonlinear Dynamics},
  volume={104},
  number={2},
  pages={1343--1362},
  year={2021},
  publisher={Springer}
}

@article{yin2015adaptive,
  title={Adaptive fuzzy control of strict-feedback nonlinear time-delay systems with unmodeled dynamics},
  author={Yin, Shen and Shi, Peng and Yang, Hongyan},
  journal={IEEE transactions on cybernetics},
  volume={46},
  number={8},
  pages={1926--1938},
  year={2015},
  publisher={IEEE}
}

@article{bahari2024system,
  title={System-Level Efficient Performance of {EMLA}-Driven Heavy-Duty Manipulators via Bilevel Optimization Framework with a Leader-Follower Scenario},
  author={Bahari, Mohammad and Paz, Alvaro and Shahna, Mehdi Heydari and Mustalahti, Pauli and Mattila, Jouni},
  journal={IEEE Trans. Autom. Sci. Eng.},
  year={2025},
  publisher={IEEE}
}

@book{tarbouriech2007advanced,
  title={Advanced strategies in control systems with input and output constraints},
  author={Tarbouriech, Sophie and Garcia, Germain and Glattfelder, Adolf H},
  volume={346},
  year={2007},
  publisher={Springer}
}

@article{karafyllis2025deadzone,
  title={Deadzone-Adapted Disturbance Suppression Control for strict-feedback systems},
  author={Karafyllis, Iasson and Krstic, Miroslav and Aslanidis, Alexandros},
  journal={Automatica},
  volume={171},
  pages={111986},
  year={2025},
  publisher={Elsevier}
}

@article{estrada2025hydraulic,
  title={Hydraulic actuator control based on continuous higher order sliding modes},
  author={Estrada, Manuel A and Ruderman, Michael and Texis-Loaiza, Oscar and Fridman, Leonid},
  journal={Control Engineering Practice},
  volume={156},
  pages={106218},
  year={2025},
  publisher={Elsevier}
}

@article{yin2024fault,
  title={Fault diagnosis of pressure relief valve based on improved deep Residual Shrinking Network},
  author={Yin, Hao and Xu, He and Fan, Weiwang and Sun, Feng},
  journal={Measurement},
  volume={224},
  pages={113752},
  year={2024},
  publisher={Elsevier}
}

@article{bahari2019new,
  title={A new variable reluctance PM-resolver},
  author={Bahari, Mohammad and Davoodi, A and Saneie, H and Tootoonchian, F and Nasiri-Gheidari, Z},
  journal={IEEE Sensors Journal},
  volume={20},
  number={1},
  pages={135--142},
  year={2019},
  publisher={IEEE}
}

@article{shahna2024robustness,
  title={Robustness-Guaranteed Observer-Based Control Strategy With Modularity for Cleantech EMLA-Driven Heavy-Duty Robotic Manipulator},
  author={Shahna, Mehdi Heydari and Bahari, Mohammad and Mattila, Jouni},
  journal={IEEE Transactions on Automation Science and Engineering},
  year={2024},
  publisher={IEEE}
}

@article{yang2016disturbance,
  title={Disturbance/uncertainty estimation and attenuation techniques in PMSM drives—A survey},
  author={Yang, Jun and Chen, Wen-Hua and Li, Shihua and Guo, Lei and Yan, Yunda},
  journal={IEEE Transactions on Industrial Electronics},
  volume={64},
  number={4},
  pages={3273--3285},
  year={2016},
  publisher={IEEE}
}

@article{wang2023concurrent,
  title={Concurrent learning control {L}yapunov and barrier functions for uncertain nonlinear safety-critical systems with high relative degree constraints},
  author={Wang, Liqi and Dong, Jiuxiang},
  journal={IEEE Transactions on Automation Science and Engineering},
  volume={21},
  number={4},
  pages={7170--7179},
  year={2023},
  publisher={IEEE}
}

@inproceedings{jin2003trade,
  title={Trade-off between performance and robustness: An evolutionary multiobjective approach},
  author={Jin, Yaochu and Sendhoff, Bernhard},
  booktitle={International conference on evolutionary multi-criterion optimization},
  pages={237--251},
  year={2003},
  organization={Springer}
}

@article{badue2021self,
  title={Self-driving cars: A survey},
  author={Badue, Claudine and Guidolini, R{\^a}nik and Carneiro, Raphael Vivacqua and Azevedo, Pedro and Cardoso, Vinicius B and Forechi, Avelino and Jesus, Luan and Berriel, Rodrigo and Paixao, Thiago M and Mutz, Filipe and others},
  journal={Expert systems with applications},
  volume={165},
  pages={113816},
  year={2021},
  publisher={Elsevier}
}

@article{katreddi2022review,
  title={A review of applications of artificial intelligence in heavy duty trucks},
  author={Katreddi, Sasanka and Kasani, Sujan and Thiruvengadam, Arvind},
  journal={Energies},
  volume={15},
  number={20},
  pages={7457},
  year={2022},
  publisher={MDPI}
}

@article{leung2023automation,
  title={Automation and Artificial Intelligence Technology in Surface Mining: A Brief Introduction to Open-Pit Operations in the Pilbara},
  author={Leung, Raymond and Hill, Andrew J and Melkumyan, Arman},
  journal={IEEE Robotics \& Automation Magazine},
  year={2023},
  publisher={IEEE}
}

@article{chi2017avoiding,
  title={Avoiding the health hazard of people from construction vehicles: a strategy for controlling the vibration of a wheel loader},
  author={Chi, Feng and Zhou, Jun and Zhang, Qi and Wang, Yong and Huang, Panling},
  journal={International journal of environmental research and public health},
  volume={14},
  number={3},
  pages={275},
  year={2017},
  publisher={MDPI}
}

@inproceedings{machado2021autonomous,
  title={Autonomous heavy-duty mobile machinery: A multidisciplinary collaborative challenge},
  author={Machado, Tyrone and Fassbender, David and Taheri, Abdolreza and Eriksson, Daniel and Gupta, Himanshu and Molaei, Amirmasoud and Forte, Paolo and Rai, Prashant Kumar and Ghabcheloo, Reza and M{\"a}kinen, Saku and others},
  booktitle={2021 IEEE International Conference on Technology and Entrepreneurship (ICTE)},
  pages={1--8},
  year={2021},
  organization={IEEE}
}

@book{klancar2017wheeled,
  title={Wheeled mobile robotics: from fundamentals towards autonomous systems},
  author={Klancar, Gregor and Zdesar, Andrej and Blazic, Saso and Skrjanc, Igor},
  year={2017},
  publisher={Butterworth-Heinemann}
}

@inproceedings{adams2025human,
  title={Human-Robot Teaming Directions for Dull, Dirty and Dangerous Domains},
  author={Adams, Julie A},
  booktitle={2025 20th ACM/IEEE International Conference on Human-Robot Interaction (HRI)},
  pages={1--1},
  year={2025},
  organization={IEEE}
}

@article{herman1998renaissance,
  title={A Renaissance robot},
  author={Herman, David},
  journal={Mechanical Engineering},
  volume={120},
  number={02},
  pages={80--82},
  year={1998},
  publisher={American Society of Mechanical Engineers}
}

@article{shahna2025anti,
  title={Anti-Slip AI-Driven Model-Free Control with Global Exponential Stability in Skid-Steering Robots},
  author={Shahna, Mehdi Heydari and Mustalahti, Pauli and Mattila, Jouni},
  journal={IEEE/RSJ International Conference on Intelligent Robots and Systems (IROS)},
  year={2025}
}

@article{cep2,
  title={Model Reference-based Control with Guaranteed Predefined Performance for Uncertain
Strict-Feedback Systems},
  author={Shahna, Mehdi Heydari and Humaloja, Jukka-Pekka and Mattila, Jouni},
  journal={Control Engineering Practice},
  volume={--},
  pages={---},
  year={2025},
  publisher={Elsevier}
}


\end{document}